%% file: root.tex
\documentclass[12pt,oneside,final]{thesis}

\makeatletter
\renewcommand{\normalsize}{\@setfontsize\normalsize\@xipt{13.6}%
\abovedisplayskip 11\p@ plus3\p@ minus6\p@
\belowdisplayskip \abovedisplayskip
\abovedisplayshortskip  \z@ plus3\p@   
\belowdisplayshortskip  6.5\p@ plus3.5\p@ minus3\p@
\let\@listi\@listI}   
\renewcommand{\small}{\@setfontsize\small\@xpt{12}%
\abovedisplayskip 10\p@ plus2\p@ minus5\p@
\belowdisplayskip \abovedisplayskip
\abovedisplayshortskip  \z@ plus3\p@   
\belowdisplayshortskip  6\p@ plus3\p@ minus3\p@
\def\@listi{\leftmargin\leftmargini 
\topsep 6\p@ plus2\p@ minus2\p@\parsep 3\p@ plus2\p@ minus\p@
\itemsep \parsep}}
\renewcommand{\footnotesize}{\@setfontsize\footnotesize\@ixpt{11}%
\abovedisplayskip 8\p@ plus2\p@ minus4\p@
\belowdisplayskip \abovedisplayskip
\abovedisplayshortskip \z@ plus\p@
\belowdisplayshortskip 4\p@ plus2\p@ minus2\p@
\def\@listi{\leftmargin\leftmargini 
\topsep 4\p@ plus2\p@ minus2\p@\parsep 2\p@ plus\p@ minus\p@
\itemsep \parsep}}
\renewcommand{\scriptsize}{\@setfontsize\scriptsize\@viiipt{9.5pt}} 
\renewcommand{\tiny}{\@setfontsize\tiny\@vipt{7pt}} 
\renewcommand{\large}{\@setfontsize\large\@xiipt{14pt}} 
\renewcommand{\Large}{\@setfontsize\Large\@xivpt{18pt}} 
\renewcommand{\LARGE}{\@setfontsize\LARGE\@xviipt{22pt}} 
\renewcommand{\huge}{\@setfontsize\huge\@xxpt{25pt}} 
\renewcommand{\Huge}{\@setfontsize\Huge\@xxvpt{30pt}} 
\makeatother

\usepackage[english]{babel} %
\usepackage{blindtext} %

\usepackage[style=authoryear-comp,citestyle=authoryear,natbib=true,backend=bibtex,maxbibnames=99,sorting=nyt,isbn=false,doi=false,dashed=false,url=true]{biblatex}
\newcommand{\newcite}[1]{\citet{#1}}
\renewcommand{\cite}[1]{\citep{#1}}

\usepackage[a-1b]{pdfx} %
\usepackage{amsmath,amsfonts}
\usepackage{amssymb}
\usepackage{amsthm}
\usepackage{graphicx}
\usepackage{array}
\usepackage{hyperref} 
\usepackage{setspace}
\usepackage{multirow}

\usepackage{times}
\usepackage{latexsym}
\usepackage{url}
\usepackage{microtype}
\usepackage{xfrac}
\usepackage{mathtools}
\usepackage{tikz}
\usepackage{caption}
\usetikzlibrary{shapes.geometric}
\usetikzlibrary{automata, positioning, arrows}
\usepackage{booktabs}
\usepackage{adjustbox}
\usepackage{xspace}
\usepackage[normalem]{ulem}
\usepackage{subfig}
\usepackage{pifont}
\usepackage{bbm}
\usepackage{xcolor}
\usepackage{algorithm}
\usepackage[noend]{algpseudocode}
\usepackage[table]{colortbl}%
\usepackage{soul}
\usepackage{arydshln}
\usepackage{enumitem}
\usepackage{gensymb}

\usepackage{enumitem}
\newlist{inlinelist}{enumerate*}{1}
\setlist*[inlinelist,1]{%
  label=(\arabic*),
}

\usepackage{fancyhdr}    %

\fancypagestyle{plain}{%
\fancyhf{} %
\fancyfoot[C]{\thepage} %

}

\usepackage{cleveref}
\crefname{section}{\S}{\S\S}
\crefname{table}{Tab.}{}
\crefname{figure}{Fig.}{}
\crefname{algorithm}{Alg.}{}
\crefname{equation}{eq.}{}
\crefname{appendix}{App.}{}
\crefformat{section}{\S#2#1#3}  %

\input{root-header}

\begin{document}

\title{\uppercase{How Do Multilingual Encoders Learn Cross-lingual Representation?}}
\author{Shijie Wu}
\degreemonth{January}
\degreeyear{2022}
\dissertation
\doctorphilosophy
\copyrightnotice

\include{src/frontmatter}

\chapter{Introduction}
\label{chap:intro}
\chaptermark{Introduction}
\include{src/introduction}

\chapter{Representation Learning in NLP}
\label{chap:background}
\chaptermark{Representation Learning in NLP}
\include{src/background}

\chapter{Does mBERT Learn Cross-lingual Representation?}
\label{chap:surprising-mbert}
\chaptermark{Does mBERT Learn Cross-lingual Representation?}
\include{src/surprise}

\chapter{How Does mBERT Learn Cross-lingual Representation?}
\label{chap:emerging-structure}
\chaptermark{How Does mBERT Learn Cross-lingual Representation?}
\include{src/emerging}

\chapter{Are All Languages Created Equal in mBERT?}
\label{chap:low-resource}
\chaptermark{Are All Languages Created Equal in mBERT?}
\include{src/low-resource}

\chapter{How To Inject Cross-lingual Signals Into Multilingual Encoders?}
\label{chap:crosslingual-signal}
\chaptermark{How To Inject Cross-lingual Signals Into Multilingual Encoders?}
\include{src/crosslingual-signal}

\chapter{Why Does Zero-shot Cross-lingual Transfer Have High Variance?}
\label{chap:analysis}
\chaptermark{Why Does Zero-shot Cross-lingual Transfer Have High Variance?}
\include{src/analysis}

\chapter{Do Data Projection and Self-training Constrain Zero-shot Cross-lingual Transfer Optimization?}
\label{chap:data-projection}
\chaptermark{Do Data Projection and Self-training Constrain Zero-shot Cross-lingual Transfer Optimization?}
\include{src/data-projection}

\chapter{Conclusions}
\label{chap:conclusions}
\chaptermark{Conclusions}
\include{src/conclusions}

\printbibliography

\include{src/CV}

\end{document}

%% file: root-header.tex
\usepackage{csquotes}
\addto\extrasenglish{

}

\newcommand{\LN}{\text{LN}\xspace}
\newcommand{\Skip}{\text{Skip}\xspace}
\newcommand{\dropout}{\text{Dropout}\xspace}
\newcommand{\FF}{\text{FF}\xspace}
\newcommand{\SA}{\text{MHSA}\xspace}

\newcommand{\gelu}{\text{GELU}\xspace}
\newcommand{\concat}{\text{Concat}\xspace}
\newcommand{\yy}{\mathbf{y}}
\newcommand{\xx}{\mathbf{x}}
\newcommand{\bb}{\mathbf{b}}
\newcommand{\ptype}{p_\text{type}}
\newcommand{\ptoken}{p_\text{token}}
\newcommand{\WW}{\mathbf{W}}
\newcommand{\calR}{\mathbb{R}}
\newcommand{\concurrent}{\textdagger\xspace}
\newcommand{\bitext}{\ensuremath{\spadesuit}\xspace}
\newcommand{\devselect}{\ensuremath{\diamondsuit}\xspace}

\newcommand{\src}{$\mathbf{S}$\space}
\newcommand{\trg}{$\mathbf{T}$\space}
\definecolor{bad}{RGB}{245, 121, 58}
\newcommand{\bad}{\cellcolor{bad}}
\definecolor{good}{RGB}{133, 192, 249}
\newcommand{\good}{\cellcolor{good}}

\newcommand\awesome{Awesome-align}

\definecolor{best}{RGB}{255, 255, 0} %
\newcommand{\best}{\cellcolor{best}}
\definecolor{muchworse}{RGB}{255, 100, 100} %
\newcommand{\muchworse}{\cellcolor{muchworse}}
\definecolor{muchbetter}{RGB}{100, 255, 100} %
\newcommand{\muchbetter}{\cellcolor{muchbetter}}

\definecolor{worse}{RGB}{255, 200, 200} %
\newcommand{\worse}{\cellcolor{worse}}
\definecolor{better}{RGB}{200, 255, 200} %
\newcommand{\better}{\cellcolor{better}}

%% file: src/frontmatter.tex
\begin{frontmatter}

\maketitle

\begin{abstract}

NLP systems typically require support for more than one language. As different languages have different amounts of supervision, cross-lingual transfer benefits languages with little to no training data by transferring from other languages. From an engineering perspective, multilingual NLP benefits development and maintenance by serving multiple languages with a single system. Both cross-lingual transfer and multilingual NLP rely on cross-lingual representations serving as the foundation. As BERT revolutionized representation learning and NLP, it also revolutionized cross-lingual representations and cross-lingual transfer. Multilingual BERT was released as a replacement for single-language BERT, trained with Wikipedia data in 104 languages. 

Surprisingly, without any explicit cross-lingual signal, multilingual BERT learns cross-lingual representations in addition to representations for individual languages. This thesis first shows such surprising cross-lingual effectiveness compared against prior art on various tasks. Naturally, it raises a set of questions, most notably how do these multilingual encoders learn cross-lingual representations. In exploring these questions, this thesis will analyze the behavior of multilingual models in a variety of settings on high and low resource languages. We also look at how to inject different cross-lingual signals into multilingual encoders, and the optimization behavior of cross-lingual transfer with these models. Together, they provide a better understanding of multilingual encoders on cross-lingual transfer. Our findings will lead us to suggested improvements to multilingual encoders and cross-lingual transfer.

{\bf{Readers:}} Mark Dredze, Benjamin Van Durme, João Sedoc
\end{abstract}

\begin{acknowledgment}

This thesis is the product of 39 months of work. In this journey, over 20 months were spent in my home without any in-person meeting with my advisor and labmates, due to the global pandemic known as COVID-19. It was not the easiest time. I owe gratitude to many people who helped me throughout my Ph.D. journey, as well as many more who made that journey possible. I would try my best to thank everyone who helped me along this journey.

I am tremendously thankful to my adviser, Mark Dredze. Mark recruited me to stay at Hopkins and has supported me throughout my Ph.D. journey. Mark offered me the freedom to dive into the research of multilingual encoders and guided me on both research and life as a grad student.

I would also like to thank Benjamin Van Durme and João Sedoc for serving on my
  committee and for mentoring me during the IARPA BETTER program.\footnote{\#2019-19051600005}. 

Before my Ph.D. journey, I spent two years at Hopkins as a Master student. I am thankful to Jason Eisner, who taught the NLP course, sparked my interest in NLP, and offered me an opportunity in NLP research. I am also thankful to Ryan Cotterell, who has collaborated with me on several papers.

No man is an island. I would like to thank my other co-authors: Aaron, Alex, Mans, Alexis, Chaitanya, Craig, Edo, Guanghui, Haoran, Haoran, Jialiang, Kenton, Luke, Mahsa, Marc, Micha, Nizar, Pamela, Patrick, Seth, Tim, Ves, and Yunmo. I would also like to thank everyone who contributed to the SIGMORPHON shared tasks in 2019, 2020, and 2021. While some of the works are not part of this thesis, this thesis will use plural pronouns after these acknowledgments to reflect that this work is not all mine.

My most heartfelt thanks goes to my family. My parents supported my curiosity in science and my pursuit of education. They gave me the freedom and support to pursue my interests, from taking me to the library to letting me pick any extracurricular classes. I could not enumerate how much they have helped me throughout my education journey. I would also like to thank my dog Lucky, who brings me joy in the final month of this journey during the writing of this thesis. My final thanks goes to Gege, who has shown me love and support every day in the past seven years.

\end{acknowledgment}

\begin{dedication}

As Max Weber put it in his speech ``Science as a Vocation'',
\begin{quote}
In science, each of us knows that what he has accomplished will be antiquated in ten, twenty, fifty years. That is the fate to which science is subjected; it is the very meaning of scientific work, to which it is devoted in a quite specific sense, as compared with other spheres of culture for which in general the same holds. Every scientific `fulfilment' raises new `questions'; it asks to be `surpassed' and outdated. Whoever wishes to serve science has to resign himself to this fact. Scientific works certainly can last as `gratifications' because of their artistic quality, or they may remain important as a means of training. Yet they will be surpassed scientifically--let that be repeated--for it is our common fate and, more, our common goal. We cannot work without hoping that others will advance further than we have. In principle, this progress goes on ad infinitum. 
\end{quote}
I dreamed to become a scientist when I was a kid, but I did not understand what it meant until I read this paragraph. NLP, ML, and AI change much faster today than science did a hundred years ago. What you and I accomplished would be antiquated in less than two or even one year, so that together as a community, we could keep moving forward faster and faster. Therefore, I am dedicating this thesis to the ad infinitum of scientific progress, answering open questions and pushing the frontier forward. I hope that I get to witness the modeling of human language on the same level or even surpass our brain one day.

\end{dedication}

\tableofcontents

\listoftables

\listoffigures

\end{frontmatter}

%% file: src/introduction.tex
Modern NLP applications typically require support for more than one language, and we want to build an equally good system for each language. As different languages usually do not have the same amount of supervision, for languages with the least or even no training data, we may rely on cross-lingual transfer---transferring knowledge from languages with more supervision to languages with less or even no supervision. From an engineering perspective, managing different systems for different languages introduces challenges for continuous development and maintenance. Thus, even if we had enough training data for each language to build one system per language, therefore eliminating the need for cross-lingual transfer, we may still want a single system for all languages, namely a multilingual system.

Cross-lingual transfer and multilingual NLP both greatly benefit from cross-lingual representation. Supposed we have access to perfect cross-lingual representation space, i.e. words with similar meaning across languages have similar vector representation, transferring knowledge across language would be straight-forward. Similarly, with such representation, multilingual NLP models only need to learn to solve the task without worrying how to encode words into vectors, leaving much less to learn.
Looking at the literature overall, the quality of cross-lingual representation tends to improve as representation learning techniques improve. NLP has moved from hand-engineered features with classical machine learning models to word embeddings with deep neural networks. In the past four years, representation learning methods like ELMo---a deep LSTM network trained with language model objective---and BERT---a deep Transformer network trained with masked language model objective---have revolutionized NLP again, including cross-lingual representation learning. In this thesis, we will refer to models like ELMo and BERT as encoders, encoding words in context into contextual vector representation with deep networks.

Around three years ago, in November 2018, a multilingual version of BERT was released, called multilingual BERT (mBERT). As the authors of BERT say ``[...] (they) do not plan to release more single-language models'', they instead train a single BERT model with Wikipedia to serve 104 languages, hence multilingual BERT. The main difference between English monolingual BERT and multilingual BERT is the training data: Wikipedia of English v.s. Wikipedia of 104 languages.
Surprisingly, even without any explicit cross-lingual signal during pretraining, mBERT shows promising zero-shot cross-lingual performance---training the model on one language then directly applying that model to another language---on a natural language inference dataset.

This thesis first fully documents the surprising cross-lingual potential of mBERT on various tasks against prior art via zero-shot cross-lingual transfer, which directly tests its cross-lingual representation. We show that mBERT is not only learning representation for each language but also learning cross-lingual representation. Such surprising cross-lingual effectiveness leads to a set of questions, most importantly how do multilingual encoders learn cross-lingual representations. This thesis attempts to answer these questions, in doing so, to better understand models behaviour and how these models learn cross-lingual representation. With these insights, we are able to identify and improve their cross-lingual representation and cross-lingual transfer with these encoders.

\section{Outline \& Contributions}

The main contribution of this thesis is to understand how multilingual encoders learn cross-lingual representations. In exploring this question, we will analyse the behavior of multilingual models in a variety of settings on high and low resource languages. Our findings will lead us to suggested improvements to these models, the testing of which will allow us to better understand how these models work and what makes them effective.
As we document the surprising cross-lingual effectiveness of these multilingual models, in each chapter, we answer different questions raised by these models. Together, they provide a better understanding of multilingual encoders on cross-lingual transfer, which leads to directions to improve these models for cross-lingual transfer. All chapters are supported by the same codebase \url{https://github.com/shijie-wu/crosslingual-nlp}.

Chapter \ref{chap:background} reviews the progress of representation learning in NLP and discusses its application in cross-lingual transfer. Improvement on representation learning typically leads to better cross-lingual representation. As BERT revolutionizes representation learning and NLP, a multilingual version of BERT called Multilingual BERT (mBERT) is also released.

\textit{Does mBERT learn cross-lingual representation?}
In \autoref{chap:surprising-mbert}, we show that surprisingly mBERT learns cross-lingual representation even without explicit cross-lingual signal, even outperforming previous state-of-the-art cross-lingual word embeddings on zero-shot cross-lingual transfer. Additionally, we probe mBERT and document the model behavior.
This work was published in \citet{wu-dredze-2019-beto}.

\textit{How does mBERT learn cross-lingual representation?}
Chapter \ref{chap:emerging-structure} presents an ablation study on mBERT, teasing apart which modeling decision contributes the most to the learning of cross-lingual representation. We show that sharing transformer parameters is the most important factor. As monolingual BERT of different languages are similar to each other, parameter sharing allows the model to natrually align the representation in a cross-lingual fashion.
This work was published in \citet{conneau-etal-2020-emerging}.

\textit{Are all languages created equal in mBERT?}
In \autoref{chap:low-resource}, we show that mBERT does not learn equally high quality representation for its lower resource languages. Such outcome is not the product of hyperparameter or multilingual joint training but the sample inefficiency of BERT objective, as monolingual BERT of these languages perform even worse and pairing them with similar high resource languages close the performance gap.
This work was published in \citet{wu-dredze-2020-languages}.

\textit{How to inject cross-lingual signals into multilingual encoders?}
Chapter \ref{chap:crosslingual-signal} introduces two approaches for injecting two types of cross-lingual signal into multilingual encoders: bilingual dictionary and bitext. For the former, we create synthetic code-switch corpus for pretraining. For the latter, we ad-hoc explicitly align the encoder representation using a contrastive alignment loss. Both methods show improvement for mBERT or smaller encoders. However, the performance gain is eclipsed by simply scaling up the model size and data size. Additionally, we observe that zero-shot cross-lingual transfer has high variance on the target language, creating challenges for comparing models fairly in the literature.
This work was published in \citet{conneau-etal-2020-emerging} and \citet{wu-dredze-2020-explicit}.

\textit{Why does zero-shot cross-lingual transfer have high variance as shown in \autoref{chap:crosslingual-signal}?}
In \autoref{chap:analysis}, we show that zero-shot cross-lingual transfer is under-specified optimization, causing its high variance on target languages and much lower variance on source language. To improve the performance of zero-shot cross-lingual transfer, addressing the under-specification could produce bigger gain.

\textit{Does data projection constrain zero-shot cross-lingual transfer optimization?}
Chapter \ref{chap:data-projection} proposes using silver target data---created automatically with machine translation based on supervision in source language---to constrain the optimization, and shows adding such constraint improves zero-shot cross-lingual transfer. We also investigate the impact of encoder on the data creation pipeline, and observe that the best setup is task specific.
This work was published in \citet{yarmohammadi-etal-2021-everything}.

Chapter \ref{chap:conclusions} recaps our contributions and discusses future work.

%% file: src/background.tex
In the past decade, representation learning has improved natural language processing (NLP) technology significantly. Representation learning learns dense representation of language using unlabeled corpus, using the corpus itself as a learning signal. As the computational infrastructure scales with the collection of Web corpus, the capability of representation learning keeps scaling \citep{radford2019language}. Every sub-fields within NLP has been revolutionized by representation learning, including cross-lingual transfer. Cross-lingual transfer attempts to transfer knowledge from one language---typically languages with lots of supervision---to another language---typically languages with less supervision. As modern NLP technology is deployed to support more than one language and different languages have different amounts of supervision for tasks of interest, cross-lingual transfer is the bedrock of NLP real world application. In this chapter, we will discuss the progress on representation learning in NLP and its impact on cross-lingual transfer.

\section{Word Embeddings}

Word embeddings encode word to dense vector representation. It had existed as a part of the neural network based NLP model before, such as language model \citep{bengio2003neural}. While global matrix factorization based methods for learning word embeddings have existed for decades, such as latent semantic analysis \citep{deerwester1990indexing} and Brown clusters \citep{brown-etal-1992-class}, online learning based approaches like Word2Vec (skip-gram and CBOW) \citep{mikolov2013efficient,mikolov2013distributed}, Glove \citep{pennington-etal-2014-glove}, and FastText \citep{bojanowski-etal-2017-enriching} pushes representation learning to a prominent position in NLP. Pretrained word embeddings became a standalone step in the pipeline of developing neural NLP systems with it as input to the neural network. \textbf{Pretraining} usually refers to the training procedure of word embeddings and later contextual word embeddings, as it is learning information from the corpus itself instead of from any particular tasks.

The learning of embedding of a word relies on its contextual information, as the distributional hypothesis states that words in similar contexts have similar meanings. Specifically, skip-gram trains a log-bilinear model to predict words within a certain window size using only the center word, while CBOW trains a similar model to predict the center word using a bag of context words. Both skip-gram and CBOW approximate the word prediction softmax loss with noise contrastive estimation and negative sampling. Glove instead trains word embeddings to predict global co-occurrence word statistics. FastText additionally extends Word2Vec by incorporating subword information. Due to efficiency consideration, word embeddings represent each word type with a single fixed-dimensional vector, trained with local co-occurrence signals regardless of order. As deep learning framework and computational infrastructure improves, such limitations would be later addressed by contextual word embeddings.

\section{Contextual Word Embeddings}
Different from word embeddings, contextual word embeddings represents word using its context processed by a deep neural network. There are many attempts on such idea with language model \citep{peters-etal-2017-semi} or machine translation \citep{mccann2017learned} as learning signal, and ELMo \cite{peters-etal-2018-deep} popularized it within the NLP community. ELMo, two deep LSTM \cite{hochreiter1997long} pretrained with right-to-left and left-to-right language modeling objective, produce contextual word embeddings by combining the output of each layer of LSTM with weighted averaging. Additionally, convolution is used to encode character-level information. This contextualized representation outperforms stand-alone word embeddings, e.g. Word2Vec and Glove, with the same task-specific architecture in various downstream tasks, and achieves state-of-the-art performance at the time of publication. Similar to word embeddings, neural network takes static representation from ELMo as input.

Instead of taking the representation from a pretrained model, GPT \cite{radford2018improving} and \newcite{howard-ruder-2018-universal} also fine-tune all the parameters of the pretrained model for a specific task, referred to as \textbf{fine-tuning}. Also, GPT uses a transformer encoder \cite{vaswani2017attention} instead of an LSTM and jointly fine-tunes with the language modeling objective. \newcite{howard-ruder-2018-universal} propose another fine-tuning strategy by using a different learning rate for each layer with learning rate warmup and gradual unfreezing.

\section{BERT}
BERT \citep{devlin-etal-2019-bert} is a deep contextual representation based on a series of transformers trained by a self-supervised objective. One of the main differences between BERT and related work like ELMo and GPT is that BERT is trained by the Cloze task \cite{taylor1953cloze}, also referred to as masked language modeling, instead of right-to-left or left-to-right language modeling. This allows the model to freely encode information from both directions in each layer, contributing to its better performance compared to ELMo and GPT. The goal of the Cloze task is to predict the center missing word based on its context. BERT could be viewed as a deep CBOW, using much deeper representation to encode much larger ordered contextual information. Softmax is used to compute the probability of the missing word based on the contextual representation.

To set up the Cloze for training, the authors propose a heuristic to replace each word with a mask or a random word with the probability of 12\% or 1.5\%, respectively. Additionally, BERT also optimizes a next sentence classification objective. At training time, 50\% of the paired sentences are consecutive sentences while the rest of the sentences are paired randomly. Instead of operating on words, BERT uses a subword vocabulary with WordPiece \cite{wu2016google}, a data-driven approach to break up a word into subwords. Using a subword vocabulary allows BERT to keep a modest vocabulary size, making the softmax prediction practical, and offers a balance between word-based vocabulary and character-level encoding like ELMo.

\subsection{Fine-tuning}
BERT shows state-of-the-art performance at the time of publication by fine-tuning the transformer encoder followed by a simple softmax classification layer on sentence classification tasks, and a sequence of shared softmax classifications for sequence tagging models on tasks like NER. Fine-tuning usually takes 3 to 4 epochs with a relatively small learning rate, for example, 3e-5. Instead of directly fine-tuning BERT on the task of interests, \citet{phang2018sentence} propose intermediate fine-tuning---fine-tuning BERT on data-rich supervised tasks---and show improvement on the final task of interests. Unlike the later GPT-2 and GPT-3, which will be discussed in \autoref{sec:gpt}, BERT typically requires task-specific data fine-tuning to perform said tasks.

\section{Transformer} Since the introduction of transformer, it has taken over NLP. Its popularity could be attributed to two factors: easy to parallelize and easy to model long range context. For a sequence with length $n$, while recurrent-based models like RNN and LSTM have $O(n)$ sequential operation, transformer has $O(1)$ sequential operation in comparison, making it much more parallelizable. Additionally, to connect any two items within a sequence, recurrent-based models need to pass through up to $O(n)$ items in between while transformer directly connect these two items, making it much easier to model long context.

For completeness, we describe the Transformer used by BERT. Let $\xx$, $\yy$ be a sequence of subwords from a sentence pair. A special token \texttt{[CLS]} is prepended to $\xx$ and \texttt{[SEP]} is appended to both $\xx$ and $\yy$. The embedding is obtained by
\begin{align}
\hat{h}^0_i &= E(x_i) + E(i) + E(\mathbbm{1}_\xx)  \\
\hat{h}^0_{j+|\mathbf{x}|} &= E(y_j) + E(j+|\mathbf{x}|) + E(\mathbbm{1}_\yy)  \\
h_\cdot^{0} &= \dropout(\LN(\hat{h}_\cdot^{0})) 
\end{align}
where $E$ is the embedding function and $\LN$ is layer normalization \cite{ba2016layer}. $M$ transformer blocks are followed by the embeddings. In each transformer block,
\begin{align}
h^{i+1}_{\cdot} &= \Skip(\FF, \Skip(\SA, h^i_{\cdot}))  \\
\Skip(f, h) &= \LN(h + \dropout(f(h)))  \\
\FF(h) &= \gelu(h \WW_1^\top + \bb_1) \WW_2^\top + \bb_2 
\end{align}
where $\gelu$ is an element-wise activation function \cite{hendrycks2016bridging}. In practice, $h^i \in \calR^{(|\xx| + |\yy|)\times d_h}$, $\WW_1 \in \calR^{4d_h \times d_h}$, $\bb_1 \in \calR^{4d_h}$, $\WW_2 \in \calR^{d_h \times 4d_h}$, and $\bb_2 \in \calR^{d_h}$. $\SA$ is the multi-heads self-attention function. We show how one new position $\hat{h}_i$ is computed.
\begin{align}
[\cdots,\hat{h}_i,\cdots] &=\SA([h_1, \cdots, h_{|\xx|+|\yy|}]) \\
&= \WW_o \concat(h^1_i, \cdots, h^N_i) + \bb_o
\end{align}
In each attention, referred to as attention head,
\begin{align}
h^j_i &= \sum^{|\xx|+|\yy|}_{k=1} \dropout(\alpha^{(i,j)}_k) \WW^j_V h_k \\
\alpha^{(i,j)}_k &= \frac{\exp \frac {(\WW^j_Q h_i)^\top \WW^j_K h_k} {\sqrt{d_h/N}} }
{\sum^{|\xx|+|\yy|}_{k'=1} \exp \frac {(\WW^j_Q h_i)^\top \WW^j_K h_{k'}} {\sqrt{d_h/N}}}
\end{align}
where $N$ is the number of attention heads, $h^j_i \in \calR^{d_h / N}$, $\WW_o \in \calR^{d_h\times d_h}$, $\bb_o \in \calR^{d_h}$, and $\WW^j_Q,\WW^j_K,\WW^j_V \in \calR^{d_h/N\times d_h}$.

\section{Generative Language Model}\label{sec:gpt}

Generative language model (LM) pretrained with language modeling objective. In this sense, ELMo and GPT are both generative LM. However, generative LM typically also refer to how the model was used. Instead of taking the representation from the model like ELMo or GPT, they instead cast the task of interest as language modeling, e.g. GPT-2 \cite{radford2019language} and GPT-3 \cite{brown2020language}. Any natural language generation tasks fall under this category, and some NLP tasks can be naturally cast as language model with prompt and template. GPT-2 shows pretrained generative LM can performs zero-shot learning on some tasks---no fine-tuning is needed. GPT-3 shows larger generative LM can performs few-shot learning with only context---again no fine-tuning is needed---although typical few-shot learning usually involve fine-tuning. One potential reason for no fine-tuning in GPT-3 could be model size with 175B parameters, making fine-tuning expensive. This thesis does not focus on generative LM as there is no publicly available multilingual generative LM until the recent mT5 \cite{xue-etal-2021-mt5}, a model with both encoder like BERT and decoder like GPT pretrained with span-corruption objective \cite{raffel2020exploring}. However, scaling beyond the current biggest encoder might need generative LM, as we discuss in \autoref{chap:conclusions}.

\section{Cross-lingual Transfer and Multilingual NLP}

Cross-lingual transfer learning is a type of transductive transfer learning with different source and target domain \citep{pan2010survey}. It attempts to transfer knowledge from one language, usually referred to as source language, to another language, usually referred to as target language. It is possible to have more than one source language or more than one target language. \textbf{Few-shot cross-lingual transfer} assumes limited training data in target languages, while \textbf{Zero-shot cross-lingual transfer} typically assumes no task specific supervision on target language. A stricter assumption further eliminates any cross-lingual signal like bilingual dictionary or bitext. A cross-lingual representation space is assumed to perform the cross-lingual transfer, and the quality of the cross-lingual space is essential for cross-lingual transfer, especially zero-shot transfer.
Multilingual NLP attempts to build a single NLP system supporting multiple languages. Cross-lingual representation benefits the development of multilingual NLP, as it alleviates the learning to solve the specific task. This thesis mainly focuses on zero-shot cross-lingual transfer as a proxy for evaluating cross-lingual representation.

\section{Cross-lingual Representation}\label{sec:crosslingual-word-embeddings-basic}
Cross-lingual representation learning follows a similar development trajectory as representation learning in NLP. Before the widespread use of cross-lingual word embeddings, task-specific models assumed coarse-grain representation like part-of-speech tags, in support of a delexicalized parser \cite{zeman-resnik-2008-cross}.

\subsection{Cross-lingual Word Embeddings}
With the progress on word embeddings, \newcite{mikolov2013exploiting} shows that embedding spaces tend to be shaped similarly across different languages. This inspired work in aligning monolingual embeddings. The alignment was done by using a bilingual dictionary to project words that have the same meaning close to each other with linear mapping \cite{mikolov2013exploiting}.
This projection aligns the words outside of the dictionary as well due to the similar shapes of the word embedding spaces. Follow-up efforts only required a very small seed dictionary (e.g., only numbers \cite{artetxe-etal-2017-learning}) or even no dictionary at all \cite{conneau2017word,zhang-etal-2017-adversarial}. 
\newcite{ruder2017survey} surveys methods for learning cross-lingual word embeddings by either joint training or post-training mappings of monolingual embeddings.
Other work has pointed out that word embeddings may not be as isomorphic as thought \cite{sogaard-etal-2018-limitations} especially for distantly related language pairs \cite{patra-etal-2019-bilingual}. \newcite{ormazabal-etal-2019-analyzing} show joint training can lead to more isomorphic word embeddings space.
On top of cross-lingual word embeddings, task-specific neural architectures have been used for tasks like named entity recognition \cite{xie-etal-2018-neural}, part-of-speech tagging \cite{kim-etal-2017-cross} and dependency parsing \cite{ahmad-etal-2019-difficulties}.

\subsection{Cross-lingual Contextual Word Embeddings}
However, cross-lingual word embeddings have similar drawbacks as word embeddings. With the success of ELMo over word embeddings, \newcite{schuster-etal-2019-cross} aligns pretrained ELMo of different languages by learning an orthogonal mapping and shows strong zero-shot and few-shot cross-lingual transfer performance on dependency parsing with 5 Indo-European languages. \newcite{mulcaire-etal-2019-polyglot} trains a single ELMo on distantly related languages and shows mixed results as to the benefit of pretraining.

\section{Multilingual BERT}\label{sec:mbert-basic}
BERT offers a multilingual model, called Multilingual BERT (mBERT), pretrained on concatenated Wikipedia data for 104 languages {\em without any explicit cross-lingual signal}, e.g. pairs of words, sentences or documents linked across languages \cite{multilingualBERTmd}. It follows the same model architecture and training procedure as BERT, except with data from Wikipedia in 104 languages.

In mBERT, the WordPiece modeling strategy allows the model to share embeddings across languages. For example, ``DNA'' has a similar meaning even in distantly related languages like English and Chinese.\footnote{``DNA'' indeed appears in the vocabulary of mBERT as a stand-alone lexicon.} To account for varying sizes of Wikipedia training data in different languages, training uses a heuristic to subsample or oversample words when running WordPiece as well as sampling a training batch, random words for cloze and random sentences for next sentence classification.

However, mBERT does surprisingly well compared to cross-lingual word embeddings on zero-shot cross-lingual transfer in XNLI \cite{conneau-etal-2018-xnli}, a natural language inference dataset \cite{multilingualBERTmd}.
While the XNLI experiment is promising, many questions remain unanswered. What does mBERT really learn: 
separate representation for each language, or some cross-lingual representation mixed with some language-specific representation? Is mBERT better than cross-lingual word embeddings in terms of cross-lingual transfer? How does its modeling decision impact its performance? In \autoref{chap:surprising-mbert}, we will conduct experiments to answer these questions.

%% file: src/surprise.tex
\section{Introduction}

As we discuss in \autoref{sec:mbert-basic}, while XNLI results are promising, the question remains: does mBERT learn a cross-lingual space that supports zero-shot transfer, even without any explicit cross-lingual signal? Does mBERT learn a cross-lingual representation, or does it produce a representation for each language in its own embedding space? 
Is mBERT better than cross-lingual word embeddings in terms of cross-lingual transfer? How does its modeling decision impact its performance?

In this chapter, grounded by models in the literature, we evaluate mBERT as a zero-shot cross-lingual transfer model on five different NLP tasks: natural language inference, document classification, named entity recognition, part-of-speech tagging, and dependency parsing.
We show that it achieves competitive or even state-of-the-art performance (at the time of publication) by simply fine-tuning all parameter of mBERT with minimal task-specific layer. This is surprising as mBERT does not have any explicit cross-lingual signal during pretraining while prior work assume various amount of cross-lingual signal.
While fine-tuning all parameters achieves strong performance, we additionally explore different fine-tuning and feature extraction schemes.
We demonstrate that we could further outperform the suggested fine-tune all approach with simple parameter freezing---freezing the bottom layer of mBERT.
Furthermore, we explore the extent to which mBERT maintains language-specific information by probing each layer of mBERT with language identification. Surprisingly, mBERT maintains strong language specific information despite having strong cross-lingual representation.
Finally, we show how subword tokenization modeling decisions impact cross-lingual transfer performance.
We observe a positive correlation between the amount of subword overlap between languages and the transfer performance across languages.

Parallel to the publication of this chapter, \newcite{lample2019cross} incorporates bitext into BERT by training on pairs of parallel sentences. \newcite{pires-etal-2019-multilingual} shows mBERT has good zero-shot cross-lingual transfer performance on NER and POS tagging. They show how subword overlap and word ordering affect mBERT transfer performance. Additionally, they show mBERT can find translation pairs and works on code-switched POS tagging. In comparison, this chapter looks at a larger set of NLP tasks including dependency parsing and ground the mBERT performance against previous state-of-the-art on zero-shot cross-lingual transfer. We also probe mBERT in different ways and show a more complete picture of the cross-lingual effectiveness of mBERT.

\section{Tasks}\label{sec:tasks}

\input{table/surprise/language}
We consider five tasks in the zero-shot transfer setting. We assume labeled training data for each task in English, and transfer the trained model to a target language. We select a range of different tasks: document classification, natural language inference, named entity recognition, part-of-speech tagging, and dependency parsing. We cover zero-shot transfer from English to 38 languages in the 5 different tasks as shown in \autoref{tab:surprise-lang}. In this section, we describe the tasks as well as task-specific layers.

\subsection{Document Classification}\label{sec:task-mldoc}
We use MLDoc \cite{schwenk-li-2018-corpus}, a balanced subset of the Reuters corpus covering 8 languages for document classification. The 4-way topic classification task decides between CCAT (Corporate/Industrial), ECAT (Economics), GCAT (Government/Social), and MCAT (Markets). We only use the first two sentences\footnote{We only use the first sentence if the document only contains one sentence. Documents are segmented into sentences with NLTK \cite{perkins2014python}.} of a document for classification due to memory constraint. The sentence pairs are provided to the mBERT encoder. The task-specific classification layer is a linear function mapping $h^{12}_0 \in \calR^d_h$ into $\calR^4$, and a softmax is used to get class distribution. We evaluate by classification accuracy.

\subsection{Natural Language Inference}\label{sec:task-xnli}
We use XNLI \cite{conneau-etal-2018-xnli} which covers 15 languages for natural language inference. The 3-way classification includes entailment, neutral, and contradiction given a pair of sentences. We feed a pair of sentences directly into mBERT and the task-specific classification layer is the same as \autoref{sec:task-mldoc}. We evaluate by classification accuracy.

\subsection{Named Entity Recognition}\label{sec:task-ner}

We use the CoNLL 2002 and 2003 NER shared tasks \cite{tjong-kim-sang-2002-introduction,tjong-kim-sang-de-meulder-2003-introduction} (4 languages) and a Chinese NER dataset \cite{levow-2006-third}. The labeling scheme is BIO with 4 types of named entities. We add a linear classification layer with softmax to obtain word-level predictions. Since mBERT operates at the subword-level while the labeling is word-level, if a word is broken into multiple subwords, we mask the prediction of non-first subwords.
NER is evaluated by F1 of predicted entities (F1). Note we adopt a simple post-processing heuristic to obtain a valid span, rewriting standalone \texttt{I-X} into \texttt{B-X} and \texttt{B-X I-Y I-Z} into \texttt{B-Z I-Z I-Z}, following the final entity type.

\subsection{Part-of-Speech Tagging}\label{sec:task-pos}

We use a subset of Universal Dependencies (UD) Treebanks (v1.4) \cite{ud1.4}, which cover 15 languages, following the setup of \newcite{kim-etal-2017-cross}. The task-specific labeling layer is the same as \autoref{sec:task-ner}. POS tagging is evaluated by the accuracy of predicted POS tags (ACC).

\subsection{Dependency parsing}\label{sec:task-parsing}

Following the setup of \newcite{ahmad-etal-2019-difficulties}, we use a subset of Universal Dependencies (UD) Treebanks (v2.2) \cite{ud2.2}, which includes 31 languages. Dependency parsing is evaluated by unlabelled attachment score (UAS) and labeled attachment score (LAS) \footnote{Punctuations (PUNCT) and symbols (SYM) are excluded.}. We only predict the coarse-grain dependency label following \citeauthor{ahmad-etal-2019-difficulties}. We use the model of \newcite{dozat2016deep}, a graph-based parser as a task-specific layer. Their LSTM encoder is replaced by mBERT. Similar to \autoref{sec:task-ner}, we only take the representation of the first subword of each word. We use masking to prevent the parser from operating on non-first subwords.

\section{Experiments}

We use the base cased multilingual BERT, which has $N=12$ attention heads and $M=12$ transformer blocks. The dropout probability is 0.1 and $d_h$ is 768. The model has 179M parameters with about 120k vocabulary.

\subsection{Training} For each task, no preprocessing is performed except tokenization of words into subwords with WordPiece. We use Adam \cite{kingma2014adam} for fine-tuning with $\beta_1$ of 0.9, $\beta_2$ of 0.999 and L2 weight decay of 0.01. We warm up the learning rate over the first 10\% of batches and linearly decay the learning rate.

\subsection{Maximum Subwords Sequence Length} \label{sec:max-seq-len} At training time, we limit the length of subwords sequence to 128 to fit in a single GPU for all tasks. For NER and POS tagging, we additionally use the sliding window approach. After the first window, we keep the last 64 subwords from the previous window as context. In other words, for a non-first window, only (up to) 64 new subwords are added for prediction. At evaluation time, we follow the same approach as training time except for parsing. We threshold the sentence length to 140 words, including words and punctuation, following \newcite{ahmad-etal-2019-difficulties}. In practice, the maximum subwords sequence length is the number of subwords of the first 140 words or 512, whichever is smaller.

\subsection{Hyperparameter Search and Model Selection} We select the best hyperparameters by searching a combination of batch size, learning rate and the number of fine-tuning epochs with the following range: learning rate $\{2\times 10^{-5},3\times 10^{-5},5\times 10^{-5}\}$; batch size $\{16, 32\}$; number of epochs: $\{3, 4\}$. Note the best hyperparameters and models are selected by development performance in \textit{English}.

\section{Is mBERT Multilingual?}\label{sec:is-mbert-multilingual}

\input{table/surprise/mldoc}

\input{table/surprise/xnli}

\input{table/surprise/ner}

\input{table/surprise/pos}

\input{table/surprise/parsing}

\subsection{MLDoc} We include two strong baselines. \citet{schwenk-li-2018-corpus} use MultiCCA, multilingual word embeddings trained with a bilingual dictionary \cite{ammar2016massively}, and convolution neural networks. Concurrent to the publication of this chapter, \citet{artetxe-schwenk-2019-massively} use bitext between English/Spanish and the rest of languages to pretrain a multilingual sentence representation with a sequence-to-sequence model where the decoder only has access to a max-pooling of the encoder hidden states.

mBERT outperforms (\autoref{tab:surprise-mldoc}) multilingual word embeddings and performs comparably with a multilingual sentence representation, even though mBERT does not have access to bitext. Interestingly, mBERT outperforms \citet{artetxe-schwenk-2019-massively} in distantly related languages like Chinese and Russian and under-performs in closely related Indo-European languages.

\subsection{XNLI} We include three strong baselines, \citet{artetxe-schwenk-2019-massively} and \citet{lample2019cross} are concurrent to the publication of this chapter. \citet{lample2019cross} with MLM is similar to mBERT; the main difference is that it only trains with the 15 languages of XNLI, has 249M parameters (around 40\% more than mBERT), and MLM+TLM also uses bitext as training data \footnote{They also use language embeddings as input and exclude the next sentence classification objective}. \citet{conneau-etal-2018-xnli} use supervised multilingual word embeddings with an LSTM encoder and max-pooling. After an English encoder and classifier are trained, the target encoder is trained to mimic the English encoder with ranking loss and bitext.

In \autoref{tab:surprise-xnli}, mBERT outperforms one model with bitext training but (as expected) falls short of models with more cross-lingual training information. Interestingly, mBERT and MLM are mostly the same except for the training languages, yet we observe that mBERT under-performs MLM by a large margin. We hypothesize that limiting pretraining to only those languages needed for the downstream task is beneficial. The gap between \citet{artetxe-schwenk-2019-massively} and mBERT in XNLI is larger than MLDoc, likely because XNLI is harder.

\subsection{NER} We use \newcite{xie-etal-2018-neural} as a zero-shot cross-lingual transfer baseline, which is state-of-the-art on CoNLL 2002 and 2003. It uses unsupervised bilingual word embeddings \cite{conneau2017word} with a hybrid of a character-level/word-level LSTM, self-attention, and a CRF. Pseudo training data is built by word-to-word translation with an induced dictionary from bilingual word embeddings.

mBERT outperforms a strong baseline by an average of 6.9 points absolute F1 and an 11.8 point absolute improvement in German with a simple one layer 0$^\text{th}$-order CRF as a prediction function (\autoref{tab:surprise-ner}). A large gap remains when transferring to distantly related languages (e.g. Chinese) compared to a supervised baseline. Further effort should focus on transferring between distantly related languages. In  \autoref{sec:mbert-corr} we show that sharing subwords across languages helps transfer.

\subsection{POS} We use \newcite{kim-etal-2017-cross} as a reference. They utilized a small amount of supervision in the target language as well as English supervision so the results are not directly comparable. \autoref{tab:surprise-pos} shows a large (average) gap between mBERT and Kim et al. Interestingly, mBERT still outperforms \newcite{kim-etal-2017-cross} with 320 sentences in German (de), Polish (pl), Slovak (sk) and Swedish (sv).

\subsection{Dependency Parsing} We use the best performing model on average in \newcite{ahmad-etal-2019-difficulties} as a zero-shot transfer baseline, i.e. transformer encoder with graph-based parser \cite{dozat2016deep}, and dictionary supervised cross-lingual embeddings \cite{smith2017offline}. Dependency parsers, including \citeauthor{ahmad-etal-2019-difficulties}, assume access to gold POS tags: a cross-lingual representation. We consider two versions of mBERT: with and without gold POS tags. When tags are available, a tag embedding is concatenated with the final output of mBERT.

\autoref{tab:surprise-parsing} shows that mBERT outperforms the baseline on average by 7.3 point UAS and 0.4 point LAS absolute improvement even without gold POS tags. Note in practice, gold POS tags are not always available, especially for low resource languages.
Interestingly, the LAS of mBERT tends to be weaker than the baseline in languages with less word order distance, in other words, more closely related to English.
With the help of gold POS tags, we further observe 1.6 points UAS and 4.7 point LAS absolute improvement on average. It appears that adding gold POS tags, which provide clearer cross-lingual representations, benefit mBERT.

\subsection{Summary} Across all five tasks, mBERT demonstrates strong (sometimes state-of-the-art) zero-shot cross-lingual performance without any cross-lingual signal. It outperforms cross-lingual embeddings in four tasks. With a small amount of target language supervision and cross-lingual signal, mBERT may improve further. In short, mBERT is a surprisingly effective cross-lingual model for many NLP tasks.

\section{Does mBERT Vary Layer-wise?}\label{sec:mbert-layer}

\begin{figure*}[ht]
\centering
\subfloat[][Document classification (ACC)]{
\includegraphics[width=0.4\columnwidth]{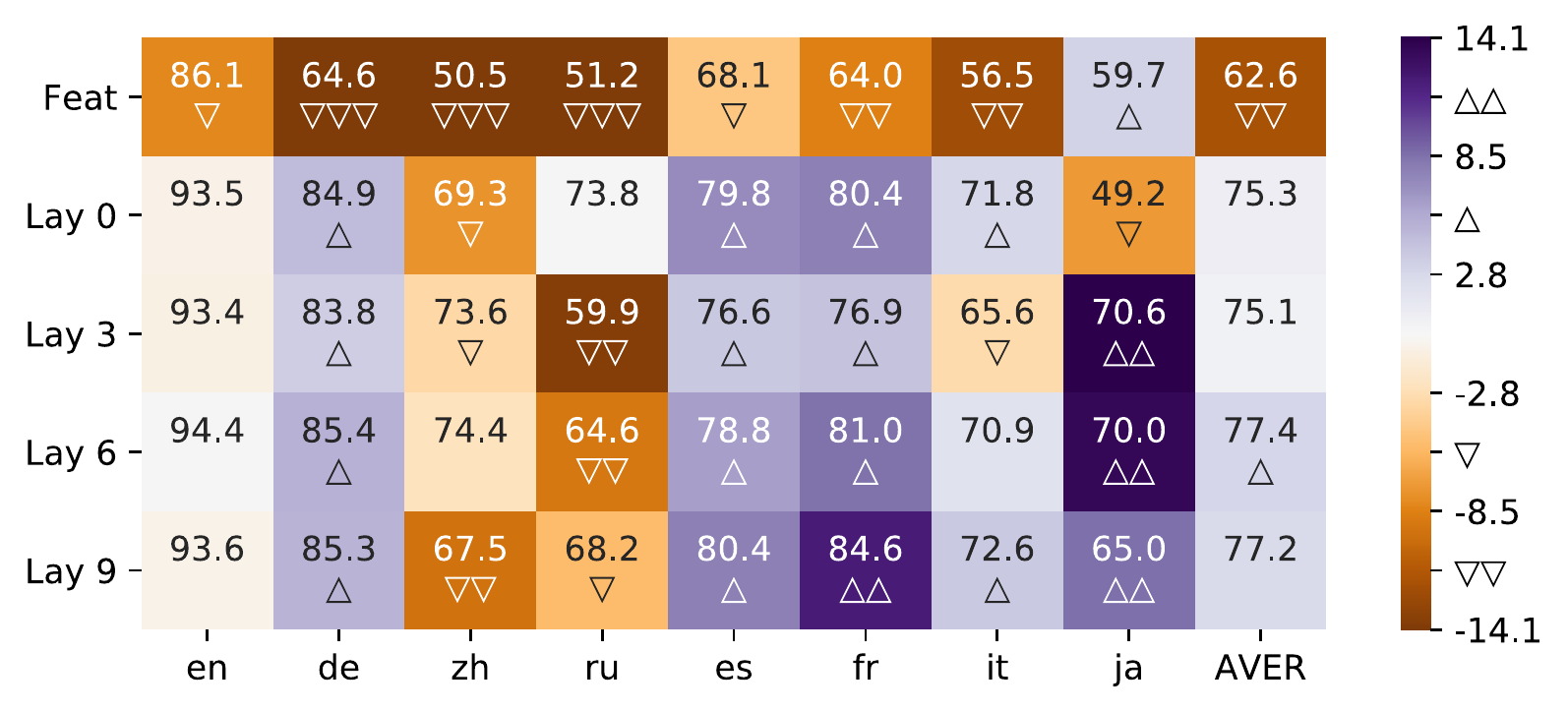}
\label{fig:heatmap-mldoc}}
\subfloat[][Natural language inference (ACC)]{
\includegraphics[width=0.59\columnwidth]{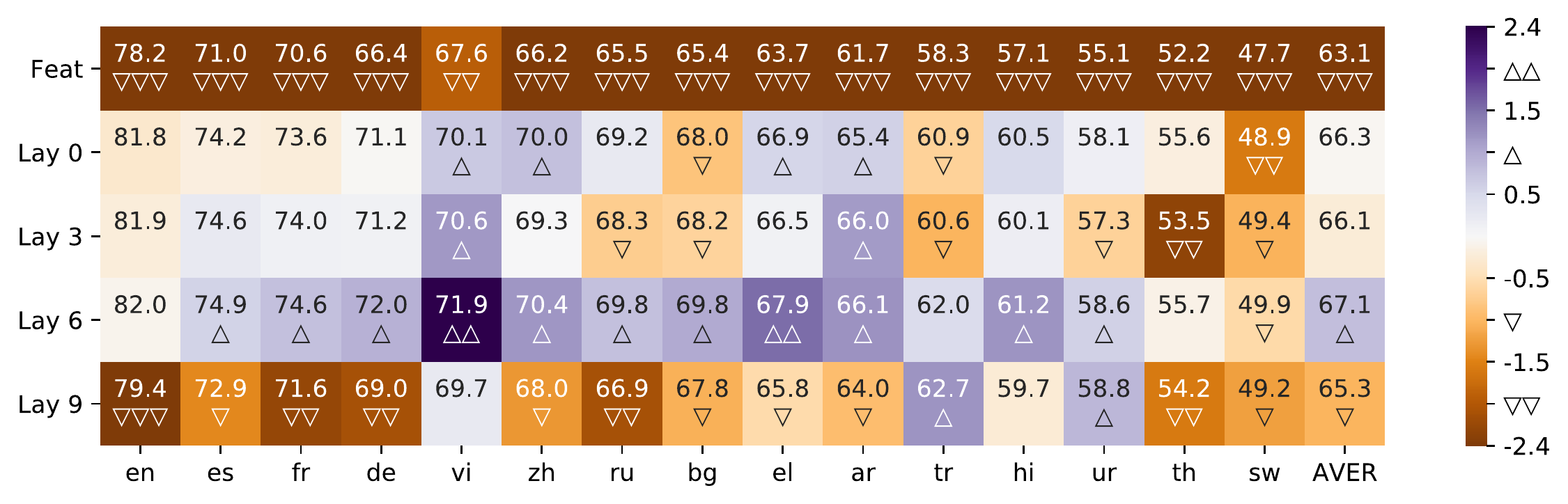}
\label{fig:heatmap-xnli}}
\qquad

\subfloat[][NER (F1)]{
\includegraphics[width=0.33\columnwidth]{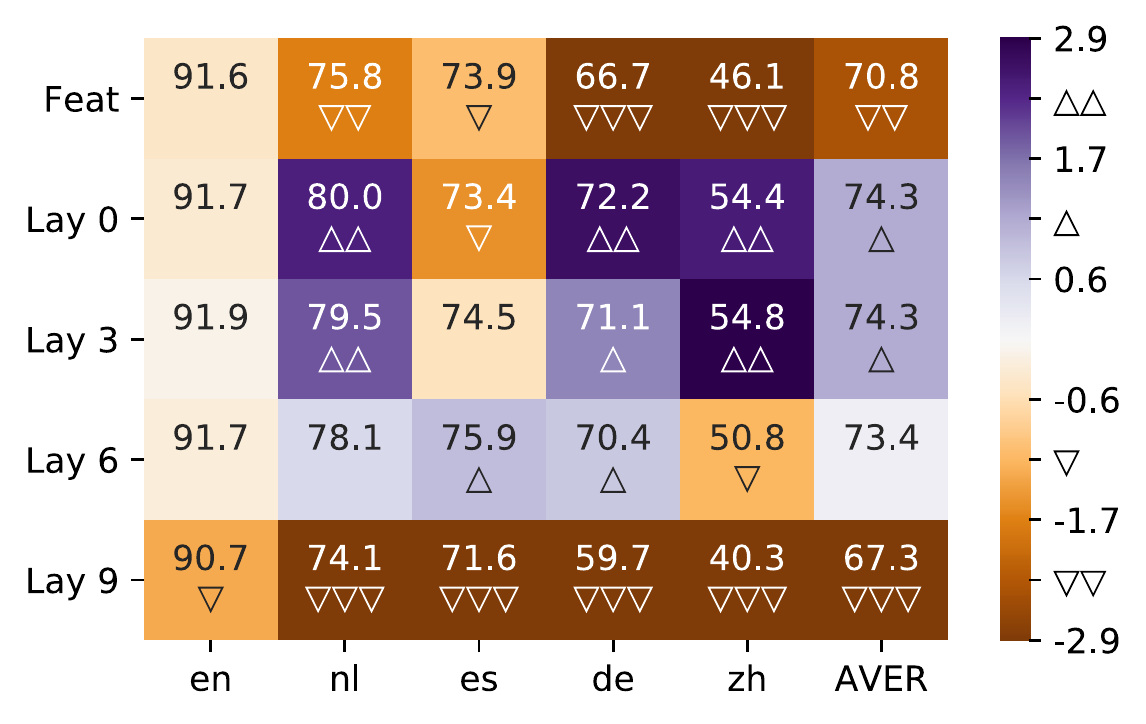}
\label{fig:heatmap-ner}}
\subfloat[][POS tagging (ACC)]{
\includegraphics[width=0.66\columnwidth]{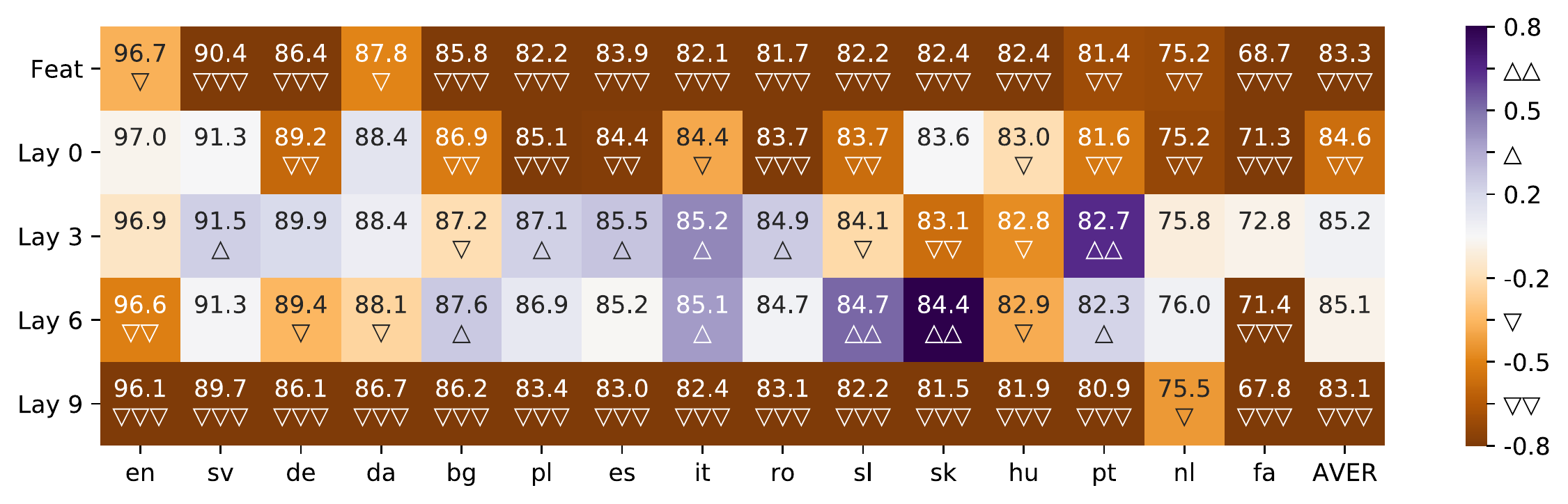}
\label{fig:heatmap-pos}}
\qquad

\subfloat[][Dependency parsing (LAS)]{
\includegraphics[width=1\columnwidth]{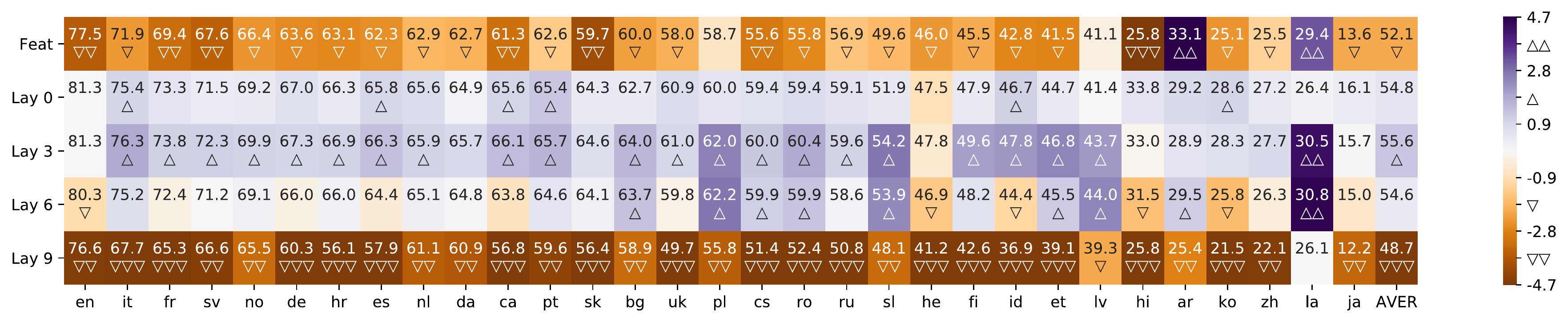}
\label{fig:heatmap-parsing}}

\caption{Performance of different fine-tuning approaches compared with fine-tuning all mBERT parameters. Color denotes absolute difference and the number in each entry is the evaluation in the corresponding setting. Languages are sorted by mBERT zero-shot transfer performance. Three downward triangles indicate performance drop more than the legend's lower limit.
}\label{fig:heatmap}
\end{figure*}

The goal of a deep neural network is to abstract to higher-order representations as you progress up the hierarchy \cite{yosinski2014transferable}. \newcite{peters-etal-2018-deep} empirically show that for ELMo in English the lower layer is better at syntax while the upper layer is better at semantics. However, it is unclear how different layers affect the quality of cross-lingual representation. For mBERT, we hypothesize a similar generalization across the 13 layers, as well as an abstraction away from a specific language with higher layers. Does the zero-shot transfer performance vary with different layers?

We consider two schemes. First, we follow the feature-based approach of ELMo by taking a learned weighted combination of all 13 layers of mBERT with a two-layer bidirectional LSTM with $d_h$ hidden size (Feat). Note the LSTM is trained from scratch and mBERT is fixed. For sentence and document classification, an additional max-pooling is used to extract a fixed-dimension vector. We train the feature-based approach with Adam and learning rate 1e-3. The batch size is 32. The learning rate is halved whenever the development evaluation does not improve. The training is stopped early when learning rate drops below 1e-5. Second, when fine-tuning mBERT, we fix the bottom $n$ layers ($n$ included) of mBERT, where layer 0 is the input embedding. We consider $n \in \{0, 3, 6, 9\}$.

Freezing the bottom layers of mBERT, in general, improves the performance of mBERT in all five tasks (\autoref{fig:heatmap}). For sentence-level tasks like document classification and natural language inference, we observe the largest improvement with $n = 6$. For word-level tasks like NER, POS tagging, and parsing, we observe the largest improvement with $n = 3$. More improvement in under-performing languages is observed.

In each task, the feature-based approach with LSTM under-performs the fine-tuning approach. We hypothesize that
initialization from pretraining with lots of languages provides a very good starting point that is hard to beat. Additionally, the LSTM could also be part of the problem. In \citet{ahmad-etal-2019-difficulties} for dependency parsing, an LSTM encoder was worse than a transformer when transferring to languages with high word ordering distance to English.

\section{Does mBERT Retain Language Specific Information?}\label{sec:mbert-langid}

\begin{figure}[]
\centering
\includegraphics{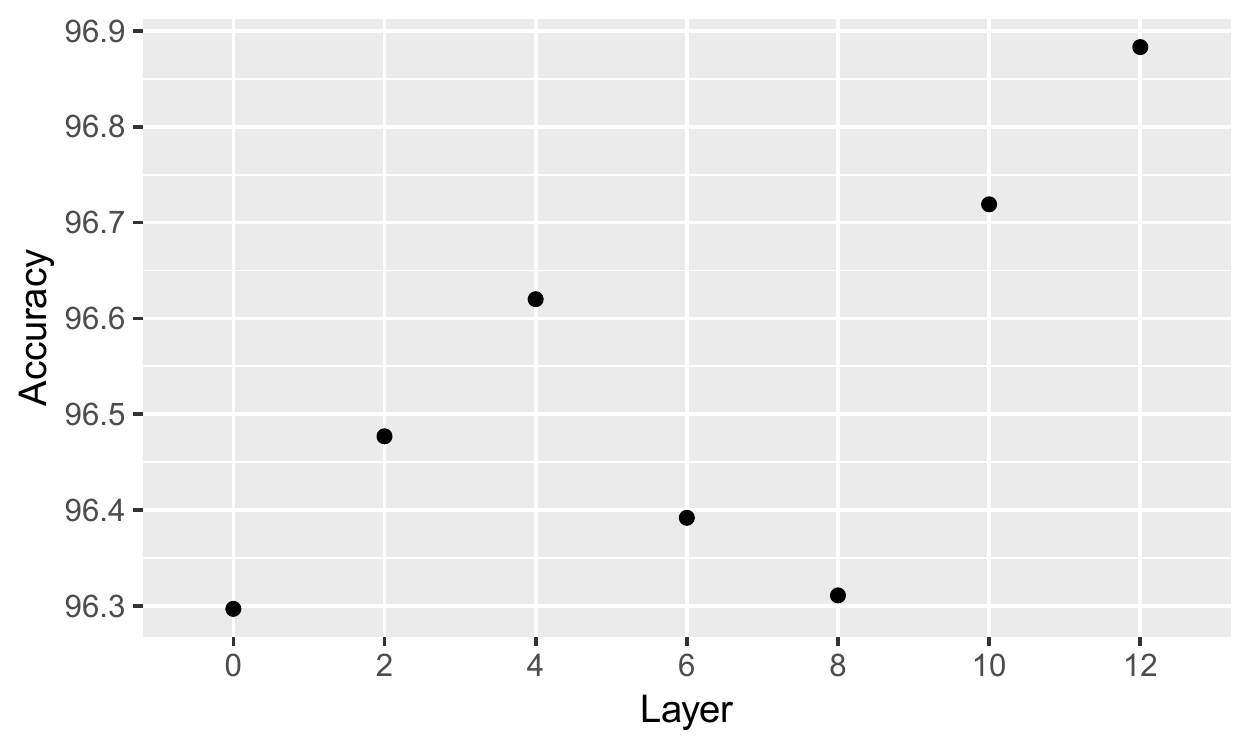}
\caption{Language identification accuracy for different layer of mBERT. layer 0 is the embedding layer and the layer $i > 0$ is the output of the i$^\text{th}$ transformer block.}
\label{fig:langid}
\end{figure}

\begin{figure*}[]
\centering
\includegraphics[width=\columnwidth]{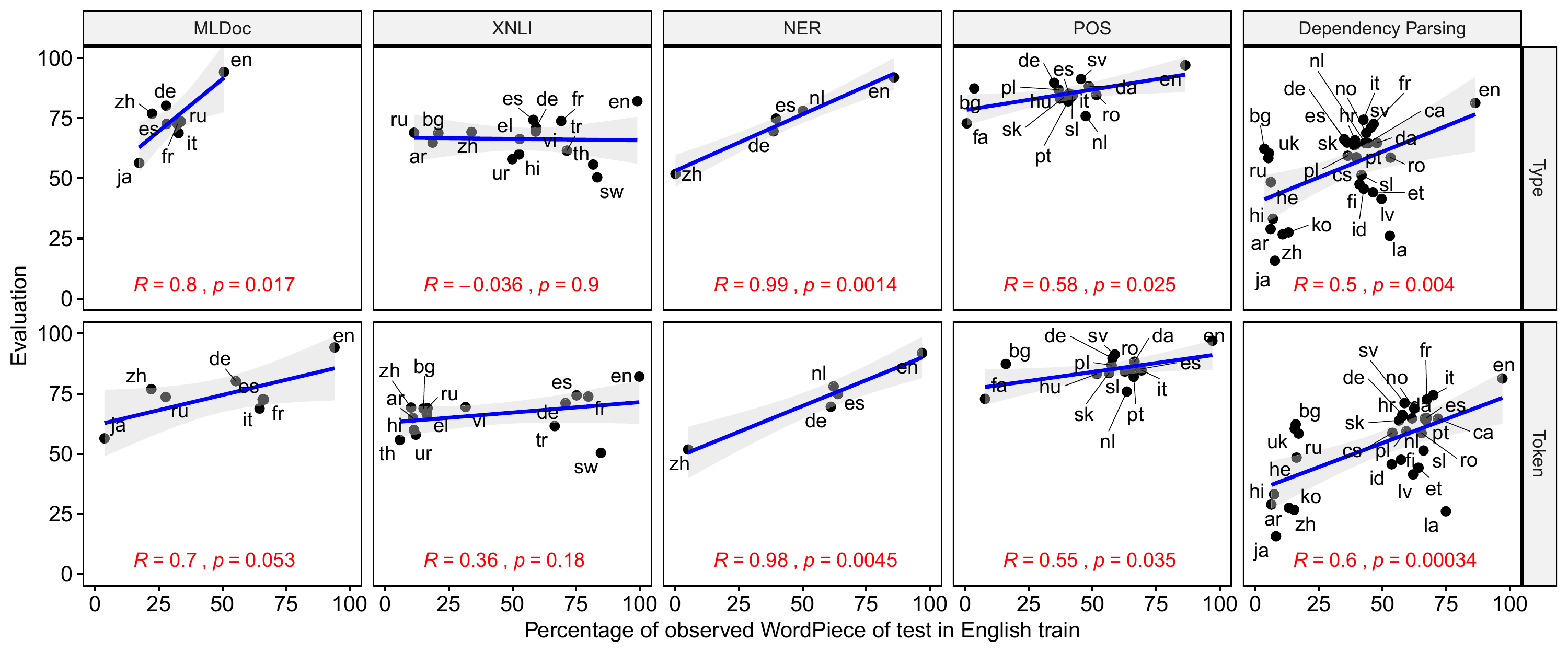}
\caption{Relation between cross-lingual zero-shot transfer performance with mBERT and percentage of observed subwords at both type-level and token-level. Pearson correlation coefficient and $p$-value are shown in red.}\label{fig:corr}
\end{figure*}

mBERT may learn a cross-lingual representation by abstracting away from language-specific information, thus losing the ability to distinguish between languages. We test this by considering language identification: does mBERT retain language-specific information? We use WiLI-2018 \cite{thoma2018wili}, which includes over 200 languages from Wikipedia. We keep only those languages included in mBERT, leaving 99 languages \footnote{Hungarian, Western-Punjabi, Norwegian-Bokmal, and Piedmontese are not covered by WiLI.}.
We take various layers of bag-of-words mBERT representation of the first two sentences of the test paragraph and add a linear classifier with softmax. We fix mBERT and train \textit{only} the classifier the same as the feature-based approach in \autoref{sec:mbert-layer}.

All tested layers achieved around 96\% accuracy (\autoref{fig:langid}), with no clear difference between layers. This suggests each layer contains language-specific information; surprising given the zero-shot cross-lingual abilities. As mBERT generalizes its representations and creates cross-lingual representations, it maintains language-specific details. This may be encouraged during pretraining since mBERT needs to retain enough language-specific information to perform the cloze task.

\section{Does mBERT Benefit by Sharing Subwords Across Languages?}\label{sec:mbert-corr}

As discussed in \autoref{sec:mbert-basic}, mBERT shares subwords in closely related languages or perhaps in distantly related languages. At training time, the representation of a shared subword is explicitly trained to contain enough information for the cloze task in all languages in which it appears. During fine-tuning for zero-shot cross-lingual transfer, if a subword in the target language test set also appears in the source language training data, the supervision could be leaked to the target language explicitly. However, all subwords interact in a non-interpretable way inside a deep network, it is hard to characterize how sharing subwords affects the transfer performance. Additionally, subword representations could overfit to the source language and potentially hurt transfer performance. In these experiments, we investigate how sharing subwords across languages affects cross-lingual transfer.

To quantify how many subwords are shared across languages in any task, we assume $V^\text{en}_{\text{train}}$ is the set of all subwords in the English training set, $V^\ell_{\text{test}}$ is the set of all subwords in language $\ell$ test set, and $c^\ell_w$ is the count of subword $w$ in test set of language $\ell$. We then calculate the percentage of observed subwords at type-level $\ptype^\ell$ and token-level $\ptoken^\ell$ for each target language $\ell$.
\begin{align}
\ptype^\ell &= \frac{|V^\ell_\text{obs}|}{|V^{\ell}_\text{test}|} \times 100 \\
\ptoken^\ell &= \frac{ \sum_{w\in V^\ell_\text{obs}} c^\ell_w } { \sum_{w\in V^\ell_\text{test}} c^\ell_w } \times 100
\end{align}
where $V^\ell_\text{obs} = V^\text{en}_{\text{train}} \cap V^\ell_{\text{test}}$.

In \autoref{fig:corr}, we show the relation between cross-lingual zero-shot transfer performance of mBERT and $\ptype^\ell$ or $\ptoken^\ell$ for all five tasks with Pearson correlation. In four out of five tasks (not XNLI) we observed a strong positive correlation  ($p<0.05$) with a correlation coefficient larger than 0.5. In Indo-European languages, we observed $\ptoken^\ell$ is usually around 50\% to 75\% while $\ptype^\ell$ is usually less than 50\%. This indicates that subwords shared across languages are usually high frequency\footnote{With the data-dependent WordPiece algorithm, subwords that appear in multiple languages with high frequency are more likely to be selected.}.

\section{Discussion}
In this chapter, we show mBERT does well in a cross-lingual zero-shot transfer setting on five different tasks covering a large number of languages, even without any explicit cross-lingual signal during pretraining. It outperforms cross-lingual embeddings, which typically have more cross-lingual supervision. By fixing the bottom layers of mBERT during fine-tuning, we observe further performance gains. Language-specific information is preserved in all layers. Sharing subwords helps cross-lingual transfer; a strong correlation is observed between the percentage of overlapping subwords and transfer performance.
mBERT effectively learns a good multilingual representation with strong cross-lingual zero-shot transfer performance in various tasks. Even without explicit cross-lingual supervision, these models do very well.

This thesis builds on top of these findings. 
In \autoref{sec:mbert-corr}, we observe a correlation between subword overlap between languages and cross-lingual transfer performance. However, this is surprisingly not causation despite being intuitive, as we will show in \autoref{chap:emerging-structure}, determining which factor contributes the most to the learning of cross-lingual representation.
While we experimented with 39 languages in this chapter, the majority of languages supported by mBERT are still untested. In \autoref{chap:low-resource}, we test the low resource languages within mBERT.
As we show with XNLI in \autoref{sec:is-mbert-multilingual}, while bitext is hard to obtain in low resource settings, a variant of mBERT pretrained with bitext \cite{lample2019cross} shows even stronger performance. In \autoref{chap:crosslingual-signal}, we explore how to introduce cross-lingual supervision into models like BERT.
With POS tagging in \autoref{sec:is-mbert-multilingual}, we show mBERT, in general, under-performs models with a small amount of supervision. \newcite{lauscher-etal-2020-zero} shows few-shot cross-lingual transfer improves zero-shot cross-lingual transfer, although the choice of shot has significant impact on the performance \cite{zhao-etal-2021-closer}. Such observation is not surprising, as we will take a deeper dive in \autoref{chap:analysis} looking at why zero-shot cross-lingual transfer has high variance.
In \autoref{chap:data-projection}, we explore how to construct better data projection pipeline to improve zero-shot cross-lingual transfer with multilingual models like mBERT.

Outside of this thesis, many papers build on top of these findings. By scaling up mBERT with bigger dataset and model, better cross-lingual representation can be achieved, including models like XLM-R \cite{conneau-etal-2020-unsupervised}, mT5 \cite{xue-etal-2021-mt5}, and XLM-R$_\text{XXL}$ \cite{goyal-etal-2021-larger}. With strong cross-lingual representation covering over 100 languages, mBERT enables massively multilingual models like multilingual parser UDify \cite{kondratyuk-straka-2019-75}. As more and more multilingual models become available, benchmarks have been introduced aggregating existing multilingual dataset \cite{hu2020xtreme,liang2020xglue}.

%% file: table/surprise/language.tex
\begin{table*}[th]
\centering
\begin{adjustbox}{width=\columnwidth}
\begin{tabular}{ccccc ccccc ccccc ccccc ccccc ccccc ccccc ccccc}
\toprule
 & ar & bg & ca & cs & da & de & el & en & es & et & fa & fi & fr & he & hi & hr & hu & id & it & ja & ko & la & lv & nl & no & pl & pt & ro & ru & sk & sl & sv & sw & th & tr & uk & ur & vi & zh \\
\midrule
MLDoc &  &  &  &  &  & $\checkmark$ &  & $\checkmark$ & $\checkmark$ &  &  &  & $\checkmark$ &  &  &  &  &  & $\checkmark$ & $\checkmark$ &  &  &  &  &  &  &  &  & $\checkmark$ &  &  &  &  &  &  &  &  &  & $\checkmark$ \\
NLI & $\checkmark$ & $\checkmark$ &  &  &  & $\checkmark$ & $\checkmark$ & $\checkmark$ & $\checkmark$ &  &  &  & $\checkmark$ &  & $\checkmark$ &  &  &  &  &  &  &  &  &  &  &  &  &  & $\checkmark$ &  &  &  & $\checkmark$ & $\checkmark$ & $\checkmark$ &  & $\checkmark$ & $\checkmark$ & $\checkmark$ \\
NER &  &  &  &  &  & $\checkmark$ &  & $\checkmark$ & $\checkmark$ &  &  &  &  &  &  &  &  &  &  &  &  &  &  & $\checkmark$ &  &  &  &  &  &  &  &  &  &  &  &  &  &  & $\checkmark$ \\
POS &  & $\checkmark$ &  &  & $\checkmark$ & $\checkmark$ &  & $\checkmark$ & $\checkmark$ &  & $\checkmark$ &  &  &  &  &  & $\checkmark$ &  & $\checkmark$ &  &  &  &  & $\checkmark$ &  & $\checkmark$ & $\checkmark$ & $\checkmark$ &  & $\checkmark$ & $\checkmark$ & $\checkmark$ &  &  &  &  &  &  &  \\
Parsing & $\checkmark$ & $\checkmark$ & $\checkmark$ & $\checkmark$ & $\checkmark$ & $\checkmark$ &  & $\checkmark$ & $\checkmark$ & $\checkmark$ &  & $\checkmark$ & $\checkmark$ & $\checkmark$ & $\checkmark$ & $\checkmark$ &  & $\checkmark$ & $\checkmark$ & $\checkmark$ & $\checkmark$ & $\checkmark$ & $\checkmark$ & $\checkmark$ & $\checkmark$ & $\checkmark$ & $\checkmark$ & $\checkmark$ & $\checkmark$ & $\checkmark$ & $\checkmark$ & $\checkmark$ &  &  &  & $\checkmark$ &  &  & $\checkmark$ \\
\bottomrule

\end{tabular}
\end{adjustbox}
\caption{The 39 languages used in the 5 tasks.}
\label{tab:surprise-lang}
\end{table*}

%% file: table/surprise/mldoc.tex
\begin{table*}[]
\begin{center}
\resizebox{\linewidth}{!}{
\begin{tabular}[b]{l|ccccc ccc|c}
\toprule
& en & de & zh & es & fr & it & ja & ru & Average \\
\midrule
\multicolumn{10}{l}{\it In language supervised learning} \\
\midrule
\citet{schwenk-li-2018-corpus} & 92.2 & 93.7 & 87.3 & 94.5 & 92.1 & 85.6 & 85.4 & 85.7 & 89.5 \\
mBERT & 94.2 & 93.3 & 89.3 & 95.7 & 93.4 & 88.0 & 88.4 & 87.5 & 91.2 \\
\midrule
\multicolumn{10}{l}{\it Zero-shot cross-lingual transfer} \\
\midrule
\citet{schwenk-li-2018-corpus} & \underline{92.2} & \underline{81.2} & \underline{74.7} & 72.5 & 72.4 & \textbf{69.4} & \textbf{67.6} & 60.8 & 73.9 \\
\citet{artetxe-schwenk-2019-massively} \bitext \concurrent & 89.9 & \textbf{84.8} & 71.9 & \textbf{77.3} & \textbf{78.0} & \textbf{69.4} & \underline{60.3} & \underline{67.8} & \textbf{74.9} \\
mBERT & \textbf{94.2} & 80.2 & \textbf{76.9} & \underline{72.6} & \underline{72.6} & \underline{68.9} & 56.5 & \textbf{73.7} & \underline{74.5} \\
\bottomrule
\end{tabular}
}
\caption{MLDoc experiments. \bitext denotes the model is pretrained with bitext, and \concurrent denotes concurrent work. Bold and underline denote best and second best.}
\label{tab:surprise-mldoc}
\end{center}
\end{table*}

%% file: table/surprise/xnli.tex
\begin{table*}[]
\begin{center}
\resizebox{1\linewidth}{!}{
\begin{tabular}[b]{l|ccccccccccccccc|c}
\toprule
& en & fr & es & de & el & bg & ru & tr & ar & vi & th & zh & hi & sw & ur & Average \\
\midrule
\multicolumn{16}{l}{\it Pseudo supervision with machine translated training data from English to target language} \\
\midrule
\citet{lample2019cross} (MLM+TLM) \bitext \concurrent & 85.0 & 80.2 & 80.8 & 80.3 & 78.1 & 79.3 & 78.1 & 74.7 & 76.5 & 76.6 & 75.5 & 78.6 & 72.3 & 70.9 & 63.2 & 76.7 \\
mBERT & 82.1 & 76.9 & 78.5 & 74.8 & 72.1 & 75.4 & 74.3 & 70.6 & 70.8 & 67.8 & 63.2 & 76.2 & 65.3 & 65.3 & 60.6 & 71.6 \\
\midrule
\multicolumn{16}{l}{\it Zero-shot cross-lingual transfer} \\ 
\midrule
\citet{conneau-etal-2018-xnli} (X-LSTM) \bitext \devselect & 73.7 & 67.7 & 68.7 & 67.7 & 68.9 & 67.9 & 65.4 & 64.2 & 64.8 & 66.4 & 64.1 & 65.8 & 64.1 & 55.7 & 58.4 & 65.6 \\
\citet{artetxe-schwenk-2019-massively} \bitext \concurrent & 73.9 & 71.9 & 72.9 & 72.6 & 73.1 & 74.2 & 71.5 & 69.7 & 71.4 & 72.0 & 69.2 & 71.4 & 65.5 & 62.2 & 61.0 & 70.2 \\
\citet{lample2019cross} (MLM+TLM) \bitext \devselect \concurrent & 85.0 & 78.7 & 78.9 & 77.8 & 76.6 & 77.4 & 75.3 & 72.5 & 73.1 & 76.1 & 73.2 & 76.5 & 69.6 & 68.4 & 67.3 & 75.1 \\
\midrule
\citet{lample2019cross} (MLM) \devselect \concurrent & 83.2 & 76.5 & 76.3 & 74.2 & 73.1 & 74.0 & 73.1 & 67.8 & 68.5 & 71.2 & 69.2 & 71.9 & 65.7 & 64.6 & 63.4 & 71.5 \\
mBERT & 82.1 & 73.8 & 74.3 & 71.1 & 66.4 & 68.9 & 69.0 & 61.6 & 64.9 & 69.5 & 55.8 & 69.3 & 60.0 & 50.4 & 58.0 & 66.3 \\
\bottomrule
\end{tabular}
}
\caption{XNLI experiments. \bitext denotes the model is pretrained with cross-lingual signal including bitext or bilingual dictionary, \concurrent denotes concurrent work, and \devselect denotes model selection with target language dev set.}
\label{tab:surprise-xnli}
\end{center}
\end{table*}

%% file: table/surprise/ner.tex
\begin{table}[]
\begin{center}
\begin{tabular}[b]{l|ccccc|c}
\toprule
& en & nl & es & de & zh & Average (-en,-zh) \\
\midrule
\multicolumn{7}{l}{\it In language supervised learning} \\
\midrule
\citet{xie-etal-2018-neural} & - & 86.40 & 86.26 & 78.16 & - & 83.61 \\
mBERT & 91.97 & 90.94 & 87.38 & 82.82 & 93.17 & 87.05 \\
\midrule
\multicolumn{7}{l}{\it Zero-shot cross-lingual transfer} \\
\midrule
\citet{xie-etal-2018-neural} \devselect & - & 71.25 & 72.37 & 57.76 & - & 67.13 \\
mBERT & 91.97 & \textbf{77.57} & \textbf{74.96} & \textbf{69.56} & 51.90 & \textbf{74.03} \\
\bottomrule
\end{tabular}
\caption{NER tagging experiments. \devselect denotes model selection with target language dev set.}
\label{tab:surprise-ner}
\end{center}
\end{table}

%% file: table/surprise/pos.tex
\begin{table*}[]
\begin{center}
\resizebox{\linewidth}{!}{
\begin{tabular}[b]{l|ccccc ccccc ccccc|c}
\toprule
lang & bg & da & de & en & es & fa & hu & it & nl & pl & pt & ro & sk & sl & sv & Average (-en) \\
\midrule
\multicolumn{10}{l}{\it In language supervised learning} \\
\midrule
mBERT & 99.0 & 97.9 & 95.2 & 97.1 & 97.1 & 97.8 & 96.9 & 98.7 & 92.1 & 98.5 & 98.3 & 97.8 & 97.0 & 98.9 & 98.4 & 97.4 \\
\midrule
\multicolumn{10}{l}{\it Low resource cross-lingual transfer} \\
\midrule
\citet{kim-etal-2017-cross} (1280) & 95.7 & 94.3 & 90.7 & - & 93.4 & 94.8 & 94.5 & 95.9 & 85.8 & 92.1 & 95.5 & 94.2 & 90.0 & 94.1 & 94.6 & 93.3 \\
\citet{kim-etal-2017-cross} (320) & 92.4 & 90.8 & 89.7 & - & 90.9 & 91.8 & 90.7 & 94.0 & 82.2 & 85.5 & 94.2 & 91.4 & 83.2 & 90.6 & 90.7 & 89.9 \\
\midrule
\multicolumn{10}{l}{\it Zero-shot cross-lingual transfer} \\
\midrule
mBERT & 87.4 & 88.3 & 89.8 & 97.1 & 85.2 & 72.8 & 83.2 & 84.7 & 75.9 & 86.9 & 82.1 & 84.7 & 83.6 & 84.2 & 91.3 & 84.3 \\

\bottomrule
\end{tabular}
}
\caption{POS tagging. \citet{kim-etal-2017-cross} use small amounts of training data in the target language.}
\label{tab:surprise-pos}
\end{center}
\end{table*}

%% file: table/surprise/parsing.tex
\begin{table}[]
\begin{center}
\begin{tabular}[b]{l ccccc}
\toprule
 & Dist & mBERT(S) & Baseline(Z) & mBERT(Z) & mBERT(Z+POS) \\
\midrule
en & 0.00 & 91.5/81.3 & 90.4/\textbf{88.4} & \underline{91.5}/81.3 & \textbf{91.8}/\underline{82.2} \\
no & 0.06 & 93.6/85.9 & \underline{80.8}/\textbf{72.8} & 80.6/68.9 & \textbf{82.7}/\underline{72.1} \\
sv & 0.07 & 91.2/83.1 & 81.0/\underline{73.2} & \underline{82.5}/71.2 & \textbf{84.3}/\textbf{73.7} \\
fr & 0.09 & 91.7/85.4 & 77.9/\underline{72.8} & \underline{82.7}/72.7 & \textbf{83.8}/\textbf{76.2} \\
pt & 0.09 & 93.2/87.2 & 76.6/\textbf{67.8} & \underline{77.1}/64.0 & \textbf{78.3}/\underline{66.9} \\
da & 0.10 & 89.5/81.9 & 76.6/\underline{67.9} & \underline{77.4}/64.7 & \textbf{79.3}/\textbf{68.1} \\
es & 0.12 & 92.3/86.5 & 74.5/\underline{66.4} & \underline{78.1}/64.9 & \textbf{79.0}/\textbf{68.9} \\
it & 0.12 & 94.8/88.7 & 80.8/\underline{75.8} & \underline{84.6}/74.4 & \textbf{86.0}/\textbf{77.8} \\
ca & 0.13 & 94.3/89.5 & 73.8/\underline{65.1} & \underline{78.1}/64.6 & \textbf{79.0}/\textbf{67.9} \\
hr & 0.13 & 92.4/83.8 & 61.9/52.9 & \textbf{80.7}/\underline{65.8} & \underline{80.4}/\textbf{68.2} \\
pl & 0.13 & 94.7/79.9 & 74.6/\underline{62.2} & \underline{82.8}/59.4 & \textbf{85.7}/\textbf{65.4} \\
sl & 0.13 & 88.0/77.8 & 68.2/\underline{56.5} & \underline{72.6}/51.4 & \textbf{75.9}/\textbf{59.2} \\
uk & 0.13 & 90.6/83.4 & 60.1/52.3 & \textbf{76.7}/\underline{60.0} & \underline{76.5}/\textbf{65.5} \\
bg & 0.14 & 95.2/85.5 & 79.4/\textbf{68.2} & \underline{83.3}/62.3 & \textbf{84.4}/\underline{68.1} \\
cs & 0.14 & 94.2/86.6 & 63.1/53.8 & \underline{76.6}/\underline{58.7} & \textbf{77.4}/\textbf{63.6} \\
de & 0.14 & 86.1/76.5 & 71.3/61.6 & \underline{80.4}/\underline{66.3} & \textbf{83.5}/\textbf{71.2} \\
he & 0.14 & 91.9/83.6 & 55.3/48.0 & \textbf{67.5}/\underline{48.4} & \underline{67.0}/\textbf{54.3} \\
nl & 0.14 & 94.0/85.0 & 68.6/60.3 & \underline{78.0}/\underline{64.8} & \textbf{79.9}/\textbf{67.1} \\
ru & 0.14 & 94.7/88.0 & 60.6/51.6 & \textbf{73.6}/\underline{58.5} & \underline{73.2}/\textbf{61.5} \\
ro & 0.15 & 92.2/83.2 & 65.1/54.1 & \textbf{77.0}/\underline{58.5} & \underline{76.9}/\textbf{62.6} \\
id & 0.17 & 86.3/75.4 & 49.2/43.5 & \textbf{62.6}/\underline{45.6} & \underline{59.8}/\textbf{48.6} \\
sk & 0.17 & 93.8/83.3 & 66.7/58.2 & \underline{82.7}/\underline{63.9} & \textbf{82.9}/\textbf{67.8} \\
lv & 0.18 & 87.3/75.3 & \textbf{70.8}/\textbf{49.3} & 66.0/41.4 & \underline{70.4}/\underline{48.5} \\
et & 0.20 & 88.8/79.7 & 65.7/\underline{44.9} & \underline{66.9}/44.3 & \textbf{70.8}/\textbf{50.7} \\
fi & 0.20 & 91.3/81.8 & 66.3/\underline{48.7} & \underline{68.4}/47.5 & \textbf{71.4}/\textbf{52.5} \\
zh* & 0.23 & 88.3/81.2 & 42.5/25.1 & \textbf{53.8}/\underline{26.8} & \underline{53.4}/\textbf{29.0} \\
ar & 0.26 & 87.6/80.6 & 38.1/28.0 & \underline{43.9}/\underline{28.3} & \textbf{44.7}/\textbf{32.9} \\
la & 0.28 & 85.2/73.1 & \underline{48.0}/\textbf{35.2} & 47.9/26.1 & \textbf{50.9}/\underline{32.2} \\
ko & 0.33 & 86.0/74.8 & 34.5/16.4 & \textbf{52.7}/\underline{27.5} & \underline{52.3}/\textbf{29.4} \\
hi & 0.40 & 94.8/86.7 & 35.5/26.5 & \underline{49.8}/\underline{33.2} & \textbf{58.9}/\textbf{44.0} \\
ja* & 0.49 & 94.2/87.4 & 28.2/\underline{20.9} & \underline{36.6}/15.7 & \textbf{41.3}/\textbf{30.9} \\
\midrule
AVER & 0.17 & 91.3/82.6 & 64.1/53.8 & \underline{71.4}/\underline{54.2} & \textbf{73.0}/\textbf{58.9} \\
\bottomrule
\end{tabular}
\caption{Dependency parsing results by language (UAS/LAS). * denotes delexicalized parsing in the baseline. S and Z denotes supervised learning and zero-shot transfer. Bold and underline denotes best and second best. We order the languages by word order distance to English.}
\label{tab:surprise-parsing}
\end{center}
\end{table}

%% file: src/emerging.tex
\section{Introduction}
In \autoref{chap:surprising-mbert}, we observe that multilingual language models such as mBERT \cite{devlin-etal-2019-bert} and XLM \cite{lample2019cross} surprisingly enable effective cross-lingual transfer---it is possible to learn a model from supervised data in one language and apply it to another with no additional training---for a wide range of tasks without any explicit cross-lingual signal.
\autoref{chap:surprising-mbert} observes that there is a positive correlation between vocabulary overlap between languages and transfer performance across languages. However, without controlled experiment, it is unclear whether such correlation is causal. More broadly, it is unclear why models like mBERT learn cross-lingual representation without any explicit cross-lingual signal.

In this chapter, we first present a detailed ablation study on the impact of each modeling decision of pretraining on the learning of cross-lingual representation. We look at factors including domain similarity of pretraining corpus, a single shared subword vocabulary across languages, vocabulary overlap between languages, random word replacement during pretraining, joint softmax prediction across languages, and transformer parameter sharing across languages.

Much to our surprise, we discover that pretrained models still learn cross-lingual representation without any shared vocabulary or domain similarity, and even when only a subset of the parameters in the joint encoder are shared.
In particular, by systematically varying the amount of shared vocabulary between two languages during pretraining, we show that the amount of overlap only accounts for a few points of performance in transfer tasks, much less than might be expected. 
By sharing transformer parameters alone, pretraining learns to map similar words and sentences to similar hidden representations. 

How does sharing transformer parameters alone allow a model to learn cross-lingual representation?
To better understand these observation, we also analyze multiple monolingual BERT models trained independently. We find that monolingual models trained in different languages learn representations that align with each other surprisingly well, even though they have no shared parameters during pretraining and completely different vocabulary. This result closely mirrors the widely observed fact that word embeddings can be effectively aligned across languages~\cite{mikolov2013exploiting}. Similar dynamics are at play in multilingual pretraining. As monolingual BERTs of different language are similar to each other, when the transformer parameters are shared across languages during pretraining, the multilingual model naturally align the representation of different language in a cross-lingual fashion.

\section{Background}

\subsection{Alignment of Embeddings}
In \autoref{sec:crosslingual-word-embeddings-basic}, we discuss the alignment of monolingual word embeddings and ELMo to produce cross-lingual representation.
\newcite{wang-etal-2019-cross} align mBERT representations and evaluate on dependency parsing.

\subsection{Neural Network Activation Similarity}
We hypothesize that similar to word embedding spaces, language-universal structures emerge in pretrained language models. While computing word embedding similarity is relatively straightforward, the same cannot be said for the deep contextualized BERT models that we study.
Researchers introduce ways to measure the similarity of neural network activation between different layers and different
models \cite{laakso2000content,li2016convergent,raghu2017svcca,morcos2018insights,wang2018towards}. For example, \newcite{raghu2017svcca} use canonical correlation analysis (CCA) and a new method, singular vector canonical correlation analysis (SVCCA), to show that early layers converge faster than upper layers in convolutional neural networks. \newcite{kudugunta-etal-2019-investigating} use SVCCA to investigate the multilingual representations obtained by the encoder of a massively multilingual neural machine translation system \cite{aharoni-etal-2019-massively}. \newcite{kornblith2019similarity} argue that CCA fails to measure meaningful similarities between representations that have a higher dimension than the number of data points and introduce the centered kernel alignment (CKA) to solve this problem. They successfully use CKA to identify correspondences between activations in networks trained from different initializations.

\begin{figure*}[]
\centering
\includegraphics[width=0.6\columnwidth]{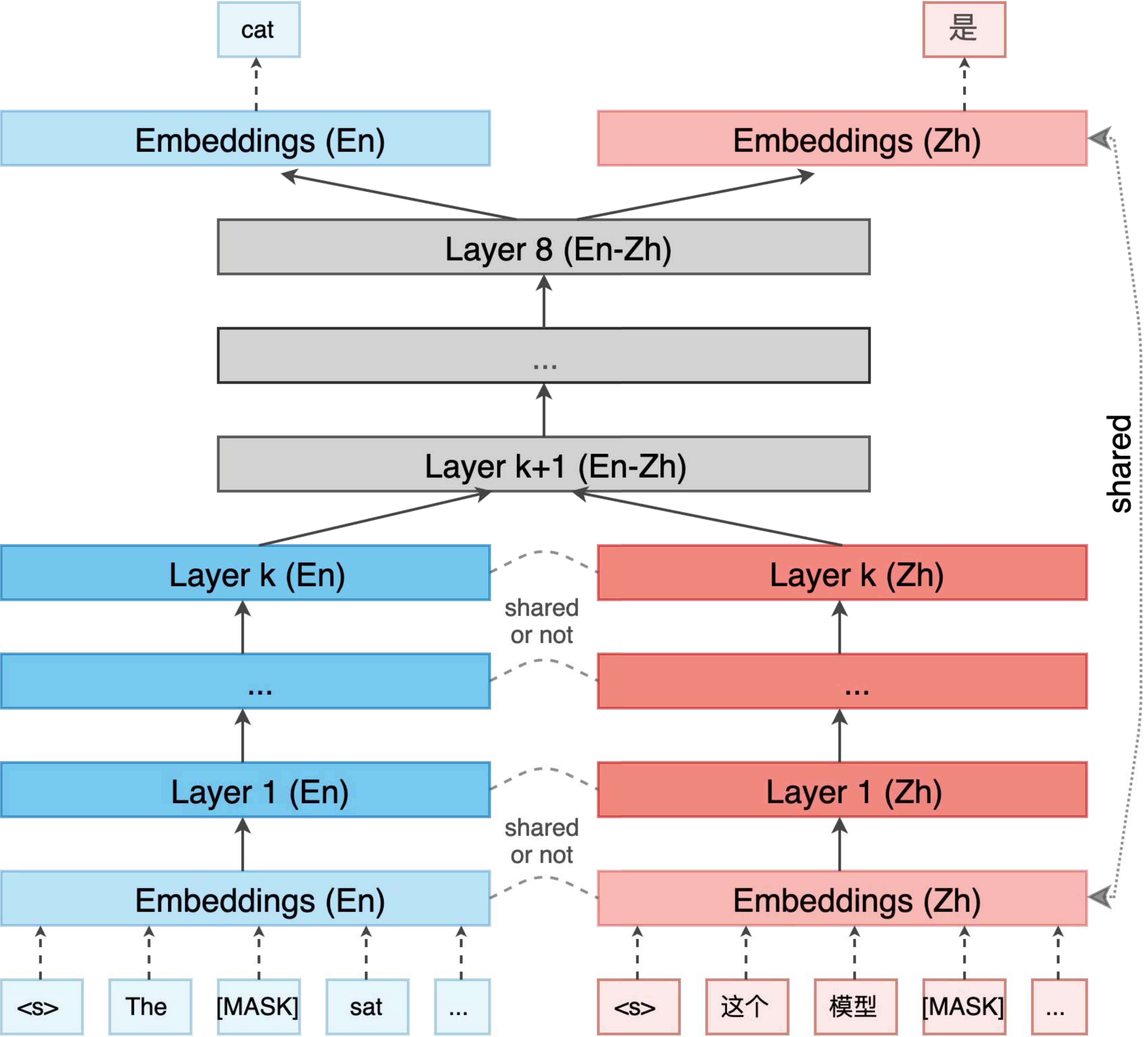}
\caption{On the impact of anchor points and parameter sharing on the emergence of multilingual representations. We train bilingual masked language models and remove parameter sharing for the embedding layers and first few Transformers layers to probe the impact of anchor points and shared structure on cross-lingual transfer.}
\label{fig:params-1}
\end{figure*}

\begin{figure*}[]
\centering
\includegraphics[width=0.6\columnwidth]{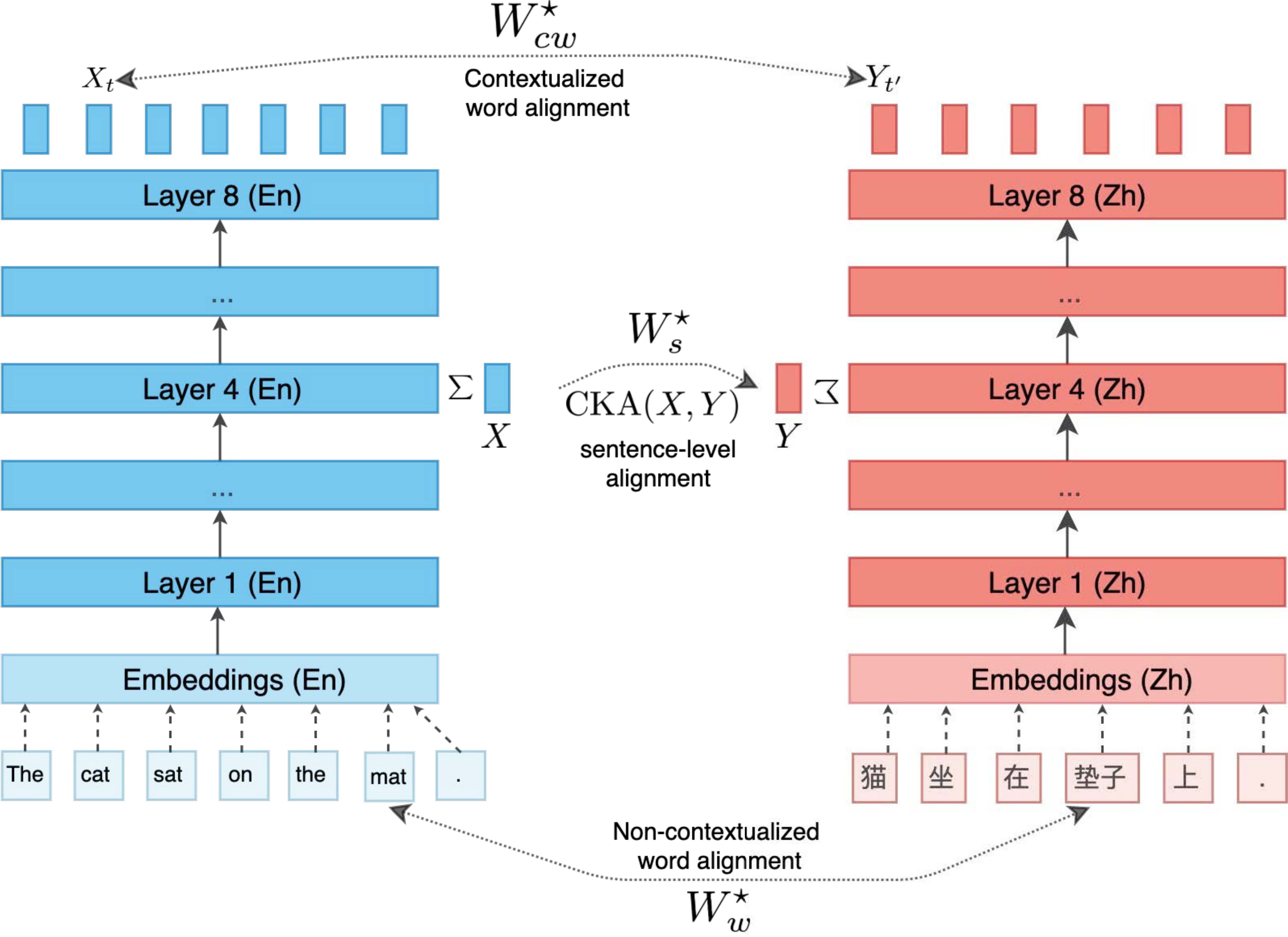}
\caption{Probing the layer similarity of monolingual BERT models. We investigate the similarity of separate monolingual BERT models at different levels. We use an orthogonal mapping between the pooled representations of each model. We also quantify the similarity using the centered kernel alignment (CKA) similarity index.}
\label{fig:params-2}
\end{figure*}

\section{Cross-lingual Pretraining}\label{sec:pretraining}
We study a standard multilingual masked language modeling formulation and evaluate performance on several different cross-lingual transfer tasks, as described in this section. 

\subsection{Multilingual Masked Language Modeling}
Our multilingual masked language models follow the setup used by both mBERT and XLM. We use the implementation of \citet{lample2019cross}. Specifically, we consider continuous streams of 256 tokens and mask 15\% of the input tokens which we replace 80\% of the time by a mask token, 10\% of the time with the original word, and 10\% of the time with a random word. Note the random words could be foreign words. The model is trained to recover the masked tokens from its context~\cite{taylor1953cloze}. The subword vocabulary and model parameters are shared across languages. Note the model has a softmax prediction layer shared across languages. We use Wikipedia for training data, preprocessed by Moses \cite{koehn-etal-2007-moses} and Stanford word segmenter (for Chinese only) and BPE~\cite{sennrich-etal-2016-neural} to learn subword vocabulary. During training, we sample a batch of continuous streams of text from one language proportionally to the fraction of sentences in each training corpus, exponentiated to the power $0.7$. 

\subsection{Pretraining Details} Each model is a Transformer \cite{vaswani2017attention} with 8 layers, 12 heads and GELU activiation functions~\cite{hendrycks2016bridging}. The output softmax layer is tied with input embeddings \cite{press-wolf-2017-using}. The embeddings dimension is 768, the hidden dimension of the feed-forward layer is 3072, and dropout is 0.1. We train our models with the Adam optimizer \cite{kingma2014adam} and the inverse square root learning rate scheduler of \newcite{vaswani2017attention} with $10^{-4}$ learning rate and 30k linear warm up steps. For each model, we train it with 8 NVIDIA V100 GPUs with 32GB of memory and mixed precision. It takes around 3 days to train one model. We use batch size 96 for each GPU and each epoch contains 200k batches. We stop training at epoch 200 and select the best model based on English dev perplexity for evaluation.

\section{Cross-lingual Evaluation}
\label{sec:emerging-eval}
We consider three NLP tasks to evaluate performance: natural language inference (NLI), named entity recognition (NER) and dependency parsing (Parsing). Similar to \autoref{chap:surprising-mbert}, we adopt the \textbf{zero-shot cross-lingual transfer} setting, where we (1) fine-tune the pretrained model on English and (2) directly transfer the model to target languages. We select the model and tune hyperparameters with the English dev set. We report the result on average of the best two sets of hyperparameters.

\subsection{Fine-tuning Details} We fine-tune the model for 10 epochs for NER and Parsing and 200 epochs for NLI. We search the following hyperparameter for NER and Parsing: batch size $\{16, 32\}$; learning rate $\{\text{2e-5},\text{3e-5},\text{5e-5}\}$. For XNLI, we search: batch size $\{4, 8\}$; encoder learning rate $\{\text{1.25e-6},\text{2.5e-6},\text{5e-6}\}$; classifier learning rate $\{\text{5e-6},\text{2.5e-5},\text{1.25e-4}\}$. We use Adam with a fixed learning rate for XNLI and warmup the learning rate for the first 10\% batch then decrease linearly to 0 for NER and Parsing. We save a checkpoint after each epoch.

\subsection{Natural Language Inference} We use the cross-lingual natural language inference (XNLI) dataset \cite{conneau-etal-2018-xnli}. The task-specific layer is a linear mapping to a softmax classifier, which takes the representation of the first token as input. 

\subsection{Named Entity Recognition} We use WikiAnn \cite{pan-etal-2017-cross}, a silver NER dataset built automatically from Wikipedia, for English-Russian and English-French. For English-Chinese, we use CoNLL 2003 English \cite{tjong-kim-sang-de-meulder-2003-introduction} and a Chinese NER dataset~\cite{levow-2006-third}, with realigned Chinese NER labels based on the Stanford word segmenter. We model NER as BIO tagging. Similar to \autoref{sec:task-ner}, the task-specific layer is a linear mapping to a softmax classifier, which takes the representation of the first subword of each word as input. We report span-level F1. We adopt the same post-processing heuristic steps as \autoref{sec:task-ner}.

\subsection{Dependency Parsing} Finally, we use the Universal Dependencies (UD v2.3) \cite{ud2.3} for dependency parsing. We consider the following four treebanks: English-EWT, French-GSD, Russian-GSD, and Chinese-GSD. The task-specific layer is a graph-based parser \cite{dozat2016deep}, using representations of the first subword of each word as inputs, same as \autoref{sec:task-parsing}. We measure performance with the labeled attachment score (LAS).

\section{What Makes mBERT Multilingual?}\label{sec:mbert-ablation}

\begin{figure*}[]
\centering
\includegraphics[width=\columnwidth]{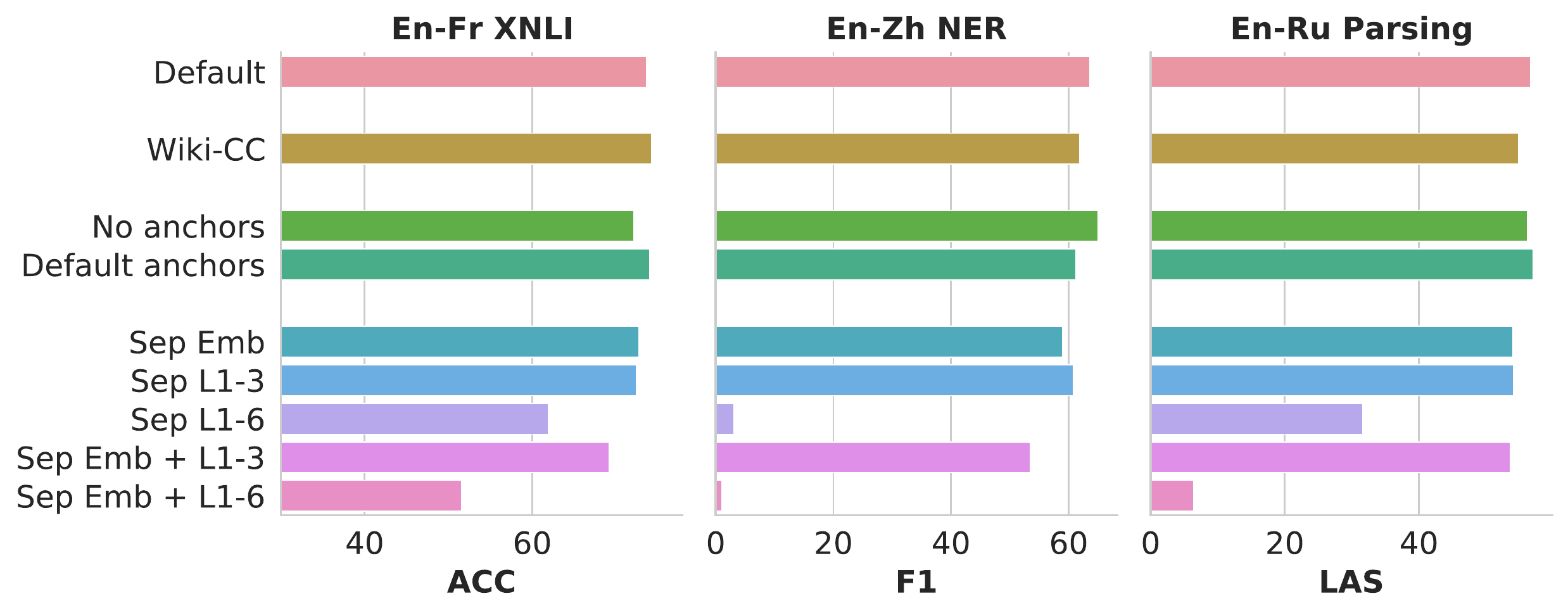}
\caption{Cross-lingual transfer of bilingual MLM on three tasks and language pairs under different settings.
Other tasks and language pairs follow similar trends. See \autoref{tab:emerging-main} for full results.}\label{fig:all}
\end{figure*}

We hypothesize that the following factors play important roles in what makes multilingual BERT multilingual: domain similarity, shared vocabulary (or anchor points), shared parameters, and language similarity. Without loss of generality, we focus on bilingual MLM. We consider three pairs of languages with different levels of language similarity: English-French, English-Russian, and English-Chinese. 

\input{table/emerging/main}

\subsection{Domain Similarity}
\label{sec:domain-similarity}

Multilingual BERT and XLM are trained on the Wikipedia comparable corpora. Domain similarity has been shown to affect the quality of cross-lingual word embeddings~\cite{conneau2017word}, but this effect is not well established for masked language models. We consider domain differences by training on Wikipedia for English and a random subset of Common Crawl of the same size for the other languages (\textbf{Wiki-CC}). We also consider a model trained with Wikipedia only (\textbf{Default}) for comparison.

The first group in \autoref{tab:emerging-main} shows domain mismatch has a relatively modest effect on performance. XNLI and parsing performance drop around 2 points while NER drops over 6 points for all languages on average. One possible reason is that the labeled WikiAnn data for NER consists of Wikipedia text; domain differences between source and target language during pretraining hurt performance more. Indeed for English and Chinese NER, where neither side comes from Wikipedia, performance only drops around 2 points.

\subsection{Anchor Points}
\label{sec:anchor-points}

Anchor points are \textit{identical strings} that appear in both languages in the training corpus. Translingual words like \textit{DNA} or \textit{Paris} appear in the Wikipedia of many languages with the same meaning. In mBERT, anchor points are naturally preserved due to joint BPE and shared vocabulary across languages. Anchor point existence has been suggested as a key ingredient for effective cross-lingual transfer since they allow the shared encoder to have at least some direct tying of meaning across different languages~\cite{lample2019cross,pires-etal-2019-multilingual,wu-dredze-2019-beto}. However, this effect has not been carefully measured. 

We present a controlled study of the impact of anchor points on cross-lingual transfer performance by varying the amount of shared subword vocabulary across languages. 
Instead of using a single joint BPE with 80k merges, we use language-specific BPE with 40k merges for each language. We then build vocabulary by taking the union of the vocabulary of two languages and train a bilingual MLM (\textbf{Default anchors}). To remove anchor points, we add a language prefix to each word in the vocabulary before taking the union. Bilingual MLM (\textbf{No anchors}) trained with such data has no shared vocabulary across languages. However, it still has a single softmax prediction layer shared across languages and tied with input embeddings.

The second group of \autoref{tab:emerging-main} shows cross-lingual transfer performance under the two anchor point conditions. 
Surprisingly, effective transfer is still possible with no anchor points. Comparing no anchors and default anchors, the performance of XNLI and parsing drops only around 1 point while NER even improves 1 point averaging over three languages.
Overall, these results show that we have previously overestimated the contribution of anchor points during multilingual pretraining.
Concurrent to the publication of this chapter, \newcite{karthikeyan2020cross} similarly find anchor points play a minor role in learning cross-lingual representation.

\subsection{Parameter Sharing}
\label{sec:param-sharing}

Given that anchor points are not required for transfer, a natural next question is the extent to which we need to tie the parameters of the transformer layers. Sharing the parameters of the top layer is necessary to provide shared inputs to the task-specific layer. However, 
as seen in \autoref{fig:params-1},
we can progressively separate the \textit{bottom} layers 1:3 and 1:6 of the Transformers and/or the embedding layers (including positional embeddings) (\textbf{Sep Emb}; \textbf{Sep L1-3}; \textbf{Sep L1-6}; \textbf{Sep Emb + L1-3}; \textbf{Sep Emb + L1-6}). Since the prediction layer is tied with the embeddings layer, separating the embeddings layer also introduces a language-specific softmax prediction layer for the cloze task. This effectively introduces language specific component into the multilingual model. In theory, such language specific component might learn to encode language specific information into a shared space, benefiting the learning of cross-lingual representation. Additionally, in this group of experiment, we only sample random words within one language during the MLM pretraining, as MLM pretraining would potentially introduce accidental anchor points during random word replacement. During fine-tuning on the English training set, we freeze the language-specific layers and only fine-tune the shared layers.

The third group in \autoref{tab:emerging-main} shows cross-lingual transfer performance under different parameter sharing conditions with ``Sep'' denoting which layers \textbf{is not} shared across languages. Sep Emb (effectively no anchor point) drops more than No anchors with 3 points on XNLI and around 1 point on NER and parsing, suggesting having a cross-language softmax layer also helps to learn cross-lingual representations. Performance degrades as fewer layers are shared for all pairs, and again the less closely related language pairs lose the most. Most notably, the cross-lingual transfer performance drops to random when separating embeddings and bottom 6 layers of the transformer. However, reasonably strong levels of transfer are still possible without tying the bottom three layers. These trends suggest that parameter sharing is the key ingredient that enables the learning of an effective cross-lingual representation space, and having language-specific capacity does not help learn a language-specific encoder for cross-lingual representation despite having extra parameters.

\subsection{Language Similarity}

Finally, in contrast to many of the experiments above, language similarity seems to be quite important for effective transfer.  
Looking at \autoref{tab:emerging-main} column by column in each task, we observe performance drops as language pairs become more distantly related.
The more complex tasks seem to have larger performance gaps and having language-specific capacity does not seem to be the solution.

\subsection{Conclusion}

Summarised by \autoref{fig:all}, parameter sharing is the most important factor.
Anchor points and shared softmax projection parameters are not necessary for effective cross-lingual transfer.
Joint BPE and domain similarity contribute a little in learning cross-lingual representation.

\section{How Does Parameter Sharing Enable Cross-lingual Representation?}

In \autoref{sec:mbert-ablation}, we observe that parameter sharing is the key for learning cross-lingual representation. How does parameter sharing enable cross-lingual representation? Our hypothesis is that the representations that the models learn for different languages are similarly shaped and
during multilingual pretraining, the models naturally align its representation across languages. If the hypothesis were true, we would be able to show that independently trained monolingual BERT models learn representations that are similar across languages, much like the widely observed similarities in word embedding spaces.

In this section, we show that independent monolingual BERT models produce highly similar representations when evaluated at the word level (\autoref{sec:align-noncontextual-word}), contextual word-level (\autoref{sec:align-contextual-word}),
and sentence level (\autoref{sec:align-sentence}) . We also plot the cross-lingual similarity of neural network activation with center kernel alignment (\autoref{sec:cka}) at each layer. We consider five languages: English, French, German, Russian, and Chinese.

\subsection{Aligning Monolingual BERTs}\label{sec:align}

To measure similarity, we learn an orthogonal mapping using the Procrustes~\cite{smith2017offline} approach:
\begin{align}
W^{\star} = \underset{W \in O_d(\mathbb{R})}{\text{argmin}} \Vert W X - Y \Vert_{F} = UV^T
\end{align}
with $U\Sigma V^T = \text{SVD}(Y X^T)$, where $X$ and $Y$ are representations of two monolingual BERT models, sampled at different granularities as described below. We apply iterative normalization on $X$ and $Y$ before learning the mapping \cite{zhang-etal-2019-girls}.

\subsubsection{Word-level Alignment}\label{sec:align-noncontextual-word}

\begin{figure*}[]
\centering
\subfloat[][Non-contextual word embeddings alignment]{
\includegraphics[width=0.5\columnwidth]{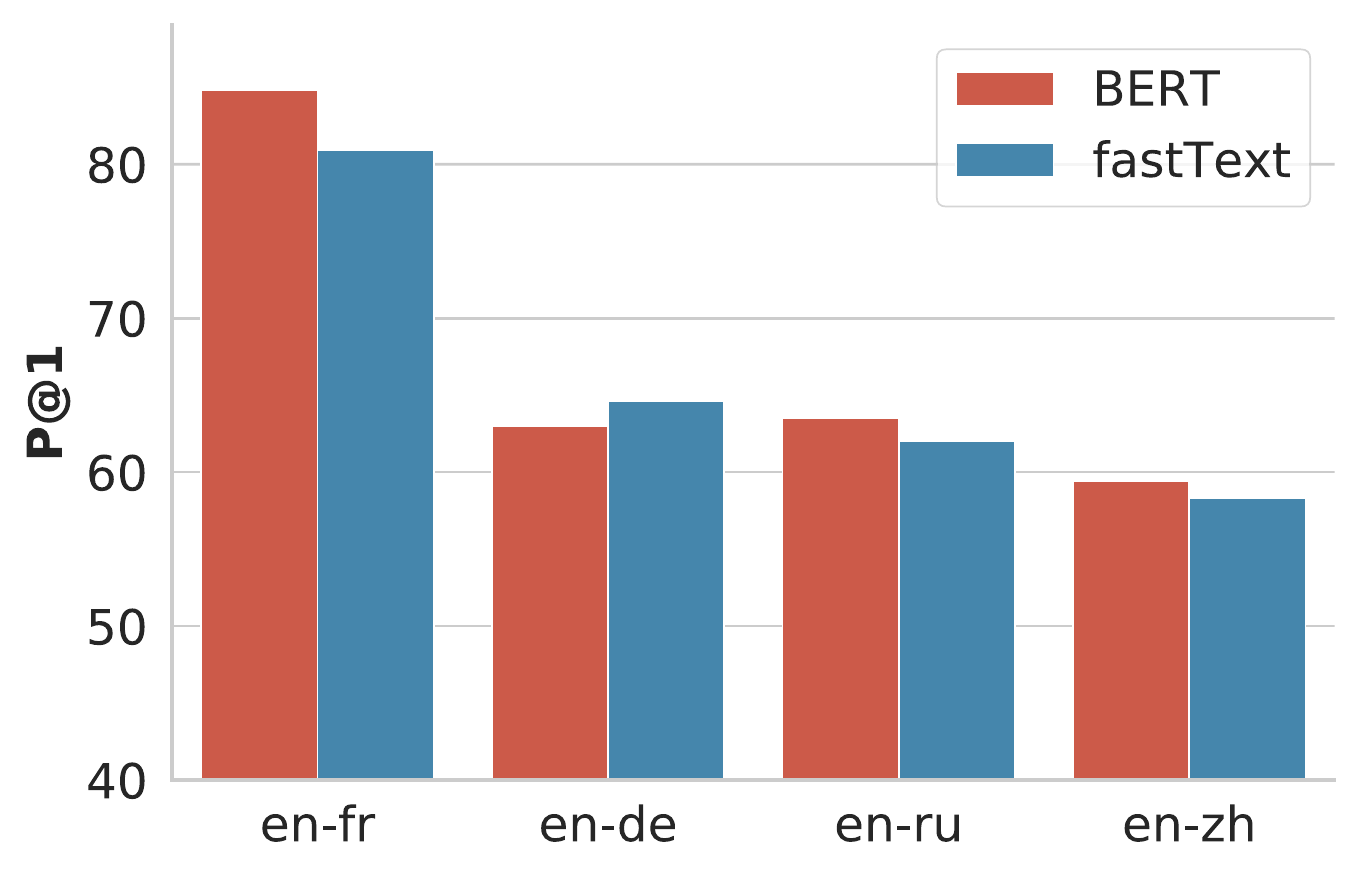}
\label{fig:align-noncontextual-word}}
\subfloat[][Contextual word embedding alignment]{
\includegraphics[width=0.5\columnwidth]{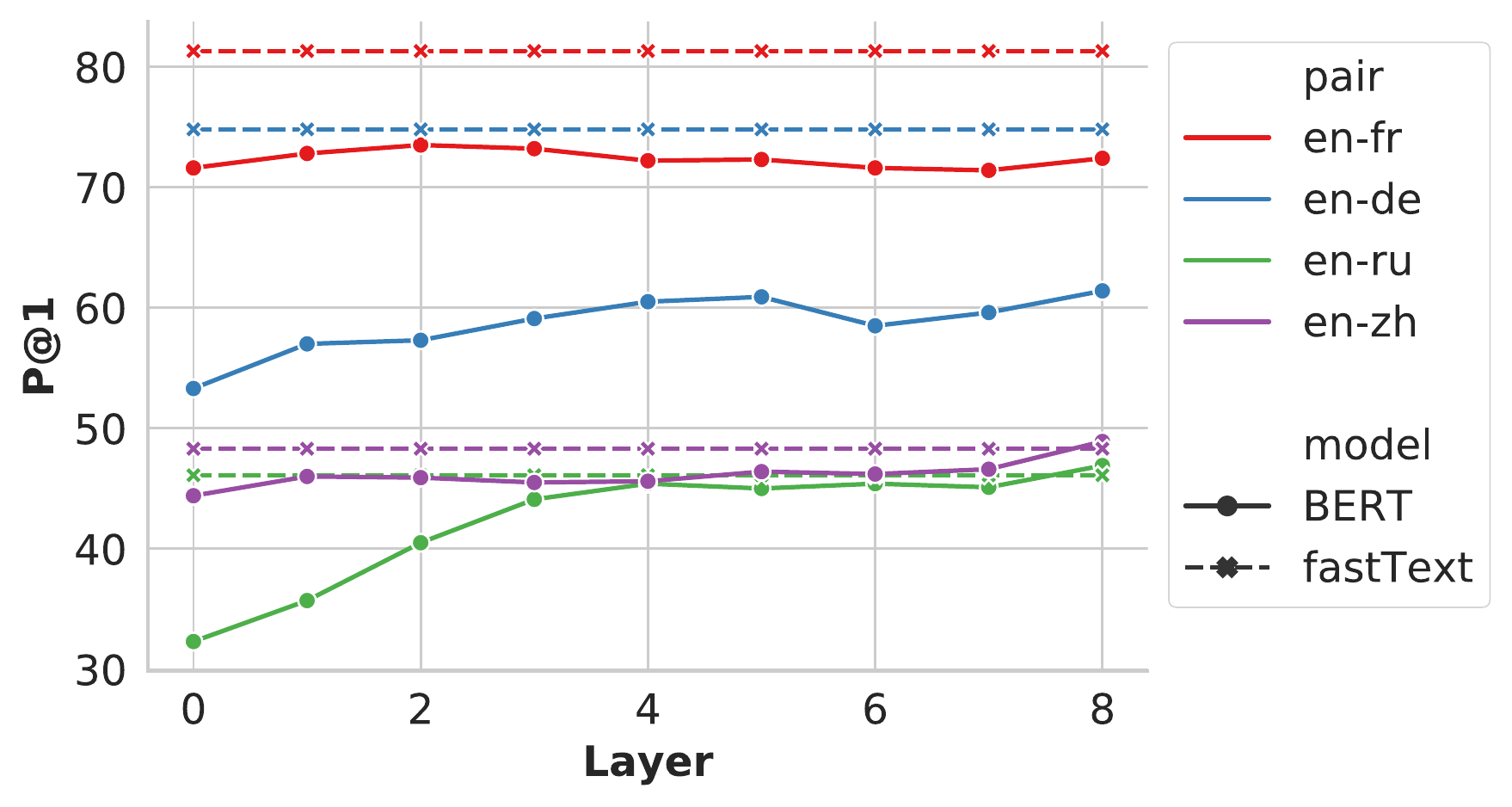}
\label{fig:align-contextual-word}}
\caption{Alignment of word-level representations from monolingual BERT models on a subset of MUSE benchmark. \autoref{fig:align-noncontextual-word} and \autoref{fig:align-contextual-word} are not comparable due to different embedding vocabularies.}\label{fig:align-word}
\end{figure*}

In this section, we align both the non-contextual word representations from the embedding layers, and the contextual word representations from the hidden states of the Transformer at each layer.

For non-contextualized word embeddings, we define $X$ and $Y$ as the word embedding layers of monolingual BERT, which contain a single embedding per word (type). Note that in this case we only keep words containing only one subword. For contextualized word representations, we first encode 500k sentences in each language. At each layer, and for each word, we collect all contextualized representations of a word in the 500k sentences and average them to get a single embedding. Since BERT operates at the subword level, for one word we consider the average of all its subword embeddings. Eventually, we get one word embedding per layer.  We use the MUSE benchmark \cite{conneau2017word}, a bilingual dictionary induction dataset for alignment supervision and evaluate the alignment on word translation retrieval.
As a baseline, we use the first 200k embeddings of fastText \cite{bojanowski-etal-2017-enriching} and learn the mapping using the same procedure as \autoref{sec:align}. Note we use a subset of 200k vocabulary of fastText, the same as BERT, to get a comparable number. We retrieve word translation using CSLS \cite{conneau2017word} with K=10.

In \autoref{fig:align-word}, we report the alignment results under these two settings. \autoref{fig:align-noncontextual-word} shows that the subword embeddings matrix of BERT, where each subword is a standalone word, can easily be aligned with an orthogonal mapping and obtain slightly better performance than the same subset of fastText. \autoref{fig:align-contextual-word} shows embeddings matrix with the average of all contextual embeddings of each word can also be aligned to obtain a decent quality bilingual dictionary, although underperforming fastText. We notice that using contextual representations from higher layers obtain better results compared to lower layers.

\subsubsection{Contextual Word-level Alignment}\label{sec:align-contextual-word}

\begin{figure*}[]
\centering
\includegraphics[width=1\columnwidth]{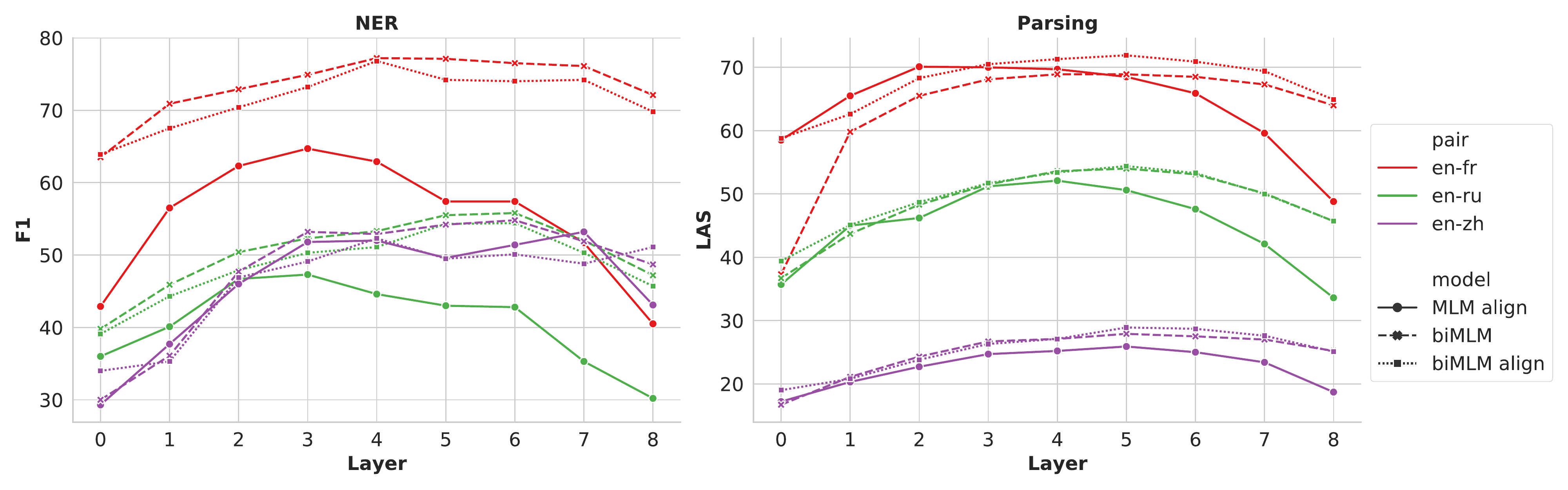}
\caption{Contextual representation alignment of different layers for zero-shot cross-lingual transfer.}\label{fig:align-contextual}
\end{figure*}

In addition to aligning word representations, we also align representations of two monolingual BERT models in contextual settings, and evaluate performance on cross-lingual transfer for NER and parsing. We take the Transformer layers of each monolingual model up to layer $i$, and learn a mapping $W$ from layer $i$ of the target model to layer $i$ of the source model. To create that mapping, we use the same Procrustes approach but use a dictionary of parallel contextual words, obtained by running the fastAlign~\cite{dyer-etal-2013-simple} model on the 10k XNLI parallel sentences. %

For each downstream task, we learn task-specific layers on top of $i$-th English layer: four Transformer layers and a task-specific layer. We learn these on the training set, but keep the first $i$ pretrained layers freezed. After training these task-specific parameters, we encode (say) a Chinese sentence with the first $i$ layers of the target Chinese BERT model, project the contextualized representations back to the English space using the $W$ we learned, and then use the task-specific layers for NER and parsing.

In \autoref{fig:align-contextual}, we vary $i$ from the embedding layer (layer 0) to the last layer (layer 8) and present the results of our approach on parsing and NER. We also report results using the first $i$ layers of a bilingual MLM (biMLM), and the same alignment step with biMLM.
We show that aligning monolingual models (MLM align) obtain relatively good performance even though they perform worse than bilingual MLM, except for parsing in English-French.
However, the same alignment step with biMLM only shows improvement in parsing.
The results of monolingual alignment generally show that we can align contextual representations of monolingual BERT models with a simple linear mapping and use this approach for cross-lingual transfer. We also observe that the model obtains the highest transfer performance with the middle layer representation alignment, and not the last layers. The performance gap between monolingual MLM alignment and bilingual MLM is higher in NER compared to parsing, suggesting the syntactic information needed for parsing might be easier to align with a simple mapping while entity information requires more explicit entity alignment.

\subsubsection{Sentence-level Alignment}\label{sec:align-sentence}

\begin{figure}[]
\begin{center}
\includegraphics[width=0.6\columnwidth]{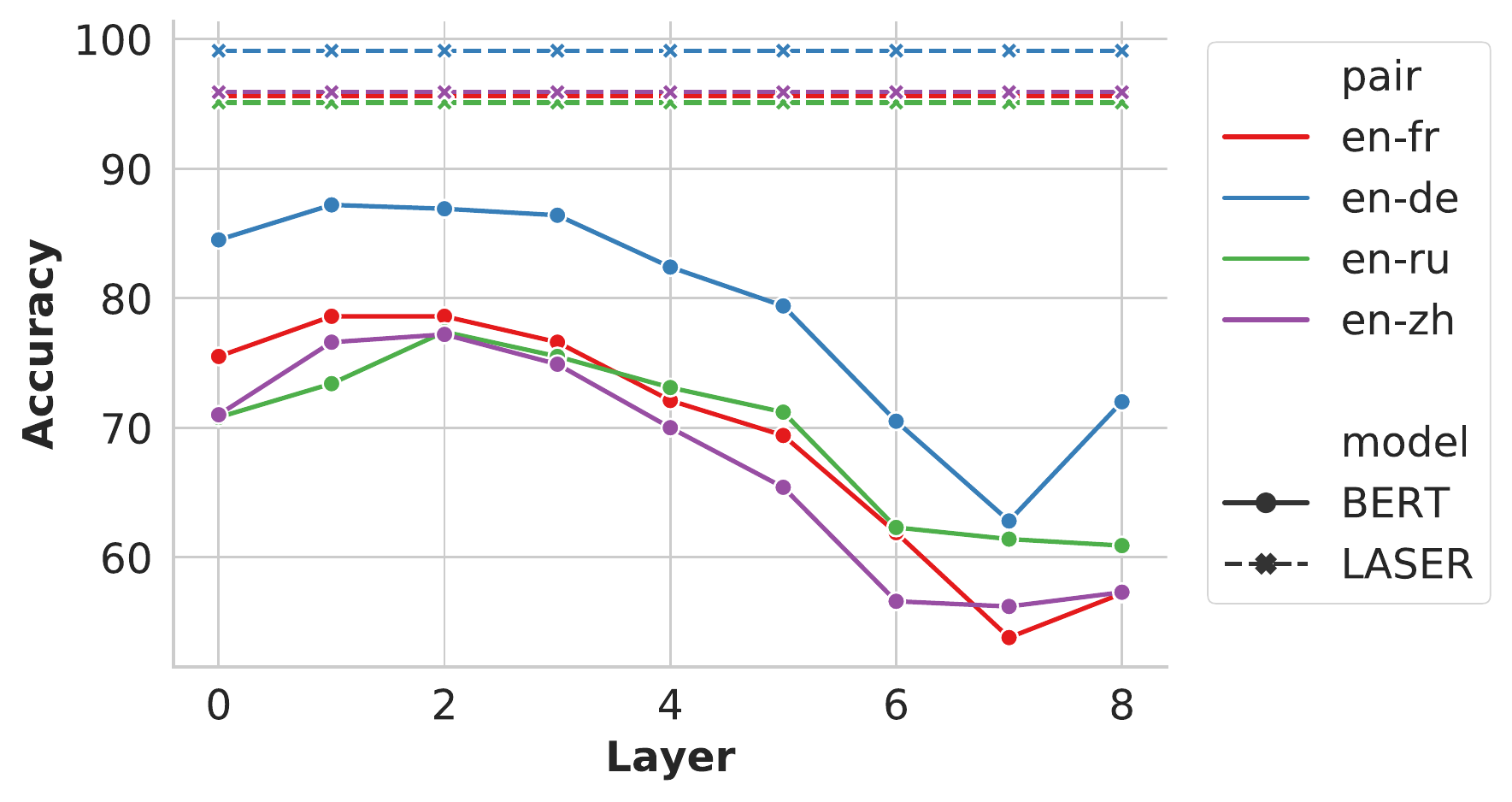}
\caption{Parallel sentence retrieval accuracy after Procrustes alignment of monolingual BERT models.
\label{fig:tatoeba}}
\end{center}
\end{figure}

In this case, $X$ and $Y$ are obtained by average pooling subword representation (excluding special token) of sentences \textit{at each layer} of monolingual BERT. We use multi-way parallel sentences from XNLI for alignment supervision and Tatoeba \cite{artetxe-schwenk-2019-massively} for evaluation.

\autoref{fig:tatoeba} shows the sentence similarity search results with nearest neighbor search and cosine similarity, evaluated by precision at 1, with four language pairs. Here the best result is obtained at lower layers. The performance is surprisingly good given we only use 10k parallel sentences to learn the alignment without fine-tuning at all. As a reference, the state-of-the-art performance is over 95\%, obtained by LASER \cite{artetxe-schwenk-2019-massively} trained with millions of parallel sentences.

\subsubsection{Conclusion} These findings demonstrate that both word-level, contextual word-level, and sentence-level BERT representations can be aligned with a simple orthogonal mapping. Similar to the alignment of word embeddings~\cite{mikolov2013exploiting}, this shows that BERT models are similar across languages. This result gives more intuition on why mere parameter sharing is sufficient for multilingual representations to emerge in multilingual masked language models.

\subsection{Neural Network Similarity}\label{sec:cka}

\begin{figure*}[]
\centering
\includegraphics[width=\columnwidth]{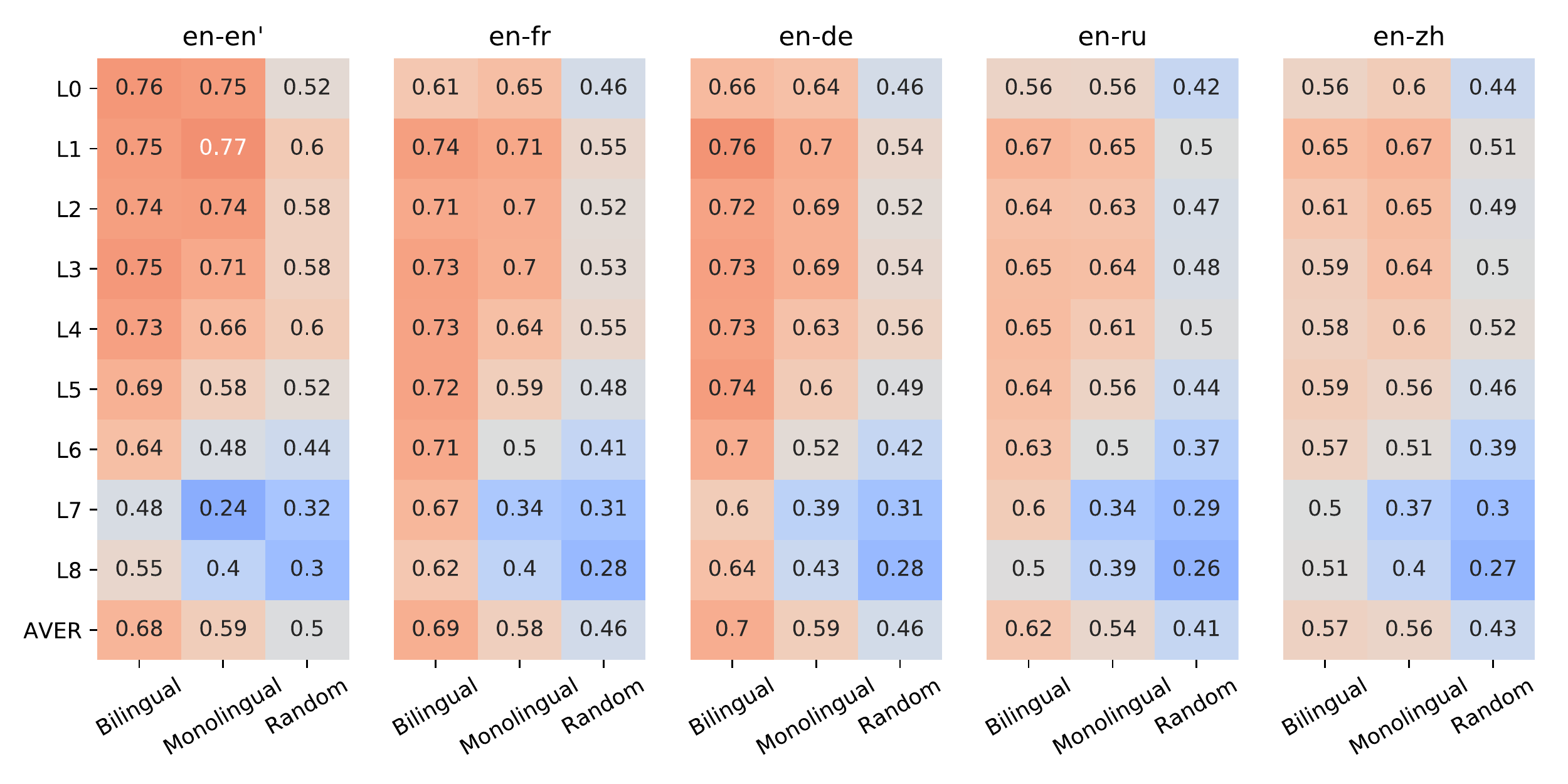}
\caption{CKA similarity of mean-pooled multi-way parallel sentence representation at each layer. Note en$^\prime$ corresponds to paraphrases of en obtained from back-translation (en-fr-en$^\prime$). Random encoder is only used by non-Engligh sentences. L0 is the embedding layer while L1 to L8 are the corresponding transformer layers. The average row is the average of 9 (L0-L8) similarity measurements.}\label{fig:cka}
\end{figure*}

Based on the work of \newcite{kornblith2019similarity}, we examine the centered kernel alignment (CKA), a neural network similarity index that improves upon canonical correlation analysis (CCA), and use it to measure the similarity across both monolingual and bilingual masked language models.
The linear CKA is both invariant to orthogonal transformation and isotropic scaling, but are not invertible to any linear transform.
The linear CKA similarity measure is defined as follows:
\begin{align}
\text{CKA}(X,Y) = \frac{\Vert Y^TX \Vert^2_{\text{F}}}{(\Vert X^TX\Vert_{\text{F}} \Vert Y^TY\Vert_{\text{F}})},
\end{align}
where $X$ and $Y$ correspond respectively to the matrix of the $d$-dimensional mean-pooled (excluding special token) subword representations at layer $l$ of the $n$ parallel source and target sentences.

In \autoref{fig:cka}, we show the CKA similarity of monolingual models, compared with bilingual models and random encoders, of multi-way parallel sentences \cite{conneau-etal-2018-xnli} for five languages pair: English to English$^\prime$ (obtained by back-translation from French), French, German, Russian, and Chinese. The monolingual en$^\prime$ is trained on the same data as en but with different random seed and the bilingual en-en$^\prime$ is trained on English data but with separate embeddings matrix as in \autoref{sec:param-sharing}. The rest of the bilingual MLM is trained with the Default setting. We only use random encoder for non-English sentences.

\autoref{fig:cka} shows bilingual models have slightly higher similarity compared to monolingual models with random encoders serving as a lower bound. Despite the slightly lower similarity between monolingual models, it still explains the alignment performance in \autoref{sec:align}. Because the measurement is also invariant to orthogonal mapping, the CKA similarity is highly correlated with the sentence-level alignment performance in \autoref{fig:tatoeba} with over 0.9 Pearson correlation for all four languages pairs. For monolingual and bilingual models, the first few layers have the highest similarity, which explains why \autoref{sec:mbert-layer} finds freezing bottom layers of mBERT helps cross-lingual transfer.
On the other hand, the final few layers have the lowest similarity, perhaps because the model needs language specific information to solve the cloze task, which explains why \autoref{sec:mbert-langid} finds that mBERT retains strong language specific information.
The similarity gap between monolingual model and bilingual model decreases as the languages pair become more distant. In other words, when languages are similar, using the same model increases representation similarity. On the other hand, when languages are dissimilar, using the same model does not help representation similarity much.

\section{Discussion}
In this chapter, we show that multilingual representations can emerge from unsupervised multilingual masked language models with only parameter sharing of some Transformer layers. Even without any anchor points, the model can still learn to map representations coming from different languages in a single shared embedding space. We also show that isomorphic embedding spaces emerge from monolingual masked language models in different languages, similar to word2vec embedding spaces~\cite{mikolov2013exploiting}. By using a linear mapping, we are able to align the embedding layers and the contextual representations of Transformers trained in different languages. We also use the CKA neural network similarity index to probe the similarity between BERT Models and show that the early layers of the Transformers are more similar across languages than the last layers. All of these effects were stronger for more closely related languages, suggesting there is room for significant improvements on more distant language pairs. 

This type of emergent language universality has interesting theoretical and practical implications. We gain insight into why the models learn cross-lingual representation and open up new lines of inquiry into the implication of such emerge universality. It should be possible to adapt multilingual pretrained models to new languages with little additional training. For example, pretrained multilingual MLM models can be rapidly fine-tuned to another language \cite{artetxe-etal-2020-cross,chau-etal-2020-parsing,wang-etal-2020-extending,pfeiffer2020unks}.

%% file: table/emerging/main.tex
\begin{table*}[]
\begin{center}
\resizebox{1\linewidth}{!}{
\begin{tabular}[b]{c|ccccc|cccc|cccc|cccc}
\toprule
\multirow{2}{*}{Model} & \multirow{2}{*}{Domain} & \multirow{2}{*}{BPE Merges} & \multirow{2}{*}{Anchors Pts} & \multirow{2}{*}{Share Param.} & \multirow{2}{*}{Softmax} & \multicolumn{4}{c|}{XNLI (Acc)} &
\multicolumn{4}{c|}{NER (F1)} &
\multicolumn{4}{c}{Parsing (LAS)} \\
& & & & & & fr & ru & zh & $\Delta$ & fr & ru & zh & $\Delta$ & fr & ru & zh & $\Delta$ \\
\midrule
\textbf{Default} & Wiki-Wiki & 80k & all & all & shared & 73.6 & 68.7 & 68.3 & 0.0 & 79.8 & 60.9 & 63.6 & 0.0 & 73.2 & 56.6 & 28.8 & 0.0 \\
\midrule 
\multicolumn{15}{l}{\textit{Domain Similarity} (\autoref{sec:domain-similarity})} \\
\midrule
\textbf{Wiki-CC} & Wiki-CC & - & - & - & - & 74.2 & 65.8 & 66.5 & -1.4 & 74.0 & 49.6 & 61.9 & -6.2 & 71.3 & 54.8 & 25.2 & -2.5 \\
\midrule 
\multicolumn{15}{l}{\textit{Anchor Points} (\autoref{sec:anchor-points})} \\
\midrule
\textbf{No anchors} & - & 40k/40k & 0 & - & - & 72.1 & 67.5 & 67.7 & -1.1 & 74.0 & 57.9 & 65.0 & -2.4 & 72.3 & 56.2 & 27.4 & -0.9 \\
\textbf{Default anchors} & - & 40k/40k & - & - & - & 74.0 & 68.1 & 68.9 & +0.1 & 76.8 & 56.3 & 61.2 & -3.3 & 73.0 & 57.0 & 28.3 & -0.1 \\
\midrule 
\multicolumn{15}{l}{\textit{Parameter Sharing} (\autoref{sec:param-sharing})} \\
\midrule
\textbf{Sep Emb} & - & 40k/40k & 0* & Sep Emb & lang-specific & 72.7 & 63.6 & 60.8 & -4.5 & 75.5 & 57.5 & 59.0 & -4.1 & 71.7 & 54.0 & 27.5 & -1.8 \\
\textbf{Sep L1-3} & - & 40k/40k & - & Sep L1-3 & - & 72.4 & 65.0 & 63.1 & -3.4 & 74.0 & 53.3 & 60.8 & -5.3 & 69.7 & 54.1 & 26.4 & -2.8 \\
\textbf{Sep L1-6} & - & 40k/40k & - & Sep L1-6 & - & 61.9 & 43.6 & 37.4 & -22.6 & 61.2 & 23.7 & 3.1 & -38.7 & 61.7 & 31.6 & 12.0 & -17.8 \\
\textbf{Sep Emb + L1-3} & - & 40k/40k & 0* & Sep Emb + L1-3 & lang-specific & 69.2 & 61.7 & 56.4 & -7.8 & 73.8 & 46.8 & 53.5 & -10.0 & 68.2 & 53.6 & 23.9 & -4.3 \\
\textbf{Sep Emb + L1-6} & - & 40k/40k & 0* & Sep Emb + L1-6 & lang-specific & 51.6 & 35.8 & 34.4 & -29.6 & 56.5 & 5.4 & 1.0 & -47.1 & 50.9 & 6.4 & 1.5 & -33.3 \\
\bottomrule
\end{tabular}
}
\caption{Dissecting bilingual MLM based on zero-shot cross-lingual transfer performance. - denote the same as the first row (\textbf{Default}). $\Delta$ denote the difference of average task performance between a model and \textbf{Default}.
\label{tab:emerging-main}}
\end{center}
\end{table*}

%% file: src/low-resource.tex
\section{Introduction}
In \autoref{chap:surprising-mbert}, we show that mBERT learns high-quality cross-lingual representation and has strong zero-shot cross-lingual transfer performance.
However, evaluations have focused on high resource languages, with cross-lingual transfer using English as a source language or within language performance. As \autoref{chap:surprising-mbert} evaluates mBERT on 39 languages, this leaves the majority of mBERT's 104 languages, most of which are low resource languages, untested. 

In this chapter, we ask the following question. \textit{Does mBERT learn equally high-quality representation for its 104 languages?} If not, which languages are hurt by its massively multilingual style pretraining? While it has been observed that for high resource languages like English, mBERT performs worse than monolingual BERT on English with the same capacity \cite{multilingualBERTmd}. 
It is unclear that for low resource languages (in terms of monolingual corpus size), how does mBERT compare to a monolingual BERT? And, does multilingual joint training help mBERT learn better representation for low resource languages?

To answer this question, we first evaluate the representation quality of mBERT on 99 languages for NER, and 54 for part-of-speech tagging and dependency parsing. 
We show mBERT does not have equally high-quality representation for all of the 104 languages, with the bottom 30\% languages performing much worse than a non-BERT model on NER. Additionally, by training various monolingual BERT for low-resource languages with the same data size, we show the low representation quality of low-resource languages is not the result of the hyperparameters of BERT or sharing the model with a large number of languages, as monolingual BERT performs worse than mBERT. On the contrary, by pairing low-resource languages with linguistically-related languages, we show low-resource languages benefit from multilingual joint training, as bilingual BERT outperforms monolingual BERT while still lacking behind mBERT.

These experiments suggest that mBERT try its best to learn representation for the low resource languages with the given data. However, as BERT pretraining objective is not known for sample efficient, the small Wikipedia of low resource languages is not enough for mBERT to learn high quality representation. To address this challenge, we either need a more sample efficient pretraining algorithm or collect more data to make low resource languages high resource.

\section{Background}

Several factors need to be considered in understanding mBERT. First, the 104 most common Wikipedia languages vary considerably in size (\autoref{tab:low-lang}). Therefore, mBERT training attempted to equalize languages by up-sampling sentences from low resource languages and down-sampling sentences from high resource languages. 
Second, while each language may be similarly represented in the training data, subwords are not evenly distributed among the languages. Many languages share common characters and cognates, biasing subword learning to some languages over others. Both of these factors may influence how well mBERT learns representations for low resource languages.
Finally, \citet{baevski-etal-2019-cloze} show that in general 
larger pretraining data for English leads to better downstream performance, yet increasing the size of pretraining data exponentially only increases downstream performance linearly. For a low resource language with limited pretraining data, it is unclear whether contextual representations outperform previous methods.

\subsection{Representations for Low Resource Languages} Embeddings with subword information, a non-contextual representation, like fastText \cite{bojanowski-etal-2017-enriching} and BPEmb \cite{heinzerling-strube-2018-bpemb} are more data-efficient compared to contextual representation like ELMo and BERT when a limited amount of text is available. For low resource languages, there are usually limits on \textbf{monolingual corpora} and \textbf{task specific supervision}. When task-specific supervision is limited, e.g. sequence labeling in low resource languages, mBERT performs better than fastText while underperforming a single BPEmb trained on all languages \cite{heinzerling-strube-2019-sequence}. Contrary to this work, we focus on mBERT from the perspective of representation learning for each language in terms of monolingual corpora resources and analyze how to improve BERT for low resource languages. We also consider parsing in addition to sequence labeling tasks.

\section{Experiments}
We begin by defining high and low resource languages in mBERT, a description of the models and downstream tasks we use for evaluation, followed by a description of the masked language model pretraining.

\subsection{High/Low Resource Languages}\label{sec:wikisize}
\input{table/low/wikirank}

Since mBERT was trained on articles from Wikipedia, a language is considered a high or low resource for mBERT based on the size of Wikipedia in that language. Size can be measured in many ways (articles, tokens, characters); we use the size of the raw dump archive file;\footnote{The size of English (en) is the size of this file: \url{https://dumps.wikimedia.org/enwiki/latest/enwiki-latest-pages-articles.xml.bz2}} for convenience we use $\log_2$ of the size in MB (\textbf{WikiSize}). English is the highest resource language (15.5GB) and Yoruba the lowest (10MB).\footnote{The ordering does not necessarily match the number of speakers for a language.} \autoref{tab:low-lang} shows languages and their relative resources.

\subsection{Downstream Tasks} 
mBERT supports 104 languages, and we seek to evaluate the learned representations for as many of these as possible. We consider three NLP tasks for which annotated task data exists in a large number of languages: named entity recognition (NER), universal part-of-speech (POS) tagging and universal dependency parsing. For each task, we fine-tune a task-specific model built on top of the mBERT using within-language supervised data.

For NER we use data created by \citet{pan-etal-2017-cross}, built automatically from Wikipedia, which covers 99 of the 104 languages supported by mBERT. We evaluate NER with entity-level F1. This data is in-domain as mBERT is pretrained on Wikipedia. For POS tagging and dependency parsing, we use Universal Dependencies (UD) v2.3 \cite{ud2.3}, which covers 54 languages (101 treebanks) supported by mBERT. We evaluate POS with accuracy (ACC) and Parsing with label attachment score (LAS) and unlabeled attachment score (UAS). For POS, we consider UPOS within the treebank. For parsing, we only consider universal dependency labels. The domain is treebank-specific so we use all treebanks of a language for completeness.

\subsubsection{Task Models} For sequence labeling tasks (NER and POS), we add a linear function with a softmax on top of mBERT.
For NER, at test time, we adopt the same post-processing step as \autoref{sec:task-ner}.
For dependency parsing, we replace the LSTM in the graph-based parser of \citet{dozat2016deep} with mBERT. For the parser, we use the original hyperparameters. Note we do not use universal part-of-speech tags as input for dependency parsing. We fine-tune all parameters of mBERT for a specific task.
We use a maximum sequence length of 128 for sequence labeling tasks. For sentences longer than 128, we use a sliding window with 64 previous tokens as context. For dependency parsing, we use sequence length 128 due to memory constraints and drop sentences with more than 128 subwords. We also adopt the same treatment for the baseline \cite{che-etal-2018-towards} to obtain comparable results. Since mBERT operates on the subword-level, we select the first subword of each word for the task-specific layer with masking.

\subsubsection{Task Optimization} We train all models with Adam \cite{kingma2014adam}. We warm up the learning rate linearly in the first 10\% steps then decrease linearly to 0. We select the hyperparameters based on dev set performance by grid search, as recommended by \citet{devlin-etal-2019-bert}. The search includes a learning rate (2e-5, 3e-5, and 5e-5), batch size (16 and 32). 
As task-specific supervision size differs by language or treebank, we fine-tune the model for 10k gradient steps and evaluate the model every 200 steps.
We select the best model and hyperparameters for a language or treebank by the corresponding dev set.

\subsubsection{Task Baselines}
We compare our mBERT models with previously published methods: \citet{pan-etal-2017-cross} for NER; For POS and dependency parsing the best performing system ranked by LAS in the 2018 universal parsing shared task \cite{che-etal-2018-towards}  \footnote{The shared task uses UD v2.2 while we use v2.3. However, treebanks contain minor changes from version to version.}, which use ELMo as well as word embeddings. Additionally, \newcite{che-etal-2018-towards} is trained on POS and dependency parsing jointly while we trained mBERT to perform each task separately. As a result, the dependency parsing with mBERT does not have access to POS tags. By comparing mBERT to these baselines, we control for task and language-specific supervised training set size.

\subsection{Masked Language Model Pretraining}
We include several experiments in which we pretrain BERT from scratch. We use the PyTorch \cite{paszke2019pytorch} implementation by \newcite{lample2019cross}, the same as \autoref{sec:pretraining}. All sentences in the corpus are concatenated. For each language, we sample a batch of $N$ sequence and each sequence contains $M$ tokens, ignoring sentence boundaries. When considering two languages, we sample each language uniformly. We then randomly select 15\% of the input tokens for masking, proportionally to the exponentiated token count of power -0.5, favoring rare tokens. We replace selected masked token with \texttt{<MASK>} 80\% of the time, the original token 10\% of the time, and uniform random token within the vocabulary 10\% of the time. The model is trained to recover the original token \cite{devlin-etal-2019-bert}. We drop the next sentence prediction task as \newcite{liu2019roberta} find it does not improve downstream performance.

\subsubsection{Data Processing}
We extract text from a Wikipedia dump with Gensim \cite{rehurek_lrec}. We learn vocabulary for the corpus using SentencePiece \cite{kudo-richardson-2018-sentencepiece} with the unigram language model \cite{kudo-2018-subword}. When considering two languages, we concatenate the corpora for the two languages while sampling the same number of sentences from both corpora when learning vocabulary. We learn a vocabulary of size $V$, excluding special tokens. Finally, we tokenized the corpora using the learned SentencePiece model and did not apply any further preprocessing.

\subsubsection{BERT Models} Following mBERT, We use 12 Transformer layers \cite{vaswani2017attention} with 12 heads, embedding dimensions of 768, hidden dimension of the feed-forward layer of 3072, dropout of 0.1 and GELU activation \cite{hendrycks2016bridging}. We tie the output softmax layer and input embeddings \cite{press-wolf-2017-using}. We consider both a 12 layer model (\textbf{base}) and a smaller 6 layer model (\textbf{small}).

\subsubsection{BERT Optimization} We train BERT with Adam and an inverse square root learning rate scheduler with warmup \cite{vaswani2017attention}. We warm up linearly for 10k steps and the learning rate is 0.0001. We use batch size $N=88$ and mixed-precision training. We trained the model for roughly 115k steps and save a checkpoint every 23k steps, which corresponds to 10 epochs. We select the best out of five checkpoints with a task-specific dev set. We train each model on a single NVIDIA RTX Titan with 24GB of memory for roughly 20 hours.

\section{Are All Languages Created Equal in mBERT?}\label{sec:is-mbert-equal}

\begin{figure*}[]
\centering
\includegraphics[width=\columnwidth]{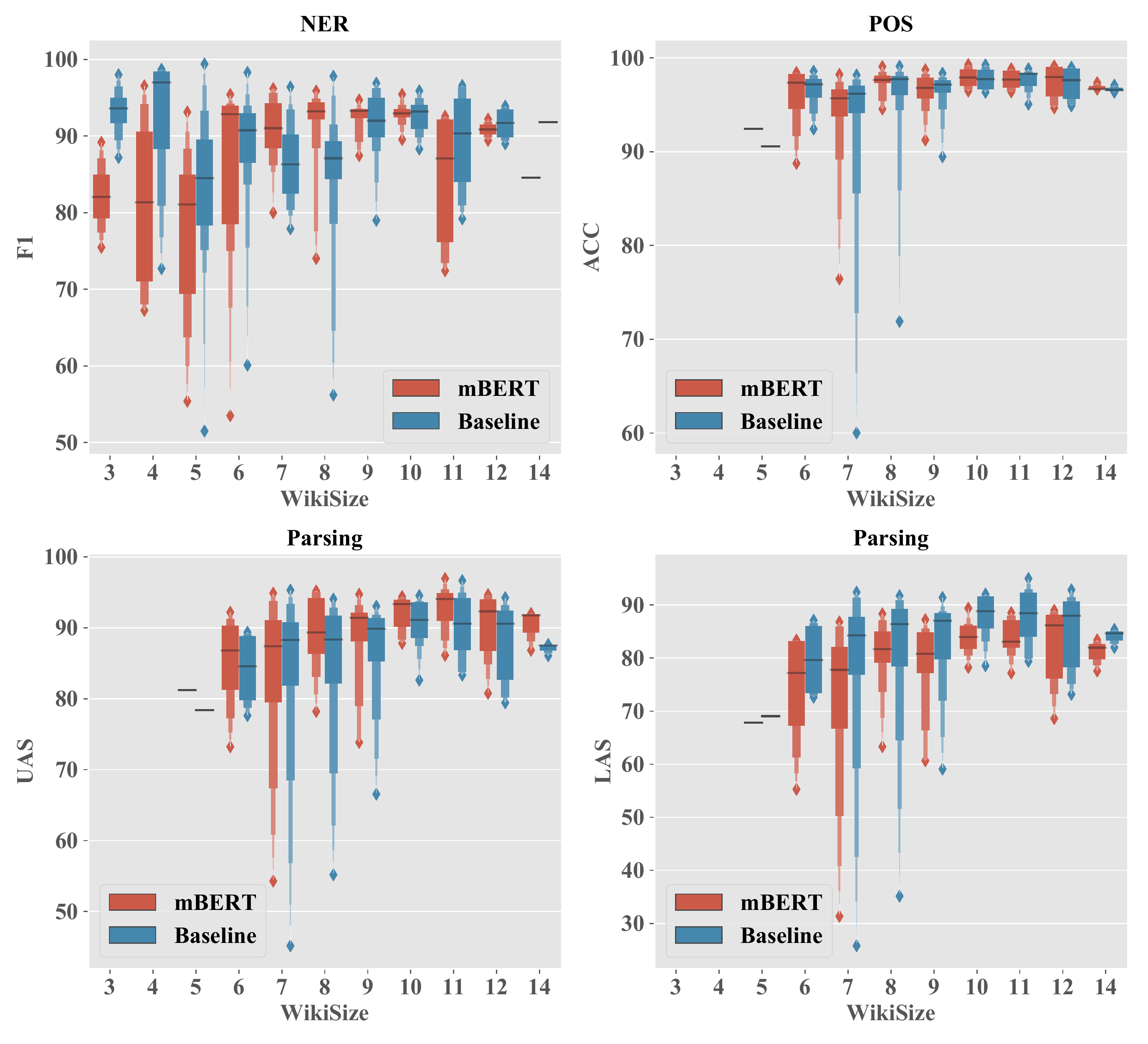}
\caption{mBERT vs baseline grouped by WikiSize. mBERT performance drops much more than baseline models on languages lower than WikiSize 6 -- the bottom 30\% languages supported by mBERT -- especially in NER, which covers nearly all mBERT supported languages.
}
\label{fig:baseline}
\end{figure*}

\begin{figure}[]
\centering
\includegraphics[width=0.6\columnwidth]{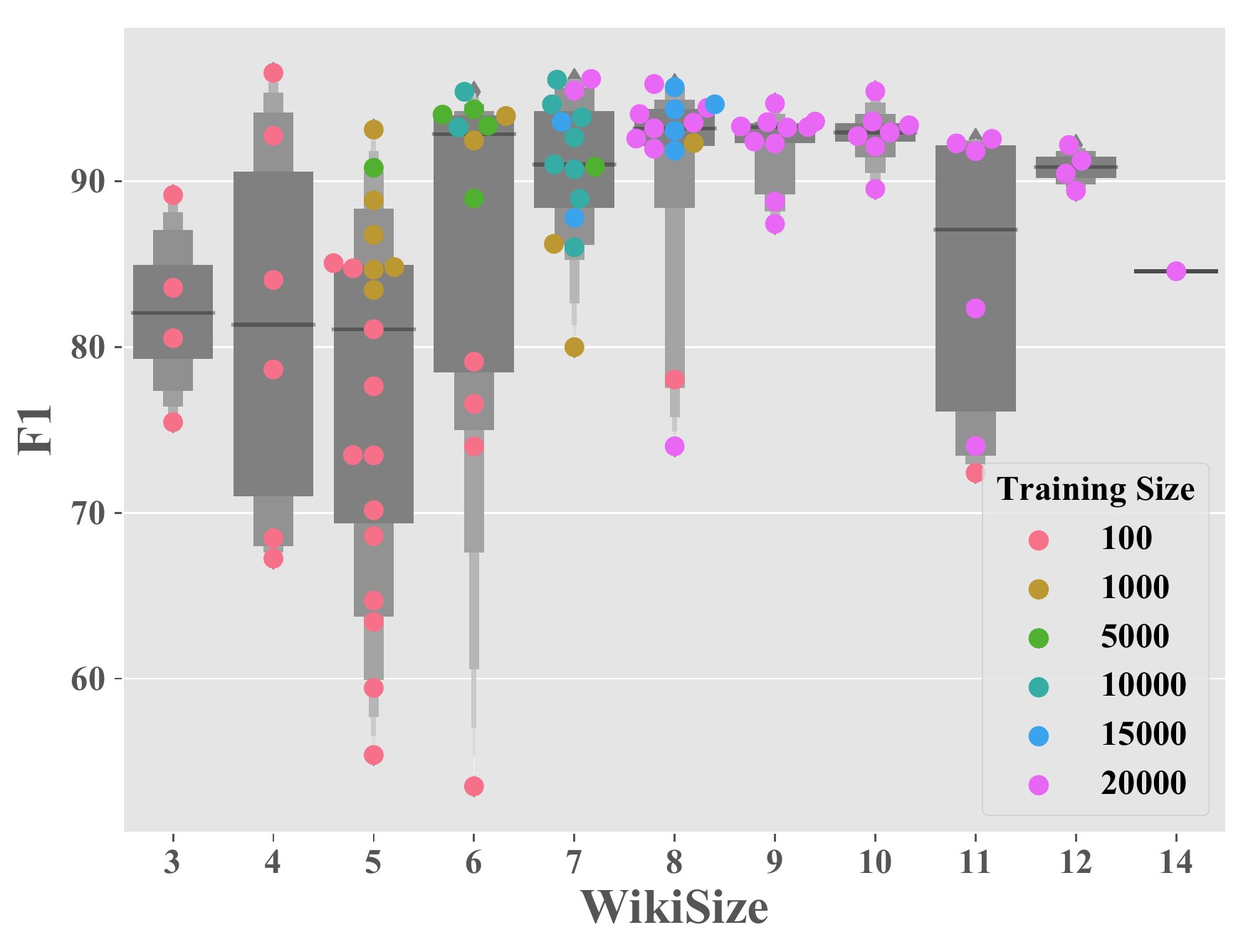}
\caption{NER with mBERT on 99 languages, ordered by size of pretraining corpus (WikiSize). Task-specific supervised training size differs by language. Performance drops dramatically with less pretraining and supervised training data.}
\label{fig:ner-only}
\end{figure}

\autoref{fig:baseline} shows the performance of mBERT and the baseline averaged across all languages by Wikipedia size (see \autoref{tab:low-lang} for groupings). 
For WikiSize over 6, mBERT is comparable or better than baselines in all three tasks, with the exception of NER. For NER in very high resource languages (WikiSize over 11, i.e. top 10\%) mBERT performs worse than baseline, suggesting high resource languages could benefit from monolingual pretraining. Note mBERT has strong UAS on parsing but weak LAS compared to the baseline; in \autoref{sec:is-mbert-multilingual} we find adding POS to mBERT improves LAS significantly. We expect multitask learning on POS and parsing could further improve LAS.
While POS and Parsing only cover half (54) of the languages, NER covers 99 of 104 languages, extending the curve to the lowest resource languages. mBERT performance drops significantly for languages with WikiSize less than 6 (bottom 30\% languages). For the smallest size, mBERT goes from being competitive with state-of-the-art to being {\em over 10 points behind.} Readers may find this surprising since while these are very low resource languages, mBERT training up-weighted these languages to counter this effect.

\autoref{fig:ner-only} shows the performance of mBERT (only) for NER over languages with {\em different resources}, where we show how much task-specific supervised training data was available for each language. For languages with only 100 labeled sentences, the performance of mBERT drops significantly as these languages also had less pretraining data.
While we may expect that pretraining representations with mBERT would be most beneficial for languages with only 100 labels, as \newcite{howard-ruder-2018-universal} show pretraining improve data-efficiency for English on text classification, our results show that on low resource languages this strategy performs much worse than a model trained directly on the available task data.
Clearly, mBERT provides variable quality representations depending on the language. While we confirm the finding of others that mBERT is excellent for high resource languages, it is much worse for low resource languages.
Our results suggest caution for those expecting a reliable model for \textit{all} 104 mBERT languages.

\section{Why Are All Languages Not Created Equal in mBERT?}\label{sec:why-not-equal}

\subsection{Statistical Analysis}\label{sec:stat-test}

\input{table/low/sigtest}

We present a statistical analysis to understand why mBERT does so poorly in some languages.
We consider three factors that might affect the downstream task performance: pretraining Wikipedia size (WikiSize), task-specific supervision size, and vocabulary size in task-specific data. Note we take $\log_2$ of training size and training vocab following WikiSize.
We consider NER because it covers nearly all languages of mBERT.

We fit a linear model to predict task performance (F1) using a single factor. \autoref{tab:low-sigtest} shows that each factor has a statistically significant positive correlation. One unit increase of training size leads to the biggest performance increase, then training vocabulary followed by WikiSize, all in log scale. 
Intuitively, training size and training vocab correlate with each other. We confirm this with a log-likelihood ratio test; adding training vocabulary to a linear model with training size yields a statistically insignificant improvement. As a result, when considering multiple factors, we consider training size and WikiSize. Interestingly, \autoref{tab:low-sigtest} shows training size still has a positive but slightly smaller slope, but the
slope of WikiSize change sign, which suggests WikiSize might correlate with training size. We confirm this by fitting a linear model with training size as $x$ and WikiSize as $y$ and the slope is over 0.5 with
$p<0.001$. This finding is unsurprising as the NER dataset is built from Wikipedia so larger Wikipedia size means larger training size.

In conclusion, the larger the task-specific supervised dataset, the better the downstream performance on NER. Unsurprisingly, while pretraining improve data-efficiency \cite{howard-ruder-2018-universal}, it still cannot solve a task with limited supervision.
Training vocabulary and Wikipedia size correlate with training size, and increasing either one factor leads to better performance. A similar conclusion could be found when we try to predict the performance ratio of mBERT and the baseline instead.
Statistical analysis shows a correlation between resource and mBERT performance but can not give a causal answer on why low resource languages within mBERT perform poorly.

\subsection{mBERT vs monolingual BERT}
\label{sec:monolingual-bert}

\input{table/low/langstat}

We have established that mBERT does not perform well in low-resource languages. Is this because we are relying on a multilingual model that favors high-resource over low-resource languages? To answer this question we train monolingual BERT models on several low resource languages with different hyperparameters. Since pretraining a BERT model from scratch is computationally intensive, we select four low resource languages: Latvian (lv), Afrikaans (af), Mongolian (mn), and Yoruba (yo). These four languages (bold font in \autoref{tab:low-langstat}) reflect varying amounts of monolingual training data. 

\begin{figure*}[]
\centering
\includegraphics[width=\columnwidth]{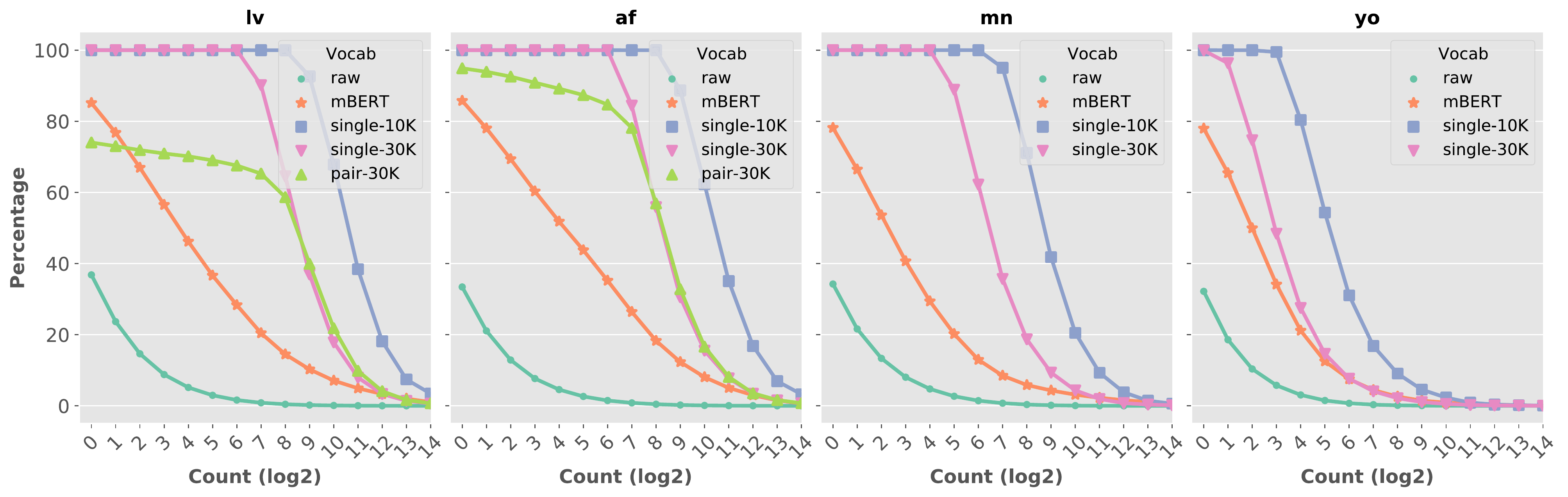}
\caption{Percentage of vocabulary containing word count larger than a threshold. ``Raw'' is the vocabulary segmented by space. Single-30K and Single-10K are 30K/10K vocabularies learned from single languages. Pair-30K is 30K vocabulary learned from the selected language and a closely related language, described in \autoref{sec:bilingual-bert}.}
\label{fig:vocab}
\end{figure*}

It turns out that these low resource languages are reasonably covered by mBERT's vocabulary: 25\% to 50\% of the subword types within the mBERT 115K vocabulary appear in these languages' Wikipedia. However, the mBERT vocabulary is by no means optimal for these languages. \autoref{fig:vocab} shows that a large amount of the mBERT vocabulary that appears in these languages is low frequency while the language-specific SentencePiece vocabulary has a much higher frequency. In other words, the vocabulary of mBERT is not distributed uniformly.

\begin{figure*}[]
\centering
\includegraphics[width=\columnwidth]{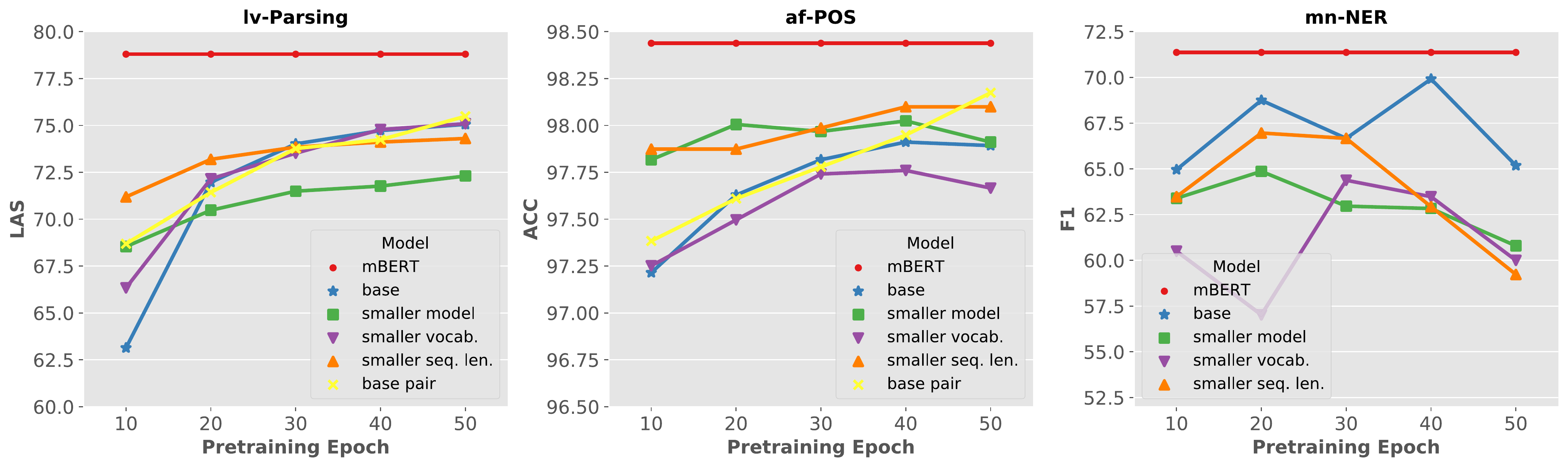}
\caption{Dev performance with different pretraining epochs on three languages and tasks. Dev performance on higher resources languages (lv, af) improves as training continues, while lower resource languages (mn) fluctuate.}
\label{fig:epoch}
\end{figure*}

To train the monolingual BERTs properly for low resource languages, we consider four different sets of hyperparameters. In \textbf{base}, we follow English monolingual BERT on learning vocabulary size $V=30K$, 12 layers of transformer (base). To ensure we have a reasonable batch size for training using our GPU, we set the training sequence length to $M=256$.
Since a smaller model can prevent overfitting smaller datasets, we consider 6 transformer layers (\textbf{small}). We do not change the batch size as a larger batch is observed to improve performance \cite{liu2019roberta}.
As low resource languages have small corpora, 30K vocabulary items might not be optimal. We consider \textbf{smaller vocabulary} with $V=10K$.
Finally, since in fine-tuning we only use a maximum sequence length of 128, in \textbf{smaller sequence length}, we match the fine-tuning phrase with $M=128$. As a benefit of half the self-attention range, we can increase the batch size over 2.5 times to $N=220$.

\input{table/low/main}

\autoref{tab:low-main} shows the performance of monolingual BERT in four settings.
The model with smaller sequence length performs best for monolingual BERT and outperforms the base model in 5 out of 8 tasks and languages combination.
The model with smaller vocabulary has mixed performance in the low resource languages (mn, yo) but falls short for (relatively) higher resource languages (lv, af).
Finally, the smaller model underperforms the base model in 5 out of 8 cases.
In conclusion, the best way to pretrain BERT with a limited amount of computation for low resource languages is to use a smaller sequence length to allow a larger batch size.

Despite these insights, no monolingual BERT outperforms mBERT (except Latvian POS). For higher resource languages (lv, af) we hypothesize that training longer with larger batch size could further improve the downstream performance as the cloze task dev perplexity was still improving. \autoref{fig:epoch} supports this hypothesis showing downstream dev performance of lv and af improves as pretraining continues. Yet for lower resource languages (mn, yo), the cloze task dev perplexity is stuck and we began to overfit the training set. At the same time, \autoref{fig:epoch} shows the downstream performance of mn fluctuates. It suggests the cloze task dev perplexity correlates with downstream performance when dev perplexity is not decreasing.

The fact that monolingual BERT underperforms mBERT on four low resource languages suggests that mBERT style multilingual training benefits low resource languages by transferring from other languages; monolingual training produces worse representations due to small corpus size. Additionally, the poor performance of mBERT on low resource languages does not emerge from balancing between languages. Instead, it appears that we do not have sufficient data, or the model is not sufficiently data-efficient.

\subsection{mBERT vs Bilingual BERT} \label{sec:bilingual-bert}
Finally, we consider a middle ground between monolingual training and massively multilingual training. We train a BERT model on a low resource language (lv and af) paired with a related higher resource language. We pair Lithuanian (lt) with Latvian and Dutch (nl) with Afrikaans.\footnote{We did not consider mn and yo since neither has a closely related language in mBERT.} Lithuanian has a similar size to Latvian while Dutch is over 10 times bigger. Lithuanian belongs to the same Genus as Latvian while Afrikaans is a daughter language of Dutch. The \textbf{base pair} model has the same hyperparameters as the base model.

\autoref{tab:low-main} shows that pairing low resource languages with closely related languages improves downstream performance. The Afrikaans-Dutch BERT improves more compared to Latvian-Lithuanian, possibly because Dutch is much larger than Afrikaans, as compared to Latvian and Lithuanian. These experiments suggest that pairing linguistically related languages can benefit representation learning and adding extra languages can further improve the performance as demonstrated by mBERT. It echoes the finding of \newcite{lample2019cross} where multilingual training improves uni-directional language model perplexity for low resource languages. Concurrent to the publication of this chapter, \newcite{conneau-etal-2020-unsupervised} shows similar findings as the performance of low resource languages (Urdu and Swahili) improves on XNLI when more languages are trained jointly then decrease with an increasing number of languages. However, they do not consider the effect of language similarity.

\section{Discussion}
While mBERT covers 104 languages, in this chapter, we find that the 30\% languages with least pretraining resources perform worse than using no pretrained language model at all. Therefore, we caution against using mBERT alone for low resource languages.
Furthermore, training a monolingual model on low resource languages does no better. Training on pairs of closely related low resource languages helps but still lags behind mBERT. Thus, mBERT is trying its best to learn representation for low resource languages. However, constrained by the sample inefficiency of BERT objective and the lack of data of low resource languages, mBERT learns low quality representation for low resource languages.
On the other end of the spectrum, the highest resource languages (top 10\%) are hurt by massively multilingual joint training. While mBERT has access to numerous languages, the resulting model is worse than a monolingual model when sufficient training data exists.

Our findings suggest, with small monolingual corpus, BERT does not learn high-quality representation for low resource languages. To learn better representation for low resource languages, we suggest either collect more data to make low resource language high resource, which leads to XLM-R \cite{conneau-etal-2020-unsupervised}, or consider more data-efficient pretraining techniques like \newcite{clark2020electra}, which leads to better performing XLM-E \cite{chi2021xlm}.
On the other hand, for high resource languages, training a monolingual model is likely to produce better representation than mBERT. Since English BERT and multilingual BERT, a large number of BERT-like models for various languages have been publicly available, e.g. Dutch \cite{delobelle-etal-2020-robbert}, French \cite{martin-etal-2020-camembert}, and Vietnamese \cite{nguyen-tuan-nguyen-2020-phobert}. In fact, by November 2021, over 1700 BERT-like models are available with the Transformers library \cite{wolf-etal-2020-transformers}.

%% file: table/low/wikirank.tex
\begin{table*}[]
\begin{center}
\resizebox{1\linewidth}{!}{
\begin{tabular}[b]{cccc}
\toprule
WikiSize & Languages & \# Languages & Size Range (GB) \\
\midrule
3 & io, pms, scn, \textbf{yo} & 4 & [0.006, 0.011] \\
4 & cv, lmo, mg, min, su, vo & 6 & [0.011, 0.022] \\
5 & an, bar, br, ce, fy, ga, gu, is, jv, ky, lb, \textbf{mn}, my, nds, ne, pa, pnb, sw, tg & 19 & [0.022, 0.044] \\
6 & \textbf{af}, ba, cy, kn, la, mr, oc, sco, sq, tl, tt, uz & 12 & [0.044, 0.088] \\
7 & az, bn, bs, eu, hi, ka, kk, lt, \textbf{lv}, mk, ml, nn, ta, te, ur & 15 & [0.088, 0.177] \\
8 & ast, be, bg, da, el, et, gl, hr, hy, ms, sh, sk, sl, th, war & 15 & [0.177, 0.354] \\
9 & fa, fi, he, id, ko, no, ro, sr, tr, vi & 10 & [0.354, 0.707] \\
10 & ar, ca, cs, hu, nl, sv, uk & 7 & [0.707, 1.414] \\
11 & ceb, it, ja, pl, pt, zh & 6 & [1.414, 2.828] \\
12 & de, es, fr, ru & 4 & [2.828, 5.657] \\
14 & en & 1 & [11.314, 22.627] \\
\bottomrule
\end{tabular}
}
\caption{List of 99 languages we consider in mBERT and its pretraining corpus size\label{tab:low-lang}. Languages in \textbf{bold} are the languages we consider in \autoref{sec:why-not-equal}.}
\end{center}
\end{table*}

%% file: table/low/sigtest.tex
\begin{table}[]
\begin{center}
\begin{tabular}[b]{cccc}
\toprule
& Coefficient &	p-value & CI \\
\midrule
\multicolumn{4}{l}{\textit{Univariate}} \\
\midrule
Training Size & 0.035 & $<$0.001 & [0.029, 0.041] \\
Training Vocab & 0.021 & $<$0.001 & [0.017, 0.025] \\
WikiSize & 0.015 & $<$0.001 & [0.007, 0.023] \\
\midrule 
\multicolumn{4}{l}{\textit{Multivariate}} \\
\midrule
Training Size & 0.029 & $<$0.001 & [0.023, 0.035] \\
WikiSize & -0.014 & $<$0.001 & [-0.022, -0.006] \\
\bottomrule
\end{tabular}
\caption{Statistical analysis on what factors predict downstream performance. We fit two types of linear models, which consider either single factor or multiple factors.
\label{tab:low-sigtest}}
\end{center}
\end{table}

%% file: table/low/langstat.tex
\begin{table}[]
\begin{center}
\begin{tabular}[b]{c cccc}
\toprule
& lv & af & mn & yo \\
\midrule
Genus & Baltic & Germanic & Mongolic & Defoid \\
Family & Indo-Eur  & Indo-Eur  & Altaic & Niger-Congo \\
WikiSize & 7 & 6 & 5 & 3 \\
\# Sentences (M) & 2.9 & 2.3 & 0.8 & 0.1 \\
\# Tokens (M) & 21.8 & 28.8 & 6.4 & 0.9 \\
mBERT vocab (K) & 56.6 & 59.0 & 42.3 & 29.3 \\
mBERT vocab (\%) & 49.2 & 51.3 & 36.8 & 25.5 \\
\bottomrule
\end{tabular}
\caption{Statistic of four low resource languages.
\label{tab:low-langstat}}
\end{center}
\end{table}

%% file: table/low/main.tex
\begin{table*}[]
\begin{center}
\resizebox{1\linewidth}{!}{
\begin{tabular}[b]{ccc| ccc| ccc| c| c}
\toprule
\multirow{2}{*}{Model Size} & \multirow{2}{*}{Vocabulary} & \multirow{2}{*}{Max Length} & \multicolumn{3}{c|}{lv} & \multicolumn{3}{c|}{af} & \multicolumn{1}{c|}{mn} & \multicolumn{1}{|c}{yo} \\
& & & NER & POS & Parsing (LAS/UAS) & NER & POS & Parsing (LAS/UAS) & NER & NER \\
\midrule 
\multicolumn{11}{l}{\textit{Baseline}} \\
\midrule
\multicolumn{3}{c|}{Baseline} & 92.10 & \textbf{96.19} & \textbf{84.47}/88.28 & \textbf{94.00} & 97.50 & \textbf{85.69}/88.67 & \textbf{76.40} & \textbf{94.00} \\
\multicolumn{3}{c|}{mBERT} & \textbf{93.88} & 95.69 & 77.78/\textbf{88.69} & 93.36 & \textbf{98.26} & 83.18/\textbf{89.69} & 64.71 & 80.54 \\
\midrule 
\multicolumn{11}{l}{\textit{Monolingual BERT} (\autoref{sec:monolingual-bert})} \\
\midrule
base & 30k & 256  & 93.02 & \underline{95.76} & \underline{74.18}/\underline{85.35} & 90.90 & 97.76 & 80.08/86.92 & 56.20 & 72.57 \\
\midrule
small & - & - & 92.75 & 95.41 & 71.67/83.34 & 90.67 & \underline{98.02} & 80.60/87.40 & \underline{58.92} & 70.80 \\
- & 10k & - & 92.68 & 95.65 & 73.94/85.20 & 89.55 & 97.66 & 79.91/86.93 & 41.70 & \underline{80.18} \\
- & - & 128 & \underline{93.38} & 95.57 & 73.21/84.53 & \underline{91.84} & 97.87 & \underline{80.83}/\underline{87.59} & 55.91 & 73.45 \\
\midrule 
\multicolumn{3}{l}{\textit{Bilingual BERT} (\autoref{sec:bilingual-bert})} & \multicolumn{3}{c}{lv + lt} & \multicolumn{3}{c}{af + nl} & \multicolumn{2}{c}{} \\
\midrule
base & 30k & 256  & 93.22 & 96.03 & 74.42/85.60 & 91.85 & 97.98 & 81.73/88.55 & n/a & n/a  \\
\bottomrule
\end{tabular}
}
\caption{Monolingual BERT on four languages with different hyperparameters. \underline{Underscore} denotes best within monolingual BERT and \textbf{bold} denotes best among all models. Monolingual BERT underperforms mBERT in most cases. ``-'' denotes same as base case. \label{tab:low-main}}
\end{center}
\end{table*}

%% file: src/crosslingual-signal.tex
\section{Introduction}

Massively multilingual encoders including multilingual BERT \citep[mBERT]{devlin-etal-2019-bert} and XLM-RoBERTa \citep[XLM-R]{conneau-etal-2020-unsupervised} are pretrained without any explicit cross-lingual signal. In this chapter, we will investigate how to inject two types of cross-lingual signal into multilingual encoders: bilingual dictionary and bitext.

Bilingual dictionary is widely available for most language pairs, and it is easy to collect bilingual dictionary for a new language pair \cite{kamholz-etal-2014-panlex}. We inject it into the pretraining process by increasing subwords overlap across languages. We achieve additional subwords overlap by creating synthetic code-switching corpus with bilingual dictionary.  As we observe in \autoref{sec:mbert-corr}, subwords overlap between languages correlates with cross-lingual transfer performance, although \autoref{sec:anchor-points} shows that subword overlap is not the necessary condition for cross-lingual representation. In \autoref{sec:bilingual-dict}, we show that the correlation indeed holds with additional subwords overlap, in other word, having extra anchor points benefit the cross-lingual representation.

Bitext is available for most high-resource language pairs (usually involving English), and researchers have proposed collecting additional bitext by mining parallel sentences from the Web \cite{schwenk-etal-2021-wikimatrix,schwenk2019ccmatrix}. While bitext can be incorporated during expensive pretraining \citep{lample2019cross,huang-etal-2019-unicoder,ji2019cross,chi-etal-2021-infoxlm}, aligning pretrained multilingual encoders with explicit alignment objective, i.e. enforcing similar words from different languages have similar representation, is much more efficient. However, as word-level alignments from an unsupervised aligner are often suboptimal, in \autoref{sec:bitext}, we develop a new cross-lingual alignment objective for training our model.
We base our objective on contrastive learning, in which two similar inputs -- such as from a bitext -- are directly optimized to be similar, relative to a negative set. These methods have been effective in computer vision tasks \citep{he2019momentum,chen2020simple}.

Most previous work on contextual alignments consider high-quality bitext like Europarl \citep{koehn2005europarl} or MultiUN \citep{eisele-chen-2010-multiun}. While helpful, these resources are unavailable for most languages for which we seek a zero-shot transfer. To better reflect the quality of bitext available for most languages, we additionally use OPUS-100 \citep{zhang-etal-2020-improving}, a randomly sampled 1 million subset (per language pair) of the OPUS collection \citep{tiedemann-2012-parallel}.
In \autoref{sec:bitext}, we show that our new contrastive learning alignment objectives outperform previous work \citep{cao2020multilingual} when applied to bitext from previous works or the OPUS-100 bitext. However, our experiments also produce a negative result. While previous work showed improvements from alignment-based objectives on zero-shot cross-lingual transfer for a single task (XNLI) with a single random seed, our more extensive analysis tells a different story. We report the mean and standard deviation of multiple runs with the same hyperparameters and different random seeds. We find that previously reported improvements disappear, even while our new method shows a small improvement. Furthermore, we extend the evaluation to multiple languages on 4 tasks, further supporting our conclusions.

\section{Background}

\subsection{Explicit Alignment Objectives}\label{sec:previous-alignment}
We begin with a presentation of explicit alignment objective functions that use parallel data across languages for training multilingual encoders.
These objectives assume multilingual data in the form of word pairs in parallel sentences. Since gold word alignments are scarce, we use an unsupervised word aligner. Let \src and \trg be the contextual hidden state matrix of corresponding words from a pretrained multilingual encoder. We assume \src is English while \trg is a combination of different target languages. As both mBERT and XLM-R operate at the subword level, we use the representation of the first subword, which is consistent with the evaluation stage. Each $s_i$ and $t_i$ are a corresponding row of \src and $\mathbf{T}$, respectively. \src and \trg come from the final layer of the encoder while $\mathbf{S}^{l}$ and $\mathbf{T}^{l}$ come from the $l^{\text{th}}$-layer.

\subsubsection{Linear Mapping} \label{sec:linear-alignment}
If \src and \trg are static feature (such as from ELMo \citep{peters-etal-2018-deep}) then \trg can be aligned so that it is close to \src via a linear mapping \citep{wang-etal-2019-cross,wang2019cross,liu-etal-2019-investigating,conneau-etal-2020-emerging}, similar to aligning monolingual embeddings to produce cross-lingual embeddings. For feature $\mathbf{S}^{l}$ and $\mathbf{T}^{l}$ from layer $l$, we can learn a mapping $\mathbf{W}^{l}$.
\begin{align}\label{eq:linear}
\mathbf{W}^{l*} = \arg\min_{\mathbf{W}^{l}} \| \mathbf{S}^{l} - \mathbf{T}^{l}\mathbf{W}^{l} \|^2_2 
\end{align}
When $\mathbf{W}^{l}$ is orthogonal, \autoref{eq:linear} is known as Procrustes problem \citep{smith2017offline} and can be solved by SVD. Alternatively, \autoref{eq:linear} can also be solved by gradient descent, without the need to store in memory huge matrices \src and $\mathbf{T}$. We adopt the latter more memory efficient approach. Following \citet{conneau2017word}, we enforce the orthogonality by alternating the gradient update and the following update rule
\begin{align}
\mathbf{W} \leftarrow (1+\beta) \mathbf{W} - \beta (\mathbf{W}\mathbf{W}^T)\mathbf{W}
\end{align}
with $\beta = 0.01$. Note we learn different $\mathbf{W}^{l}$ for each target language.

This approach has yielded improvements in several studies.
In \autoref{sec:align-contextual-word}, we use bilingual BERT and 10k parallel sentences from XNLI \citep{conneau-etal-2018-xnli} to improve dependency parsing (but not NER) on French, Russian, and Chinese. 
\citet{wang-etal-2019-cross} use mBERT and 10k parallel sentences from Europarl to improve dependency parsing.
\citet{wang2019cross} use mBERT and 30k parallel sentences from Europarl to improve named entity recognition (NER) on Spanish, Dutch, and German.
\citet{liu-etal-2019-investigating} do not evaluate on cross-lingual transfer tasks.

\subsubsection{L2 Alignment} \label{sec:l2-alignment}
Instead of using \src and \trg as static features, \citet{cao2020multilingual} propose fine-tuning the entire encoder
\begin{align}\label{eq:l2}
\mathcal{L}_\text{L2} (\theta) = \text{mean}_i( \| s_i - t_i \|^2_2 )
\end{align}
where $\theta$ is the encoder parameters. 
To prevent a degenerative solution, they additionally use a regularization term
\begin{align}\label{eq:src-hidden}
\mathcal{L}_\text{reg-hidden} (\theta) = \| \bar{\mathbf{S}} - \bar{\mathbf{S}}_\text{pretrained} \|^2_2
\end{align}
where $\bar{\mathbf{S}}$ denote \textbf{all} hidden states of the source sentence including unaligned words, encouraging the source hidden states to stay close to the pretrained hidden states. With mBERT and 20k to 250k parallel sentences from Europarl and MultiUN, \citeauthor{cao2020multilingual} show improvement on XNLI but not parsing.\footnote{The authors state they did not observe improvements on parsing in the NLP Hightlights podcast (\#112) \citep{ai2_2020}.}

In preliminary experiments, we found constraining parameters to stay close to their original pretrained values also prevents degenerative solutions
\begin{align}\label{eq:src-param}
\mathcal{L}_\text{reg-param} (\theta) = \| \theta - \theta_\text{pretrained} \|^2_2
\end{align}
while being more efficient than \autoref{eq:src-hidden}. As a result, we adopt the following objective (with $\lambda=1$):
\begin{align}\label{eq:l2_and_reg}
\mathcal{L} (\theta) = \mathcal{L}_\text{L2} (\theta) + \lambda \mathcal{L}_\text{reg-param} (\theta)
\end{align}

\section{Bilingual Dictionary}\label{sec:bilingual-dict}

\subsection{Experiments}

As \autoref{sec:mbert-corr} suggests that there may be correlation between cross-lingual performance and anchor points,
we additionally increase anchor points by using bilingual dictionary to create code switch data for training bilingual MLM. Specifically, for two languages, $\ell_1$ and $\ell_2$, with bilingual dictionary entries $d_{\ell_1, \ell_2}$, we add anchors to the training data as follows. 
For each training word $w_{\ell_1}$ in the bilingual dictionary, we either leave it as is (70\% of the time) or randomly replace it with one of the possible translations from the dictionary (30\% of the time).
We change at most 15\% of the words in a batch and sample word translations from  PanLex~\cite{kamholz-etal-2014-panlex} bilingual dictionary, weighted according to their translation quality.\footnote{Although we only consider pairs of languages, this procedure naturally scales to multiple languages.}
We pretrain two bilingual encoders for each language pair: with or without synthetic code-switching corpus. We consider the same three language pairs as \autoref{sec:mbert-ablation}: English-French, English-Russian, and English-Chinese. The rest of the pretraining is the same as \autoref{sec:pretraining}. Recall that each encoder is a 8-layer Transformer. To ensure a fair comparison, both models have the same number of gradient updates. For this section, we adapt the same zero-shot cross-lingual evaluation on XNLI, NER, and dependency parsing as \autoref{sec:emerging-eval}

\subsection{Fingdings}

\input{table/signal/dict}

\autoref{tab:signal-dict} shows using bilingual dictionary to create synthetic code-switching corpus overall benefit cross-lingual representation. Anchor points have a clear effect on performance and more anchor points help, especially in the less closely related language pairs (e.g. English-Chinese has a larger effect than English-French with over 3 points improvement on NER and XNLI).

\section{Bitext}\label{sec:bitext}

\begin{figure*}[]
\begin{center}
\resizebox{\linewidth}{!}{\includegraphics{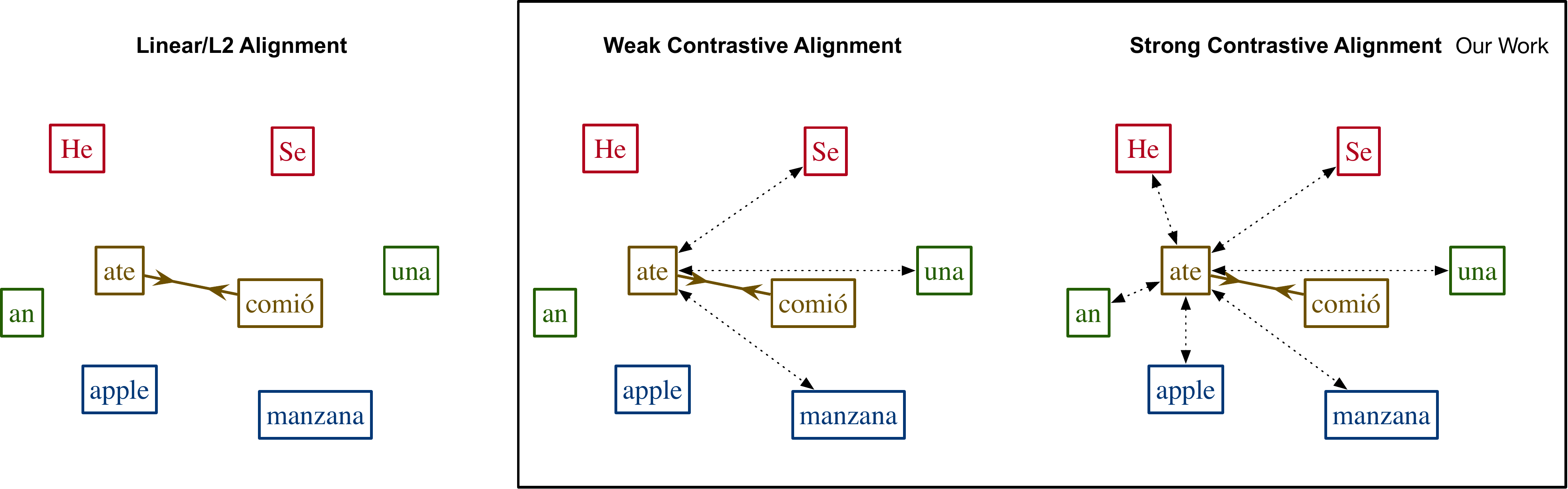}}
\caption{Explicit alignment with different objectives. We use a parallel sentence ``He ate an apple'' and ``Se comió una manzana'' as an example. While linear or L2 alignment optimizes for absolute distance, making ``ate'' and ``comió'' as close as possible (solid line), contrastive alignment optimizes for relative distance, making ``ate'' and ``comió'' closer (solid line) and pushing other away (dotted line).}
\label{fig:mapping}
\end{center}
\end{figure*}

\subsection{Contrastive Alignment}
Inspired by the contrastive learning framework of \citet{chen2020simple}, we propose a contrastive loss to align \src and \trg by fine-tuning the encoder. Assume in each batch, we have corresponding $(s_i, t_i)$ where $i \in \{1,\dots,B\}$.
Instead of optimizing the absolute distance between $s_i$ and $t_i$ like \autoref{eq:linear} or \autoref{eq:l2} in \autoref{sec:previous-alignment}, contrastive loss allows more flexibility by encouraging $s_i$ and $t_i$ to be closer as compared with any other hidden state. In other words, our proposed contrastive alignment optimizes the relative distance between $s_i$ and $t_i$ (see \autoref{fig:mapping} for visualization). As the alignment signal is often suboptimal, our alignment objective is more robust to errors in unsupervised word-level alignment. Additionally, unlike previous works, we select different sets of negative examples to enforce different levels of cross-lingual alignment. Finally, it naturally scales to multiple languages.

\subsubsection{Weak alignment} 
When the negative examples only come from target languages, we enforce a weak cross-lingual alignment, i.e. $s_i$ should be closer to $t_i$ than any other $t_j, \forall j\neq i$. The same is true in the other direction. The loss of a batch is
\begin{align}
\mathcal{L}_\text{weak} (\theta) = \frac{1}{2B} \sum^{B}_{i=1} (\log\frac{ \exp (\text{sim}(s_i, t_i ) / T) } { \sum^{B}_{j=1} \exp (\text{sim}(s_i, t_j ) / T) } + \log\frac{ \exp (\text{sim}(s_i, t_i ) / T) } { \sum^{B}_{j=1} \exp (\text{sim}(s_j, t_i ) / T) } )
\end{align}
where $T = 0.1$ is a temperature hyperparameter and $\text{sim}(a, b)$ measures the similarity of $a$ and $b$.

We use a learned cosine similarity $\text{sim}(a, b) = \cos(f(a), f(b))$ where $f$ is a feed-forward feature extractor with one hidden layer (768-768-128) and ReLU. It can learn to discard language-specific information and only align the align-able information. \citet{chen2020simple} find that this similarity measure learns better representation for computer vision. After alignment, $f$ is discarded as most cross-lingual transfer tasks do not need this feature extractor, though tasks like parallel sentence retrieval might find it helpful. This learned similarity cannot be applied to an absolute distance objective like \autoref{eq:l2} as it can produce degenerate solutions.

\subsubsection{Strong alignment} 
If the negative examples include both source and target languages, we enforce a strong cross-lingual alignment, i.e. $s_i$ should be closer to $t_i$ than any other $t_j, \forall j \neq i$ and $s_j, \forall j \neq i$.
\begin{align}
\mathcal{L}_\text{strong} (\theta) = \frac{1}{2B} \sum_{h\in \mathcal{H}} \log\frac{ \exp (\text{sim}(h, \text{aligned}(h) ) / T) } { \sum_{h'\in \mathcal{H}, h'\neq h} \exp (\text{sim}(h, h') / T) }
\end{align}
where $\text{aligned}(h)$ is the aligned hidden state of $h$ and $\mathcal{H} = \{s_1, \dots, s_B, t_1, \dots, t_B\}$.

For both weak and strong alignment objectives, we add a regularization term \autoref{eq:src-param} with $\lambda = 1$.

\subsection{Experiments}

\subsubsection{Multilingual Alignment} We consider alignment and transfer from English to 8 target languages: Arabic, German, English, Spanish, French, Hindi, Russian, Vietnamese, and Chinese. We use two sets of bitexts: (1) bitext used in previous works \citep{lample2019cross} and (2) the OPUS-100 bitext \citep{zhang-etal-2020-improving}. (1) For bitext used in previous works, we use MultiUN for Arabic, Spanish, French, Russian or Chinese, EUBookshop \citep{skadins-etal-2014-billions} for German, IIT Bombay corpus \citep{kunchukuttan-etal-2018-iit} for Hindi and OpenSubtitles \citep{lison-etal-2018-opensubtitles2018} for Vietnamese. We sample 1M bitext for each target language. (2) The OPUS-100 covers 100 languages with English as the center, and sampled from the OPUS collection randomly, which better reflects the average quality of bitext for most languages. It contains 1M bitext for each target language, except Hindi (0.5M).

We tokenize the bitext with Moses \citep{koehn-etal-2007-moses} and segment Chinese with \citet{chang-etal-2008-optimizing}. %
We use \texttt{fast\_align} \citep{dyer-etal-2013-simple} to produce unsupervised word alignments in both directions and symmetrize with the \textit{grow-diag-final-and} heuristic. We only keep one-to-one alignment and discard any trivial alignment where the source and target words are identical. %

We train the L2 (\autoref{sec:l2-alignment}), weak, and strong alignment objectives in a multilingual fashion. Each batch contains examples from all target languages. Following \citet{devlin-etal-2019-bert}, we optimize with Adam \citep{kingma2014adam}, learning rate $\texttt{1e-4}$, 128 batch size, 100k total steps ($\approx$ 2 epochs), 4k steps linear warmup and linear decay. We use 16-bit precision and train each model on a single RTX TITAN for around 18 hours. We set the maximum sequence length to 96. For linear mapping (\autoref{sec:linear-alignment}), we use a linear decay learning rate from $\texttt{1e-4}$ to $0$ in 20k steps ($\approx$ 3 epochs), and train for 3 hours for each language pairs.

\subsubsection{Evaluation} We consider zero-shot cross-lingual transfer with XNLI \citep{conneau-etal-2018-xnli}, NER \citep{pan-etal-2017-cross}, POS tagging and dependency parsing \citep{ud2.6}.\footnote{We use the following treebanks: Arabic-PADT, German-GSD, English-EWT, Spanish-GSD, French-GSD, Hindi-HDTB, Russian-GSD, Vietnamese-VTB, and Chinese-GSD.} We evaluate XNLI and POS tagging with accuracy (ACC), NER with span-level F1, and parsing with labeled attachment score (LAS). For the task-specific layer, we use a linear classifier for XNLI, NER, and POS tagging, and use \citet{dozat2016deep} for dependency parsing. We fine-tune all parameters on English training data and directly transfer to target languages. We optimize with Adam, learning rate $\texttt{2e-5}$ with 10\% steps linear warmup and linear decay, 5 epochs, and 32 batch size. For the linear mapping alignment, we use an ELMo-style feature-based model\footnote{We take the weighted average of representations in all layers of the encoder.} with 4 extra Transformer layers \citep{vaswani2017attention}, a CRF instead of a linear classifier for NER, and train for 20 epochs, a batch size of 128 and learning rate $\texttt{1e-3}$ (except NER and XNLI with $\texttt{1e-4}$). All token level tasks use the first subword as the word representation for task-specific layers similar to previous chapters. Model selection is done on the English dev set. We report the mean and standard derivation of test performance of 5 evaluation runs with different random seeds\footnote{We pick 5 random seeds before the experiment and use the same seeds for each task and model.} and the same hyperparameters.

We set the maximum sequence length to 128 during fine-tuning. For NER and POS tagging, we additionally use a sliding window of context to include subwords beyond the first 128. At test time, we use the same maximum sequence length except for parsing. At test time for parsing, we only use the first 128 words of a sentence instead of subwords to make sure we compare different models consistently.
We ignore words with POS tags of \texttt{SYM} and \texttt{PUNCT} during parsing evaluation.
We adopt the same post-processing heuristic steps as \autoref{sec:task-ner} during NER evaluation.
As the supervision on Chinese NER is on character-level, we segment the character into word using the Stanford Word Segmenter and realign the label.

\subsection{Findings}

\input{table/signal/main}

\input{table/signal/xlm}

\input{table/signal/opus}

\subsubsection{Robustness of Previous Methods} With a more robust evaluation scheme and 1 million parallel sentences (4$\times$ to 100$\times$ of previously considered data), the previously proposed Linear Mapping or L2 Alignment does not consistently outperform a no alignment setting more than one standard deviation in all cases (\autoref{tab:signal-main}).
With mBERT, L2 Alignment performs comparably to no alignment on all 4 tasks (XNLI, NER, POS tagging, and parsing). Compared to no alignment, Linear Mapping performs much worse on NER, performs better on POS tagging and parsing, and performs comparably on XNLI. While previous work observes small improvements on selected languages and tasks, it likely depends on the randomness during evaluation. Based on a more comprehensive evaluation including 4 tasks and multiple seeds, the previously proposed methods do not consistently perform better than no alignment with millions of parallel sentences.

\subsubsection{Contrastive Alignment} In \autoref{tab:signal-main}, with mBERT, both proposed contrastive alignment methods consistently perform as well as no alignment while outperforming more than 1 standard deviation on POS tagging and/or parsing. This suggests the proposed methods are more robust to suboptimal alignments. We hypothesize that learned cosine similarity and contrastive alignment allow the model to recover from suboptimal alignments. Both weak and strong alignment perform comparably.

\subsubsection{Alignment with XLM-R} XLM-R, trained on 2.5TB of text, has the same number of transformer layers
as mBERT but with a larger vocabulary. It performs much better than mBERT. Therefore, we wonder if an explicit alignment objective can similarly lead to better cross-lingual representations. Unfortunately, in \autoref{tab:signal-main}, we find all alignment methods we consider do not improve over no alignment. Compared to no alignment, Linear Mapping and L2 Alignment have worse performance in 3 out of 4 tasks (except POS tagging). In contrast to previous work, both contrastive alignment objectives perform comparably to no alignment in all 4 tasks.

\subsubsection{Impact of Bitext Quality} Even though the OPUS-100 bitext has lower quality compared to bitext used in previous works (due to its greater inclusion of bitext from various sources), by comparing \autoref{tab:signal-main-xlm} and \autoref{tab:signal-main-opus}, we observe that it has minimum impact on each alignment method we consider. This is good news for the lower resource languages, as not all languages are covered by MultiUN or Europarl.

\subsubsection{Model Capacity vs Alignment} XLM-R$_\text{large}$ has nearly twice the number of parameters as XLM-R$_\text{base}$. Even trained on the same data, it performs much better than XLM-R$_\text{base}$, with or without alignment, as shown in \autoref{tab:signal-main}. This suggests increasing model capacity likely leads to better cross-lingual representations than using an explicit alignment objective.

\section{Discussion}

In this chapter, we discuss how to inject cross-lingual signals into multilingual encoders. For type-level cross-lingual signal like bilingual dictionary, we show that adding additional subwords overlap by creating synthetic code-switching corpus with bilingual dictionary benefits cross-lingual representation. For sentence-level cross-lingual signal like bitext, we propose contrastive alignment objective and show that it outperforms L2 Alignment \citep{cao2020multilingual} and consistently performs as well as or better than no alignment using various quality bitext on 4 NLP tasks under a comprehensive evaluation with multiple seeds. 

However, to our surprise, previously proposed methods do not show consistent improvement over no alignment in this setting.
Therefore, we make the following recommendations for future work on cross-lingual alignment or multilingual representations: 1) Evaluations should consider average quality data, not exclusively high-quality bitext. 2) Evaluation must consider multiple NLP tasks or datasets. 3) Evaluation should report \textbf{mean and variance over multiple seeds}, not a single run. More broadly, the community must establish a robust evaluation scheme for zero-shot cross-lingual transfer as a single run with one random seed does not reflect the variance of the method (especially in a zero-shot or few-shot setting).\footnote{This includes zero-shot cross-lingual transfer benchmarks like XGLUE \citep{liang2020xglue} and XTREME \citep{hu2020xtreme}.} While \citet{keung-etal-2020-dont} advocate using oracle for model selection, we instead argue reporting the variance of test performance, following the few-shot learning literature. 

Finally, no explicit alignment methods with bitext improve XLM-R and the larger XLM-R$_\text{large}$ performs much better. While bilingual dictionary contributes to improved cross-lingual representation for 8-layers encoders, the performance gain is likely eclipsed by scaling up the model size. Indeed, as \newcite{kale-etal-2021-nmt5} find that the gain from incorporating bitext into pretraining decreases as model size increase. For smaller model, incorporating cross-lingual signal explicitly might still offers good performance gain. However, as raw text is easier to obtain than bitext, scaling models to more raw text and larger capacity models may be more beneficial for producing better cross-lingual models, as evidenced by \citet{xue-etal-2021-mt5} and \citet{goyal-etal-2021-larger}.

In this chapter, we observe that zero-shot cross-lingual transfer has low variance on source language generalization performance but high variance on target language generalization performance (\autoref{tab:signal-main-xlm} or \autoref{tab:signal-main-opus}). In \autoref{chap:analysis}, we will investigate why zero-shot cross-lingual transfer has high variance.

%% file: table/signal/dict.tex
\begin{table*}[]
\begin{center}
\begin{tabular}[b]{l|ccc|ccc|ccc}
\toprule
 & \multicolumn{3}{c|}{XNLI (Acc)} & \multicolumn{3}{c|}{NER (F1)} & \multicolumn{3}{c}{Parsing (LAS)} \\
 & fr & ru & zh & fr & ru & zh & fr & ru & zh \\
\midrule
Default     & 73.6 & 68.7 & 68.3 & \textbf{79.8} & \textbf{60.9} & 63.6 & 73.2 & 56.6 & 28.8 \\
\midrule
+ Bi. Dict. & \textbf{74.0} & \textbf{69.8} & \textbf{72.1} & 76.1 & 59.7 & \textbf{66.8} & \textbf{73.3} & \textbf{56.9} & \textbf{29.2} \\
\bottomrule
\end{tabular}
\caption{Impact of extra anchor points with synthetic code-switching corpus based on bilingual dictionary.
\label{tab:signal-dict}}
\end{center}
\end{table*}

%% file: table/signal/main.tex
\begin{table*}[]
\centering
\resizebox{0.8\linewidth}{!}{
\begin{tabular}[b]{l|cccc}
\toprule
 & XNLI & NER & POS & Parsing \\
\midrule

mBERT & 70.1$_{\pm 0.8}$ & 67.7$_{\pm 1.3}$ & 78.3$_{\pm 0.5}$ & 52.6$_{\pm 0.4}$ \\
+ Linear Mapping & 70.0$_{\pm 0.6}$ & \bad 63.7$_{\pm 1.5}$ & \good 79.5$_{\pm 0.5}$ & \good 53.6$_{\pm 0.3}$ \\
+ L2 Align & 69.7$_{\pm 0.4}$ & 67.1$_{\pm 1.0}$ & 78.0$_{\pm 1.3}$ & 52.2$_{\pm 0.7}$ \\
+ Weak Align (Our) & 70.5$_{\pm 0.7}$ & 68.0$_{\pm 1.3}$ & \good 78.8$_{\pm 0.7}$ & \good 53.1$_{\pm 0.6}$ \\
+ Strong Align (Our) & 70.4$_{\pm 0.7}$ & 67.7$_{\pm 1.1}$ & \good 79.0$_{\pm 0.7}$ & 53.0$_{\pm 0.6}$ \\
\midrule

XLM-R$_{\text{base}}$ & 76.4$_{\pm 0.5}$ & 66.4$_{\pm 0.9}$ & 81.2$_{\pm 0.6}$ & 57.3$_{\pm 0.6}$ \\
+ Linear Mapping & \bad 73.4$_{\pm 0.6}$ & \bad 54.1$_{\pm 0.9}$ & 81.3$_{\pm 0.5}$ & \bad 55.6$_{\pm 0.5}$ \\
+ L2 Align & \bad 75.7$_{\pm 0.5}$ & 65.7$_{\pm 1.2}$ & 81.3$_{\pm 0.9}$ & \bad 56.2$_{\pm 0.7}$ \\
+ Weak Align (Our) & 76.1$_{\pm 0.7}$ & 66.0$_{\pm 1.0}$ & 81.5$_{\pm 0.5}$ & 57.4$_{\pm 0.4}$ \\
+ Strong Align (Our) & 76.0$_{\pm 0.6}$ & 66.1$_{\pm 0.9}$ & 81.4$_{\pm 0.6}$ & 57.4$_{\pm 0.5}$ \\
\midrule

XLM-R$_{\text{large}}$ & 80.4$_{\pm 0.6}$  & 71.0$_{\pm 1.4}$  & 82.6$_{\pm 0.5}$  & 59.4$_{\pm 0.8}$  \\
\bottomrule
\end{tabular}
}
\caption{Alignment with bitext used in previous works}
\label{tab:signal-main-xlm}
\resizebox{0.8\linewidth}{!}{
\begin{tabular}[b]{l|cccc}
\toprule
 & XNLI & NER & POS & Parsing \\
\midrule

mBERT & 70.1$_{\pm 0.8}$ & 67.7$_{\pm 1.3}$ & 78.3$_{\pm 0.5}$ & 52.6$_{\pm 0.4}$ \\
+ Linear Mapping & 70.2$_{\pm 0.6}$ & \bad 63.8$_{\pm 1.3}$ & \good 80.1$_{\pm 0.4}$ & \good 53.6$_{\pm 0.3}$ \\
+ L2 Align & 70.3$_{\pm 0.5}$ & 67.8$_{\pm 1.4}$ & 78.2$_{\pm 1.2}$ & 52.8$_{\pm 0.7}$ \\
+ Weak Align (Our) & 70.8$_{\pm 0.7}$ & 67.3$_{\pm 0.9}$ & \good 78.8$_{\pm 0.6}$ & 52.9$_{\pm 0.6}$ \\
+ Strong Align (Our) & 70.4$_{\pm 0.7}$ & 67.2$_{\pm 1.1}$ & \good 79.0$_{\pm 0.7}$ & \good 53.3$_{\pm 0.6}$ \\
\midrule

XLM-R$_{\text{base}}$ & 76.4$_{\pm 0.5}$ & 66.4$_{\pm 0.9}$ & 81.2$_{\pm 0.6}$ & 57.3$_{\pm 0.6}$ \\
+ Linear Mapping & \bad 73.5$_{\pm 0.5}$ & \bad 54.2$_{\pm 0.8}$ & 81.7$_{\pm 0.6}$ & \bad 56.1$_{\pm 0.4}$ \\
+ L2 Align & \bad 75.8$_{\pm 0.5}$ & \bad 65.5$_{\pm 1.2}$ & 81.4$_{\pm 0.8}$ & \bad 55.9$_{\pm 0.6}$ \\
+ Weak Align (Our) & 76.0$_{\pm 0.4}$ & 66.2$_{\pm 1.2}$ & 81.5$_{\pm 0.5}$ & 57.4$_{\pm 0.5}$ \\
+ Strong Align (Our) & 76.1$_{\pm 0.4}$ & 66.2$_{\pm 1.0}$ & 81.5$_{\pm 0.6}$ & 57.4$_{\pm 0.5}$ \\
\midrule

XLM-R$_{\text{large}}$ & 80.4$_{\pm 0.6}$  & 71.0$_{\pm 1.4}$  & 82.6$_{\pm 0.5}$  & 59.4$_{\pm 0.8}$  \\
\bottomrule
\end{tabular}
}
\caption{Alignment with the OPUS-100 bitext}
\label{tab:signal-main-opus}

\caption{Zero-shot cross-lingual transfer result, average over 9 languages. Breakdown can be found in \autoref{tab:signal-xlm} and \autoref{tab:signal-opus}.
\textcolor{good}{Blue} or \textcolor{bad}{orange} indicates the mean performance is one standard derivation \textcolor{good}{above} or \textcolor{bad}{below} the mean of baseline. While mBERT benefits from alignment in some cases, extra alignment does not improve XLM-R.\label{tab:signal-main}}
\end{table*}

%% file: table/signal/xlm.tex
\begin{table*}[]
\begin{center}
\resizebox{1\linewidth}{!}{
\begin{tabular}[b]{l|ccc ccc ccc|c}
\toprule

 & \textbf{ar} & \textbf{de} & \textbf{en} & \textbf{es} & \textbf{fr} & \textbf{hi} & \textbf{ru} & \textbf{vi} & \textbf{zh} & \textbf{AVER} \\

\midrule
\multicolumn{11}{l}{\textbf{XNLI (Accuracy)}} \\
\midrule
mBERT & 64.2$_{\pm 0.9}$ & 70.5$_{\pm 0.2}$ & 82.5$_{\pm 0.3}$ & 74.2$_{\pm 1.2}$ & 73.8$_{\pm 0.8}$ & 59.4$_{\pm 0.7}$ & 68.3$_{\pm 0.9}$ & 69.6$_{\pm 0.7}$ & 68.6$_{\pm 0.9}$ & 70.1$_{\pm 0.8}$ \\
+ Linear Mapping & 63.8$_{\pm 0.6}$ & 70.4$_{\pm 0.4}$ & \bad 81.0$_{\pm 0.5}$ & 73.9$_{\pm 0.9}$ & \bad 72.5$_{\pm 0.8}$ & \good 61.2$_{\pm 0.7}$ & \bad 67.1$_{\pm 0.4}$ & 70.2$_{\pm 0.5}$ & \good 70.1$_{\pm 0.8}$ & 70.0$_{\pm 0.6}$ \\
+ L2 Align & 64.1$_{\pm 0.4}$ & \bad 70.0$_{\pm 0.7}$ & 82.2$_{\pm 0.4}$ & 73.9$_{\pm 0.5}$ & 73.8$_{\pm 0.2}$ & \bad 58.5$_{\pm 0.3}$ & 67.9$_{\pm 0.4}$ & 69.4$_{\pm 0.6}$ & 67.9$_{\pm 0.4}$ & 69.7$_{\pm 0.4}$ \\
+ Weak Align (Our) & 64.9$_{\pm 0.8}$ & \good 71.0$_{\pm 0.8}$ & 82.3$_{\pm 0.4}$ & 74.6$_{\pm 0.7}$ & 73.8$_{\pm 0.4}$ & 59.8$_{\pm 0.3}$ & 68.5$_{\pm 1.0}$ & 70.3$_{\pm 0.8}$ & 69.4$_{\pm 1.0}$ & 70.5$_{\pm 0.7}$ \\
+ Strong Align (Our) & 64.8$_{\pm 0.8}$ & 70.5$_{\pm 0.9}$ & 82.3$_{\pm 0.5}$ & 74.4$_{\pm 0.6}$ & 74.1$_{\pm 0.7}$ & 59.8$_{\pm 0.9}$ & 68.2$_{\pm 0.6}$ & 70.1$_{\pm 0.8}$ & 69.0$_{\pm 1.0}$ & 70.4$_{\pm 0.7}$ \\

\midrule
XLM-R$_\text{base}$ & 71.8$_{\pm 0.2}$ & 77.3$_{\pm 0.5}$ & 85.1$_{\pm 0.3}$ & 79.3$_{\pm 0.5}$ & 78.8$_{\pm 0.4}$ & 70.3$_{\pm 0.6}$ & 75.9$_{\pm 0.5}$ & 74.8$_{\pm 0.4}$ & 74.1$_{\pm 0.5}$ & 76.4$_{\pm 0.5}$ \\
+ Linear Mapping & \bad 69.7$_{\pm 0.6}$ & \bad 74.3$_{\pm 0.3}$ & \bad 82.5$_{\pm 0.6}$ & \bad 76.4$_{\pm 0.5}$ & \bad 75.5$_{\pm 0.4}$ & \bad 67.2$_{\pm 0.9}$ & \bad 73.2$_{\pm 0.3}$ & \bad 72.5$_{\pm 0.5}$ & \bad 68.9$_{\pm 1.2}$ & \bad 73.4$_{\pm 0.6}$ \\
+ L2 Align & 71.6$_{\pm 0.8}$ & \bad 76.0$_{\pm 0.5}$ & \bad 84.5$_{\pm 0.5}$ & \bad 78.6$_{\pm 0.3}$ & \bad 77.9$_{\pm 0.3}$ & 69.8$_{\pm 0.7}$ & \bad 75.3$_{\pm 0.3}$ & \bad 74.0$_{\pm 0.4}$ & 73.7$_{\pm 0.7}$ & \bad 75.7$_{\pm 0.5}$ \\
+ Weak Align (Our) & 71.7$_{\pm 0.7}$ & \bad 76.5$_{\pm 0.6}$ & \bad 84.7$_{\pm 0.6}$ & \bad 78.7$_{\pm 0.6}$ & \bad 78.1$_{\pm 0.7}$ & 70.4$_{\pm 0.9}$ & 75.8$_{\pm 0.6}$ & 74.5$_{\pm 0.5}$ & 74.2$_{\pm 0.7}$ & 76.1$_{\pm 0.7}$ \\
+ Strong Align (Our) & 71.6$_{\pm 0.5}$ & \bad 76.6$_{\pm 0.4}$ & \bad 84.7$_{\pm 0.5}$ & 79.0$_{\pm 0.4}$ & \bad 78.3$_{\pm 0.3}$ & 70.0$_{\pm 1.0}$ & 75.7$_{\pm 0.7}$ & 74.7$_{\pm 0.4}$ & 73.7$_{\pm 0.8}$ & 76.0$_{\pm 0.6}$ \\

\midrule
XLM-R$_{\text{large}}$ & 77.5$_{\pm 0.6}$ & 81.7$_{\pm 0.4}$ & 88.0$_{\pm 0.3}$ & 83.3$_{\pm 0.6}$ & 82.0$_{\pm 0.5}$ & 75.1$_{\pm 0.8}$ & 79.2$_{\pm 0.7}$ & 78.4$_{\pm 0.6}$ & 78.3$_{\pm 0.6}$ & 80.4$_{\pm 0.6}$ \\

\midrule
\multicolumn{11}{l}{\textbf{NER (Entity-level F1)}} \\
\midrule
mBERT & 42.0$_{\pm 2.9}$ & 79.0$_{\pm 0.3}$ & 84.1$_{\pm 0.2}$ & 73.3$_{\pm 2.5}$ & 78.9$_{\pm 0.3}$ & 65.7$_{\pm 1.4}$ & 65.2$_{\pm 1.4}$ & 69.7$_{\pm 1.8}$ & 51.7$_{\pm 0.8}$ & 67.7$_{\pm 1.3}$ \\
+ Linear Mapping & \bad 36.9$_{\pm 1.1}$ & \bad 76.1$_{\pm 0.4}$ & \bad 82.8$_{\pm 0.1}$ & \bad 70.4$_{\pm 2.1}$ & \bad 77.4$_{\pm 0.7}$ & 64.5$_{\pm 1.4}$ & \bad 59.5$_{\pm 2.5}$ & \bad 65.2$_{\pm 2.7}$ & \bad 40.5$_{\pm 2.0}$ & \bad 63.7$_{\pm 1.5}$ \\
+ L2 Align & 39.7$_{\pm 1.6}$ & \bad 77.7$_{\pm 0.8}$ & 84.0$_{\pm 0.1}$ & 72.5$_{\pm 1.5}$ & 79.1$_{\pm 0.3}$ & \bad 63.3$_{\pm 1.8}$ & 64.3$_{\pm 1.0}$ & 71.2$_{\pm 0.9}$ & 52.1$_{\pm 1.1}$ & 67.1$_{\pm 1.0}$ \\
+ Weak Align (Our) & 42.3$_{\pm 2.7}$ & 78.7$_{\pm 0.3}$ & 84.2$_{\pm 0.2}$ & 71.6$_{\pm 2.2}$ & \good 79.4$_{\pm 0.6}$ & \good 67.6$_{\pm 1.3}$ & 64.8$_{\pm 0.8}$ & 70.0$_{\pm 2.3}$ & \good 52.9$_{\pm 0.9}$ & 68.0$_{\pm 1.3}$ \\
+ Strong Align (Our) & 40.6$_{\pm 1.0}$ & \bad 78.7$_{\pm 0.3}$ & 84.2$_{\pm 0.2}$ & 72.2$_{\pm 2.5}$ & 79.0$_{\pm 0.5}$ & \good 67.2$_{\pm 0.7}$ & 64.5$_{\pm 1.7}$ & 70.1$_{\pm 2.5}$ & \good 52.5$_{\pm 0.8}$ & 67.7$_{\pm 1.1}$ \\

\midrule
XLM-R$_\text{base}$ & 44.0$_{\pm 1.3}$ & 75.0$_{\pm 0.3}$ & 82.2$_{\pm 0.2}$ & 76.0$_{\pm 2.4}$ & 77.6$_{\pm 0.7}$ & 65.7$_{\pm 0.6}$ & 64.1$_{\pm 0.7}$ & 68.0$_{\pm 1.2}$ & 45.1$_{\pm 0.8}$ & 66.4$_{\pm 0.9}$ \\
+ Linear Mapping & \bad 30.8$_{\pm 2.1}$ & \bad 69.0$_{\pm 0.6}$ & \bad 78.3$_{\pm 0.3}$ & \bad 59.8$_{\pm 0.5}$ & \bad 67.8$_{\pm 0.7}$ & \bad 57.9$_{\pm 1.5}$ & \bad 48.0$_{\pm 1.0}$ & \bad 54.4$_{\pm 0.5}$ & \bad 21.0$_{\pm 0.9}$ & \bad 54.1$_{\pm 0.9}$ \\
+ L2 Align & 44.9$_{\pm 2.1}$ & 74.9$_{\pm 0.6}$ & 82.1$_{\pm 0.3}$ & 75.0$_{\pm 3.1}$ & 77.1$_{\pm 0.6}$ & 65.5$_{\pm 1.3}$ & \bad 63.2$_{\pm 0.3}$ & \bad 66.3$_{\pm 2.2}$ & \bad 42.4$_{\pm 0.7}$ & 65.7$_{\pm 1.2}$ \\
+ Weak Align (Our) & \good 45.6$_{\pm 1.4}$ & 75.0$_{\pm 0.5}$ & 82.2$_{\pm 0.2}$ & 74.2$_{\pm 2.4}$ & 77.2$_{\pm 0.8}$ & 65.8$_{\pm 1.1}$ & 63.6$_{\pm 1.1}$ & 67.6$_{\pm 0.7}$ & \bad 42.8$_{\pm 0.6}$ & 66.0$_{\pm 1.0}$ \\
+ Strong Align (Our) & \good 45.7$_{\pm 1.7}$ & 75.1$_{\pm 0.6}$ & 82.1$_{\pm 0.3}$ & \bad 73.5$_{\pm 1.7}$ & 77.2$_{\pm 0.6}$ & 65.8$_{\pm 1.7}$ & 63.7$_{\pm 0.5}$ & 68.1$_{\pm 0.8}$ & \bad 43.2$_{\pm 0.4}$ & 66.1$_{\pm 0.9}$ \\

\midrule
XLM-R$_{\text{large}}$ & 46.8$_{\pm 4.3}$ & 79.1$_{\pm 0.5}$ & 84.2$_{\pm 0.2}$ & 75.7$_{\pm 2.9}$ & 80.7$_{\pm 0.5}$ & 71.6$_{\pm 1.1}$ & 71.7$_{\pm 0.5}$ & 77.4$_{\pm 1.3}$ & 51.5$_{\pm 1.4}$ & 71.0$_{\pm 1.4}$ \\

\midrule
\multicolumn{11}{l}{\textbf{POS (Accuracy)}} \\
\midrule
mBERT & 60.3$_{\pm 0.9}$ & 90.4$_{\pm 0.3}$ & 96.9$_{\pm 0.1}$ & 87.7$_{\pm 0.2}$ & 88.9$_{\pm 0.3}$ & 68.0$_{\pm 0.8}$ & 82.5$_{\pm 0.7}$ & 62.7$_{\pm 0.2}$ & 67.1$_{\pm 1.1}$ & 78.3$_{\pm 0.5}$ \\
+ Linear Mapping & \good 73.6$_{\pm 0.7}$ & \bad 88.2$_{\pm 0.5}$ & \bad 96.3$_{\pm 0.0}$ & \bad 87.4$_{\pm 0.1}$ & 88.9$_{\pm 0.3}$ & \good 77.3$_{\pm 0.6}$ & \bad 78.0$_{\pm 1.0}$ & \bad 60.4$_{\pm 0.5}$ & \bad 65.7$_{\pm 1.3}$ & \good 79.5$_{\pm 0.5}$ \\
+ L2 Align & \good 63.4$_{\pm 2.6}$ & \bad 89.3$_{\pm 0.7}$ & \bad 96.7$_{\pm 0.2}$ & \bad 86.7$_{\pm 0.3}$ & \bad 87.9$_{\pm 0.5}$ & \bad 65.2$_{\pm 3.9}$ & \good 83.6$_{\pm 0.9}$ & \bad 62.3$_{\pm 0.8}$ & 66.5$_{\pm 1.5}$ & 78.0$_{\pm 1.3}$ \\
+ Weak Align (Our) & \good 61.6$_{\pm 2.0}$ & 90.3$_{\pm 0.7}$ & 96.9$_{\pm 0.1}$ & 87.5$_{\pm 0.6}$ & \bad 88.6$_{\pm 0.3}$ & \good 70.3$_{\pm 0.9}$ & 83.1$_{\pm 0.6}$ & \good 63.2$_{\pm 0.3}$ & 68.1$_{\pm 0.9}$ & \good 78.8$_{\pm 0.7}$ \\
+ Strong Align (Our) & \good 61.9$_{\pm 2.0}$ & 90.4$_{\pm 0.7}$ & 96.9$_{\pm 0.0}$ & 87.5$_{\pm 0.5}$ & \bad 88.5$_{\pm 0.4}$ & \good 71.1$_{\pm 1.2}$ & 83.0$_{\pm 0.5}$ & \good 63.2$_{\pm 0.2}$ & 68.0$_{\pm 0.6}$ & \good 79.0$_{\pm 0.7}$ \\

\midrule
XLM-R$_\text{base}$ & 70.2$_{\pm 1.6}$ & 91.6$_{\pm 0.3}$ & 97.5$_{\pm 0.0}$ & 88.5$_{\pm 0.2}$ & 89.4$_{\pm 0.3}$ & 71.7$_{\pm 1.3}$ & 86.1$_{\pm 0.3}$ & 64.5$_{\pm 0.5}$ & 71.4$_{\pm 0.5}$ & 81.2$_{\pm 0.6}$ \\
+ Linear Mapping & \good 74.3$_{\pm 1.1}$ & \bad 90.7$_{\pm 0.5}$ & \bad 96.9$_{\pm 0.0}$ & \bad 88.2$_{\pm 0.1}$ & 89.3$_{\pm 0.3}$ & \good 82.1$_{\pm 0.9}$ & \bad 82.7$_{\pm 0.4}$ & \bad 62.6$_{\pm 0.4}$ & \bad 65.3$_{\pm 1.0}$ & 81.3$_{\pm 0.5}$ \\
+ L2 Align & 71.1$_{\pm 1.8}$ & 91.4$_{\pm 0.3}$ & \bad 97.4$_{\pm 0.0}$ & \bad 88.2$_{\pm 0.2}$ & \bad 89.0$_{\pm 0.3}$ & 73.0$_{\pm 3.8}$ & \good 86.6$_{\pm 0.2}$ & 64.4$_{\pm 0.4}$ & \bad 70.8$_{\pm 0.8}$ & 81.3$_{\pm 0.9}$ \\
+ Weak Align (Our) & \good 72.8$_{\pm 0.7}$ & \bad 91.1$_{\pm 0.2}$ & \bad 97.4$_{\pm 0.0}$ & 88.3$_{\pm 0.2}$ & 89.2$_{\pm 0.2}$ & 72.4$_{\pm 1.6}$ & 86.4$_{\pm 0.1}$ & 64.7$_{\pm 0.4}$ & 71.6$_{\pm 1.2}$ & 81.5$_{\pm 0.5}$ \\
+ Strong Align (Our) & \good 72.5$_{\pm 0.9}$ & \bad 91.1$_{\pm 0.3}$ & \bad 97.4$_{\pm 0.0}$ & \bad 88.3$_{\pm 0.2}$ & \bad 89.1$_{\pm 0.1}$ & 72.0$_{\pm 2.1}$ & 86.4$_{\pm 0.1}$ & 64.8$_{\pm 0.4}$ & 71.4$_{\pm 1.1}$ & 81.4$_{\pm 0.6}$ \\

\midrule
XLM-R$_{\text{large}}$ & 73.9$_{\pm 1.0}$ & 91.9$_{\pm 0.3}$ & 98.0$_{\pm 0.0}$ & 89.2$_{\pm 0.2}$ & 89.8$_{\pm 0.1}$ & 78.4$_{\pm 2.1}$ & 86.5$_{\pm 0.2}$ & 64.8$_{\pm 0.3}$ & 71.0$_{\pm 0.3}$ & 82.6$_{\pm 0.5}$ \\

\midrule
\multicolumn{11}{l}{\textbf{Parsing (Labeled Attachment Score)}} \\
\midrule
mBERT & 28.8$_{\pm 0.4}$ & 67.8$_{\pm 0.5}$ & 79.7$_{\pm 0.1}$ & 69.1$_{\pm 0.1}$ & 73.3$_{\pm 0.2}$ & 31.0$_{\pm 0.5}$ & 60.2$_{\pm 0.6}$ & 33.5$_{\pm 0.5}$ & 29.5$_{\pm 0.4}$ & 52.6$_{\pm 0.4}$ \\
+ Linear Mapping & \good 44.1$_{\pm 0.3}$ & \bad 64.4$_{\pm 0.4}$ & \good 80.5$_{\pm 0.2}$ & \good 70.2$_{\pm 0.3}$ & \good 73.9$_{\pm 0.1}$ & \good 32.2$_{\pm 0.3}$ & \bad 56.7$_{\pm 0.5}$ & \bad 32.1$_{\pm 0.2}$ & \bad 28.1$_{\pm 0.3}$ & \good 53.6$_{\pm 0.3}$ \\
+ L2 Align & \good 29.6$_{\pm 1.6}$ & \bad 66.9$_{\pm 0.2}$ & \bad 79.2$_{\pm 0.2}$ & \bad 68.2$_{\pm 0.4}$ & \bad 72.5$_{\pm 0.5}$ & 30.8$_{\pm 1.9}$ & 60.0$_{\pm 0.6}$ & 33.3$_{\pm 0.4}$ & 29.5$_{\pm 0.4}$ & 52.2$_{\pm 0.7}$ \\
+ Weak Align (Our) & \good 30.7$_{\pm 0.9}$ & 67.6$_{\pm 0.6}$ & \good 79.8$_{\pm 0.1}$ & \good 69.7$_{\pm 0.4}$ & \good 73.6$_{\pm 0.4}$ & 31.2$_{\pm 0.8}$ & \good 61.3$_{\pm 0.7}$ & 33.5$_{\pm 0.6}$ & \good 30.5$_{\pm 0.6}$ & \good 53.1$_{\pm 0.6}$ \\
+ Strong Align (Our) & \good 31.2$_{\pm 1.1}$ & 67.5$_{\pm 0.4}$ & 79.8$_{\pm 0.1}$ & \good 69.4$_{\pm 0.3}$ & 73.4$_{\pm 0.5}$ & 30.7$_{\pm 1.5}$ & \good 61.3$_{\pm 0.8}$ & 33.5$_{\pm 0.6}$ & \good 30.0$_{\pm 0.5}$ & \good 53.0$_{\pm 0.6}$ \\

\midrule
XLM-R$_\text{base}$ & 43.7$_{\pm 1.7}$ & 69.0$_{\pm 0.4}$ & 80.5$_{\pm 0.2}$ & 71.0$_{\pm 0.4}$ & 73.6$_{\pm 0.5}$ & 41.2$_{\pm 0.9}$ & 66.3$_{\pm 0.9}$ & 36.6$_{\pm 0.2}$ & 34.2$_{\pm 0.7}$ & 57.3$_{\pm 0.6}$ \\
+ Linear Mapping & \good 47.2$_{\pm 0.6}$ & \bad 66.7$_{\pm 0.3}$ & \good 81.4$_{\pm 0.1}$ & \good 72.6$_{\pm 0.2}$ & \good 74.4$_{\pm 0.4}$ & 41.4$_{\pm 0.7}$ & \bad 60.8$_{\pm 0.6}$ & \bad 34.3$_{\pm 0.3}$ & \bad 21.5$_{\pm 1.1}$ & \bad 55.6$_{\pm 0.5}$ \\
+ L2 Align & \bad 41.3$_{\pm 1.8}$ & \bad 68.1$_{\pm 0.3}$ & \bad 79.7$_{\pm 0.2}$ & \bad 70.0$_{\pm 0.5}$ & \bad 73.0$_{\pm 0.5}$ & \bad 40.2$_{\pm 1.6}$ & \bad 63.7$_{\pm 0.9}$ & 36.5$_{\pm 0.5}$ & \bad 32.9$_{\pm 0.3}$ & \bad 56.2$_{\pm 0.7}$ \\
+ Weak Align (Our) & 44.6$_{\pm 1.0}$ & 68.8$_{\pm 0.4}$ & 80.4$_{\pm 0.1}$ & 71.4$_{\pm 0.2}$ & 73.9$_{\pm 0.2}$ & 41.0$_{\pm 0.6}$ & 65.7$_{\pm 0.4}$ & 36.7$_{\pm 0.4}$ & 33.8$_{\pm 0.3}$ & 57.4$_{\pm 0.4}$ \\
+ Strong Align (Our) & 44.8$_{\pm 0.9}$ & 68.9$_{\pm 0.5}$ & 80.4$_{\pm 0.1}$ & 71.3$_{\pm 0.2}$ & 73.9$_{\pm 0.1}$ & 40.7$_{\pm 0.8}$ & 66.2$_{\pm 0.4}$ & 36.7$_{\pm 0.3}$ & 34.0$_{\pm 0.8}$ & 57.4$_{\pm 0.5}$ \\

\midrule
XLM-R$_{\text{large}}$ & 48.2$_{\pm 1.5}$ & 67.8$_{\pm 0.6}$ & 82.6$_{\pm 0.3}$ & 73.9$_{\pm 0.4}$ & 76.4$_{\pm 0.4}$ & 41.8$_{\pm 2.5}$ & 69.6$_{\pm 0.4}$ & 38.9$_{\pm 0.6}$ & 35.4$_{\pm 0.5}$ & 59.4$_{\pm 0.8}$ \\

\bottomrule
\end{tabular}
}
\caption{Zero-shot cross-lingual transfer result with bitext from previous works. 
\textcolor{good}{Blue} or \textcolor{bad}{orange} indicates the mean performance is one standard derivation \textcolor{good}{above} or \textcolor{bad}{below} the mean of baseline.
\label{tab:signal-xlm}}
\end{center}
\end{table*}

%% file: table/signal/opus.tex
\begin{table*}[]
\begin{center}
\resizebox{1\linewidth}{!}{
\begin{tabular}[b]{l|ccc ccc ccc|c}
\toprule

 & \textbf{ar} & \textbf{de} & \textbf{en} & \textbf{es} & \textbf{fr} & \textbf{hi} & \textbf{ru} & \textbf{vi} & \textbf{zh} & \textbf{AVER} \\

\midrule
\multicolumn{11}{l}{\textbf{XNLI (Accuracy)}} \\
\midrule

mBERT & 64.2$_{\pm 0.9}$ & 70.5$_{\pm 0.2}$ & 82.5$_{\pm 0.3}$ & 74.2$_{\pm 1.2}$ & 73.8$_{\pm 0.8}$ & 59.4$_{\pm 0.7}$ & 68.3$_{\pm 0.9}$ & 69.6$_{\pm 0.7}$ & 68.6$_{\pm 0.9}$ & 70.1$_{\pm 0.8}$ \\
+ Linear Mapping & 64.1$_{\pm 0.7}$ & \bad 70.0$_{\pm 0.6}$ & \bad 81.0$_{\pm 0.5}$ & 74.1$_{\pm 0.6}$ & \bad 72.9$_{\pm 0.9}$ & \good 61.8$_{\pm 0.7}$ & \bad 67.4$_{\pm 0.6}$ & 70.2$_{\pm 0.5}$ & \good 70.2$_{\pm 0.8}$ & 70.2$_{\pm 0.6}$ \\
+ L2 Align & 64.3$_{\pm 0.5}$ & 70.7$_{\pm 1.0}$ & 82.5$_{\pm 0.5}$ & 74.3$_{\pm 0.3}$ & 74.0$_{\pm 0.4}$ & 59.3$_{\pm 0.4}$ & 68.6$_{\pm 0.7}$ & 69.7$_{\pm 0.4}$ & 69.1$_{\pm 0.5}$ & 70.3$_{\pm 0.5}$ \\
+ Weak Align (Our) & 65.1$_{\pm 0.9}$ & \good 70.9$_{\pm 0.6}$ & 82.6$_{\pm 0.5}$ & 74.9$_{\pm 0.6}$ & 74.1$_{\pm 0.4}$ & \good 60.3$_{\pm 0.6}$ & 68.9$_{\pm 0.8}$ & \good 70.6$_{\pm 0.6}$ & \good 69.6$_{\pm 1.0}$ & 70.8$_{\pm 0.7}$ \\
+ Strong Align (Our) & 64.7$_{\pm 0.9}$ & \good 70.8$_{\pm 0.7}$ & 82.4$_{\pm 0.1}$ & 74.5$_{\pm 0.7}$ & 73.9$_{\pm 0.7}$ & 59.6$_{\pm 0.6}$ & 68.5$_{\pm 1.1}$ & \good 70.4$_{\pm 0.6}$ & 69.1$_{\pm 1.0}$ & 70.4$_{\pm 0.7}$ \\

\midrule
XLM-R$_\text{base}$ & 71.8$_{\pm 0.2}$ & 77.3$_{\pm 0.5}$ & 85.1$_{\pm 0.3}$ & 79.3$_{\pm 0.5}$ & 78.8$_{\pm 0.4}$ & 70.3$_{\pm 0.6}$ & 75.9$_{\pm 0.5}$ & 74.8$_{\pm 0.4}$ & 74.1$_{\pm 0.5}$ & 76.4$_{\pm 0.5}$ \\
+ Linear Mapping & \bad 69.9$_{\pm 0.4}$ & \bad 74.3$_{\pm 0.3}$ & \bad 82.5$_{\pm 0.6}$ & \bad 76.4$_{\pm 0.5}$ & \bad 75.5$_{\pm 0.6}$ & \bad 67.2$_{\pm 1.0}$ & \bad 72.7$_{\pm 0.2}$ & \bad 72.7$_{\pm 0.5}$ & \bad 70.1$_{\pm 0.8}$ & \bad 73.5$_{\pm 0.5}$ \\
+ L2 Align & 71.9$_{\pm 0.6}$ & \bad 76.4$_{\pm 0.4}$ & \bad 84.6$_{\pm 0.3}$ & \bad 78.4$_{\pm 0.5}$ & \bad 77.8$_{\pm 0.3}$ & 69.9$_{\pm 0.8}$ & \bad 75.2$_{\pm 0.5}$ & \bad 74.2$_{\pm 0.5}$ & 73.7$_{\pm 0.5}$ & \bad 75.8$_{\pm 0.5}$ \\
+ Weak Align (Our) & 71.8$_{\pm 0.6}$ & \bad 76.5$_{\pm 0.5}$ & \bad 84.6$_{\pm 0.2}$ & 79.0$_{\pm 0.4}$ & 78.4$_{\pm 0.5}$ & 70.0$_{\pm 0.5}$ & 75.7$_{\pm 0.3}$ & 74.7$_{\pm 0.3}$ & \bad 73.4$_{\pm 0.6}$ & 76.0$_{\pm 0.4}$ \\
+ Strong Align (Our) & 72.0$_{\pm 0.5}$ & \bad 76.6$_{\pm 0.4}$ & 84.8$_{\pm 0.1}$ & 79.0$_{\pm 0.4}$ & 78.6$_{\pm 0.5}$ & 70.1$_{\pm 0.3}$ & 75.7$_{\pm 0.4}$ & 74.8$_{\pm 0.6}$ & 73.8$_{\pm 0.6}$ & 76.1$_{\pm 0.4}$ \\

\midrule
XLM-R$_{\text{large}}$ & 77.5$_{\pm 0.6}$ & 81.7$_{\pm 0.4}$ & 88.0$_{\pm 0.3}$ & 83.3$_{\pm 0.6}$ & 82.0$_{\pm 0.5}$ & 75.1$_{\pm 0.8}$ & 79.2$_{\pm 0.7}$ & 78.4$_{\pm 0.6}$ & 78.3$_{\pm 0.6}$ & 80.4$_{\pm 0.6}$ \\

\midrule
\multicolumn{11}{l}{\textbf{NER (Entity-level F1)}} \\
\midrule

mBERT & 42.0$_{\pm 2.9}$ & 79.0$_{\pm 0.3}$ & 84.1$_{\pm 0.2}$ & 73.3$_{\pm 2.5}$ & 78.9$_{\pm 0.3}$ & 65.7$_{\pm 1.4}$ & 65.2$_{\pm 1.4}$ & 69.7$_{\pm 1.8}$ & 51.7$_{\pm 0.8}$ & 67.7$_{\pm 1.3}$ \\
+ Linear Mapping & \bad 36.9$_{\pm 0.9}$ & \bad 76.2$_{\pm 0.3}$ & \bad 82.8$_{\pm 0.1}$ & 71.2$_{\pm 1.5}$ & \bad 77.4$_{\pm 0.7}$ & \bad 62.4$_{\pm 2.2}$ & \bad 59.6$_{\pm 2.4}$ & \bad 65.4$_{\pm 2.6}$ & \bad 42.3$_{\pm 1.4}$ & \bad 63.8$_{\pm 1.3}$ \\
+ L2 Align & 41.3$_{\pm 3.2}$ & \bad 78.2$_{\pm 1.0}$ & 84.1$_{\pm 0.1}$ & 73.4$_{\pm 2.4}$ & \good 79.7$_{\pm 0.8}$ & 64.9$_{\pm 1.5}$ & 64.9$_{\pm 1.6}$ & \good 71.8$_{\pm 0.9}$ & 52.4$_{\pm 1.3}$ & 67.8$_{\pm 1.4}$ \\
+ Weak Align (Our) & 40.3$_{\pm 1.1}$ & \bad 78.7$_{\pm 0.3}$ & 84.0$_{\pm 0.1}$ & \bad 70.7$_{\pm 2.1}$ & 79.0$_{\pm 0.4}$ & \good 67.2$_{\pm 1.2}$ & 64.9$_{\pm 1.2}$ & 69.1$_{\pm 0.8}$ & 52.0$_{\pm 1.1}$ & 67.3$_{\pm 0.9}$ \\
+ Strong Align (Our) & 40.7$_{\pm 1.9}$ & \bad 78.3$_{\pm 0.3}$ & 84.2$_{\pm 0.1}$ & \bad 70.0$_{\pm 2.6}$ & 78.8$_{\pm 0.3}$ & 66.7$_{\pm 1.4}$ & 64.8$_{\pm 0.9}$ & 69.5$_{\pm 1.4}$ & 52.1$_{\pm 0.6}$ & 67.2$_{\pm 1.1}$ \\

\midrule
XLM-R$_\text{base}$ & 44.0$_{\pm 1.3}$ & 75.0$_{\pm 0.3}$ & 82.2$_{\pm 0.2}$ & 76.0$_{\pm 2.4}$ & 77.6$_{\pm 0.7}$ & 65.7$_{\pm 0.6}$ & 64.1$_{\pm 0.7}$ & 68.0$_{\pm 1.2}$ & 45.1$_{\pm 0.8}$ & 66.4$_{\pm 0.9}$ \\
+ Linear Mapping & \bad 30.8$_{\pm 1.6}$ & \bad 69.3$_{\pm 0.6}$ & \bad 78.3$_{\pm 0.3}$ & \bad 60.2$_{\pm 0.8}$ & \bad 67.9$_{\pm 0.5}$ & \bad 58.2$_{\pm 0.7}$ & \bad 47.7$_{\pm 0.8}$ & \bad 54.1$_{\pm 0.3}$ & \bad 21.6$_{\pm 1.2}$ & \bad 54.2$_{\pm 0.8}$ \\
+ L2 Align & 44.1$_{\pm 1.2}$ & \bad 74.2$_{\pm 0.7}$ & \bad 81.9$_{\pm 0.3}$ & 74.9$_{\pm 3.3}$ & \bad 76.9$_{\pm 0.6}$ & \bad 64.7$_{\pm 0.5}$ & \bad 61.9$_{\pm 1.4}$ & 68.4$_{\pm 2.2}$ & \bad 42.1$_{\pm 1.1}$ & \bad 65.5$_{\pm 1.2}$ \\
+ Weak Align (Our) & \good 45.5$_{\pm 2.8}$ & 75.0$_{\pm 0.8}$ & 82.2$_{\pm 0.2}$ & \bad 73.7$_{\pm 1.8}$ & 77.3$_{\pm 0.6}$ & \good 66.6$_{\pm 1.3}$ & 64.0$_{\pm 1.2}$ & 67.5$_{\pm 1.4}$ & \bad 43.9$_{\pm 1.2}$ & 66.2$_{\pm 1.2}$ \\
+ Strong Align (Our) & 45.3$_{\pm 1.5}$ & 75.1$_{\pm 0.4}$ & 82.2$_{\pm 0.2}$ & 74.6$_{\pm 2.5}$ & 77.4$_{\pm 0.6}$ & 66.0$_{\pm 1.2}$ & 63.7$_{\pm 0.9}$ & 68.0$_{\pm 1.1}$ & \bad 43.3$_{\pm 0.4}$ & 66.2$_{\pm 1.0}$ \\

\midrule
XLM-R$_{\text{large}}$ & 46.8$_{\pm 4.3}$ & 79.1$_{\pm 0.5}$ & 84.2$_{\pm 0.2}$ & 75.7$_{\pm 2.9}$ & 80.7$_{\pm 0.5}$ & 71.6$_{\pm 1.1}$ & 71.7$_{\pm 0.5}$ & 77.4$_{\pm 1.3}$ & 51.5$_{\pm 1.4}$ & 71.0$_{\pm 1.4}$ \\

\midrule
\multicolumn{11}{l}{\textbf{POS (Accuracy)}} \\
\midrule

mBERT & 60.3$_{\pm 0.9}$ & 90.4$_{\pm 0.3}$ & 96.9$_{\pm 0.1}$ & 87.7$_{\pm 0.2}$ & 88.9$_{\pm 0.3}$ & 68.0$_{\pm 0.8}$ & 82.5$_{\pm 0.7}$ & 62.7$_{\pm 0.2}$ & 67.1$_{\pm 1.1}$ & 78.3$_{\pm 0.5}$ \\
+ Linear Mapping & \good 76.2$_{\pm 0.5}$ & \good 91.2$_{\pm 0.1}$ & \bad 96.3$_{\pm 0.0}$ & 87.6$_{\pm 0.1}$ & 89.0$_{\pm 0.2}$ & \good 74.9$_{\pm 1.1}$ & \bad 80.6$_{\pm 0.3}$ & \bad 60.4$_{\pm 0.5}$ & \bad 64.8$_{\pm 1.3}$ & \good 80.1$_{\pm 0.4}$ \\
+ L2 Align & \good 62.7$_{\pm 2.9}$ & \bad 89.5$_{\pm 0.8}$ & \bad 96.8$_{\pm 0.1}$ & \bad 87.1$_{\pm 0.3}$ & \bad 88.3$_{\pm 0.2}$ & \bad 65.2$_{\pm 3.7}$ & \good 83.8$_{\pm 1.0}$ & 62.8$_{\pm 0.5}$ & 67.3$_{\pm 1.1}$ & 78.2$_{\pm 1.2}$ \\
+ Weak Align (Our) & 61.1$_{\pm 1.3}$ & 90.4$_{\pm 0.8}$ & 96.9$_{\pm 0.0}$ & 87.7$_{\pm 0.5}$ & 88.7$_{\pm 0.3}$ & \good 70.3$_{\pm 1.2}$ & 83.2$_{\pm 0.6}$ & \good 63.3$_{\pm 0.3}$ & 68.0$_{\pm 0.5}$ & \good 78.8$_{\pm 0.6}$ \\
+ Strong Align (Our) & \good 61.7$_{\pm 1.7}$ & 90.5$_{\pm 0.7}$ & 96.9$_{\pm 0.0}$ & 87.7$_{\pm 0.6}$ & 88.7$_{\pm 0.4}$ & \good 70.5$_{\pm 1.0}$ & \good 83.3$_{\pm 0.7}$ & \good 63.1$_{\pm 0.3}$ & \good 68.2$_{\pm 0.8}$ & \good 79.0$_{\pm 0.7}$ \\

\midrule
XLM-R$_\text{base}$ & 70.2$_{\pm 1.6}$ & 91.6$_{\pm 0.3}$ & 97.5$_{\pm 0.0}$ & 88.5$_{\pm 0.2}$ & 89.4$_{\pm 0.3}$ & 71.7$_{\pm 1.3}$ & 86.1$_{\pm 0.3}$ & 64.5$_{\pm 0.5}$ & 71.4$_{\pm 0.5}$ & 81.2$_{\pm 0.6}$ \\
+ Linear Mapping & \good 76.0$_{\pm 0.9}$ & \good 92.0$_{\pm 0.1}$ & \bad 96.9$_{\pm 0.0}$ & 88.7$_{\pm 0.2}$ & 89.5$_{\pm 0.3}$ & \good 78.9$_{\pm 2.1}$ & \bad 83.9$_{\pm 0.3}$ & \bad 62.5$_{\pm 0.4}$ & \bad 66.5$_{\pm 1.0}$ & 81.7$_{\pm 0.6}$ \\
+ L2 Align & 71.0$_{\pm 0.9}$ & \bad 91.2$_{\pm 0.5}$ & \bad 97.3$_{\pm 0.0}$ & \bad 87.9$_{\pm 0.3}$ & \bad 88.8$_{\pm 0.4}$ & \good 74.8$_{\pm 2.9}$ & \good 86.9$_{\pm 0.8}$ & \bad 64.0$_{\pm 0.6}$ & \bad 70.6$_{\pm 0.5}$ & 81.4$_{\pm 0.8}$ \\
+ Weak Align (Our) & \good 72.5$_{\pm 0.8}$ & \bad 91.2$_{\pm 0.3}$ & \bad 97.4$_{\pm 0.0}$ & \bad 88.2$_{\pm 0.2}$ & 89.2$_{\pm 0.2}$ & 72.7$_{\pm 1.3}$ & 86.2$_{\pm 0.2}$ & 64.7$_{\pm 0.4}$ & 71.8$_{\pm 1.4}$ & 81.5$_{\pm 0.5}$ \\
+ Strong Align (Our) & \good 72.5$_{\pm 0.6}$ & \bad 91.2$_{\pm 0.2}$ & 97.4$_{\pm 0.1}$ & 88.3$_{\pm 0.2}$ & 89.2$_{\pm 0.2}$ & 72.0$_{\pm 1.9}$ & \good 86.5$_{\pm 0.2}$ & 64.8$_{\pm 0.4}$ & 71.7$_{\pm 1.7}$ & 81.5$_{\pm 0.6}$ \\

\midrule
XLM-R$_{\text{large}}$ & 73.9$_{\pm 1.0}$ & 91.9$_{\pm 0.3}$ & 98.0$_{\pm 0.0}$ & 89.2$_{\pm 0.2}$ & 89.8$_{\pm 0.1}$ & 78.4$_{\pm 2.1}$ & 86.5$_{\pm 0.2}$ & 64.8$_{\pm 0.3}$ & 71.0$_{\pm 0.3}$ & 82.6$_{\pm 0.5}$ \\

\midrule
\multicolumn{11}{l}{\textbf{Parsing (Labeled Attachment Score)}} \\
\midrule

mBERT & 28.8$_{\pm 0.4}$ & 67.8$_{\pm 0.5}$ & 79.7$_{\pm 0.1}$ & 69.1$_{\pm 0.1}$ & 73.3$_{\pm 0.2}$ & 31.0$_{\pm 0.5}$ & 60.2$_{\pm 0.6}$ & 33.5$_{\pm 0.5}$ & 29.5$_{\pm 0.4}$ & 52.6$_{\pm 0.4}$ \\
+ Linear Mapping & \good 45.0$_{\pm 0.3}$ & 67.7$_{\pm 0.2}$ & \good 80.5$_{\pm 0.2}$ & \good 70.0$_{\pm 0.3}$ & \good 73.9$_{\pm 0.2}$ & \bad 28.4$_{\pm 0.2}$ & \bad 57.2$_{\pm 0.4}$ & \bad 32.0$_{\pm 0.3}$ & \bad 28.1$_{\pm 0.2}$ & \good 53.6$_{\pm 0.3}$ \\
+ L2 Align & \good 29.7$_{\pm 0.6}$ & 67.7$_{\pm 0.7}$ & \bad 79.3$_{\pm 0.4}$ & \bad 68.9$_{\pm 0.6}$ & 73.4$_{\pm 0.5}$ & \good 31.7$_{\pm 1.8}$ & \good 61.3$_{\pm 1.2}$ & 33.6$_{\pm 0.5}$ & 29.7$_{\pm 0.2}$ & 52.8$_{\pm 0.7}$ \\
+ Weak Align (Our) & \good 29.9$_{\pm 1.0}$ & 67.6$_{\pm 0.4}$ & \good 79.8$_{\pm 0.0}$ & \good 69.6$_{\pm 0.3}$ & 73.5$_{\pm 0.5}$ & 31.0$_{\pm 1.6}$ & \good 61.2$_{\pm 0.9}$ & 33.4$_{\pm 0.7}$ & \good 30.0$_{\pm 0.5}$ & 52.9$_{\pm 0.6}$ \\
+ Strong Align (Our) & \good 30.8$_{\pm 0.9}$ & 68.0$_{\pm 0.4}$ & \good 79.8$_{\pm 0.1}$ & \good 69.9$_{\pm 0.3}$ & \good 73.7$_{\pm 0.5}$ & \good 31.5$_{\pm 1.5}$ & \good 61.8$_{\pm 0.6}$ & 33.5$_{\pm 0.6}$ & \good 30.4$_{\pm 0.4}$ & \good 53.3$_{\pm 0.6}$ \\

\midrule
XLM-R$_\text{base}$ & 43.7$_{\pm 1.7}$ & 69.0$_{\pm 0.4}$ & 80.5$_{\pm 0.2}$ & 71.0$_{\pm 0.4}$ & 73.6$_{\pm 0.5}$ & 41.2$_{\pm 0.9}$ & 66.3$_{\pm 0.9}$ & 36.6$_{\pm 0.2}$ & 34.2$_{\pm 0.7}$ & 57.3$_{\pm 0.6}$ \\
+ Linear Mapping & \good 48.0$_{\pm 0.5}$ & 69.2$_{\pm 0.2}$ & \good 81.4$_{\pm 0.1}$ & \good 72.4$_{\pm 0.1}$ & \good 74.8$_{\pm 0.3}$ & \bad 38.8$_{\pm 0.9}$ & \bad 61.8$_{\pm 0.5}$ & \bad 34.2$_{\pm 0.3}$ & \bad 24.2$_{\pm 0.9}$ & \bad 56.1$_{\pm 0.4}$ \\
+ L2 Align & \bad 39.4$_{\pm 0.5}$ & \bad 68.0$_{\pm 0.5}$ & \bad 79.9$_{\pm 0.2}$ & \bad 69.9$_{\pm 0.5}$ & \bad 72.8$_{\pm 0.5}$ & \bad 40.2$_{\pm 1.1}$ & \bad 63.8$_{\pm 0.8}$ & 36.4$_{\pm 0.6}$ & \bad 32.3$_{\pm 0.9}$ & \bad 55.9$_{\pm 0.6}$ \\
+ Weak Align (Our) & 44.5$_{\pm 1.3}$ & 68.7$_{\pm 0.7}$ & 80.4$_{\pm 0.1}$ & 71.3$_{\pm 0.3}$ & 73.8$_{\pm 0.3}$ & 41.4$_{\pm 0.8}$ & 65.7$_{\pm 0.4}$ & 36.7$_{\pm 0.4}$ & 34.0$_{\pm 0.7}$ & 57.4$_{\pm 0.5}$ \\
+ Strong Align (Our) & 44.9$_{\pm 1.0}$ & 68.8$_{\pm 0.6}$ & 80.4$_{\pm 0.1}$ & 71.2$_{\pm 0.2}$ & 73.8$_{\pm 0.2}$ & 41.1$_{\pm 0.8}$ & 65.9$_{\pm 0.5}$ & 36.6$_{\pm 0.3}$ & 33.9$_{\pm 0.7}$ & 57.4$_{\pm 0.5}$ \\

\midrule
XLM-R$_{\text{large}}$ & 48.2$_{\pm 1.5}$ & 67.8$_{\pm 0.6}$ & 82.6$_{\pm 0.3}$ & 73.9$_{\pm 0.4}$ & 76.4$_{\pm 0.4}$ & 41.8$_{\pm 2.5}$ & 69.6$_{\pm 0.4}$ & 38.9$_{\pm 0.6}$ & 35.4$_{\pm 0.5}$ & 59.4$_{\pm 0.8}$ \\

\bottomrule
\end{tabular}
}
\caption{Zero-shot cross-lingual transfer result with the OPUS-100 bitext. 
\textcolor{good}{Blue} or \textcolor{bad}{orange} indicates the mean performance is one standard derivation \textcolor{good}{above} or \textcolor{bad}{below} the mean of baseline.
\label{tab:signal-opus}}
\end{center}
\end{table*}

%% file: src/analysis.tex
\section{Introduction}

In \autoref{chap:crosslingual-signal}, we observe that While the generalization performance on the source language has low variance, on the target language the variance is much higher with zero-shot cross-lingual transfer, making it difficult to compare different models in the literature and meta-benchmark. Similarly, pretrained monolingual encoders also have unstable performance during fine-tuning \citep{devlin-etal-2019-bert,phang2018sentence}.

Why are these models so sensitive to the random seed? Many theories have bee offered: catastrophic forgetting of the pretrained task \citep{phang2018sentence,Lee2020Mixout,keung-etal-2020-dont}, small data size \citep{devlin-etal-2019-bert}, impact of random seed on task-specific layer initialization and data ordering \citep{dodge2020fine}, the Adam optimizer without bias correction \citep{mosbach2021on,zhang2021revisiting}, and a different generalization error with similar training loss \citep{mosbach2021on}. However, none of these factors fully explain the high generalization error variance of zero-shot cross-lingual transfer on target language but low variance on source language.

In this chapter, we offer a new explanation for high variance in target language performance: \textit{the zero-shot cross-lingual transfer optimization problem is under-specified}.
Based on the well-established linear interpolation of 1-dimensional plot and contour plot \citep{goodfellow2014qualitatively,li2018visualizing}, we empirically show that any linear-interpolated model between the monolingual source model and bilingual source and target model has equally low source language generation error. Yet the target language generation error surprisingly reduces smoothly and linearly as we move from a monolingual model to a bilingual model. To the best of our knowledge, no other paper documents this finding.

This result provides a new answer to our mystery: only a small subset of the solution space for the source language solves the target language on par with models with actual target language supervision; the optimization could not find such a solution without target language supervision, hence an under-specified optimization problem. If target language supervision were available, as it was in the counterfactual bilingual model, the optimization finds the smaller subset. By comparing both mBERT and XLM-R, we find that the generalization error surface of XLM-R is flatter than mBERT, contributing to its better performance compared to mBERT. Thus, zero-shot cross-lingual transfer has high variance, as the solution found by zero-shot cross-lingual transfer lies in the non-flat region of the target language generalization error surface.

\section{Existing Hypotheses}
Prior studies have observed encoder model instability, and have offered various hypotheses to explain this behavior.
Catastrophic forgetting -- when neural networks trained on one task forget that task after training on a second task 
 \cite{McCloskey1989CatastrophicII,kirkpatrick2017overcoming}
---has been credited as the source of high variance in both monolingual fine-tuning \cite{phang2018sentence,Lee2020Mixout} and zero-shot cross-lingual transfer \cite{keung-etal-2020-dont}.  \citet{mosbach2021on} wonder why preserving cloze capability is important. In zero-shot cross-lingual transfer, deliberately preserving the multilingual cloze capability with regularization improves performance but does not eliminate the zero-shot transfer gap \cite{aghajanyan2021better,liu-etal-2021-preserving}.

Small training data size often seems to have higher variance in performance \cite{devlin-etal-2019-bert}, but \citet{mosbach2021on} found that when controlling the number of gradient updates, smaller data size has the similar variance as larger data size. 

In the pretraining-then-fine-tune paradigm, random seeds mainly impact the initialization of task-specific layers and data ordering during fine-tuning. \citet{dodge2020fine} show development set performance has high variance with respect to seeds. Additionally, Adam optimizer without bias correction---an Adam \citep{kingma2014adam} variant (inadvertently) introduced by the implementation of \citet{devlin-etal-2019-bert}---has been identified as the source of high variance during monolingual fine-tuning \citep{mosbach2021on,zhang2021revisiting}. However, in zero-shot cross-lingual transfer, while different random seeds lead to high variance in target languages, the source language has much smaller variance in comparison even with standard Adam \citep{wu-dredze-2020-explicit}.

Beyond optimizers, \citet{mosbach2021on} attributes high variance to generalization issues: despite having similar training loss, different models exhibit vastly different development set performance. However, in zero-shot cross-lingual transfer, the development or test performance variance is much smaller on the source language compared to target language.

\section{Zero-shot Cross-lingual Transfer is Under-specified Optimization}
\label{sec:underspecified}

Existing hypotheses do not explain the high variance of zero-shot cross-lingual transfer:
much higher variance on generalization error of the target language compared to the source language. 
We propose a new explanation: \textit{zero-shot cross-lingual transfer is an under-specified optimization problem}. 
Optimizing a multilingual model for a specific task using only source language annotation allows choices of many good solutions in terms of generalization error. However, unbeknownst to the optimizer, these solutions have wildly different generalization errors performance on the target language. In fact, a small subset has similar low generalization error as models trained on target language. Yet without the guidance of target data, the zero-shot cross-lingual optimization could not find this smaller subset. As we will show in \cref{sec:analysis-result}, the solution found by zero-shot transfer lies in a non-flat region of target language generalization error, causing its high variance.

\subsection{Linear Interpolation}\label{sec:interpolation}
We test this hypothesis via a linear interpolation between two models to explore the neural network parameter space.
Consider three sets of neural network parameters: $\theta_{src}$, $\theta_{tgt}$, $\theta_{\{src,tgt\}}$ for a model trained on task data for the source language only, target language only and both languages, respectively. This includes both task-specific layers and encoders.\footnote{We experiment with interpolating the encoder parameters only and observe similar findings. On the other hand, interpolating the  task-specific layer only has a negligible effect.}
Note all three models have the same initialization before fine-tuning, making the bilingual model a counterfactual setup if the corresponding target language supervision was available.
We obtain the 1-dimensional (1D) linear interpolation of a monolingual (source) task trained model and bilingual task trained model with
\begin{align}\label{eq:src-bi}
\theta(\alpha) = \alpha\theta_{\{src,tgt\}} + (1-\alpha) \theta_{src}
\end{align}
or we could swap source and target by
\begin{align}\label{eq:tgt-bi}
\theta(\alpha) = \alpha\theta_{\{src,tgt\}} + (1-\alpha) \theta_{tgt}
\end{align}
where $\alpha$ is a scalar mixing coefficient \cite{goodfellow2014qualitatively}. 
Additionally, we can compute a 2-dimensional linear interpolation as
\begin{align}\label{eq:tri}
\theta(\alpha_1, \alpha_2) = \theta_{\{src,tgt\}} + \alpha_1 \delta_{src} + \alpha_2 \delta_{tgt} 
\end{align}

\begin{figure*}[]
\centering
\includegraphics[width=\columnwidth]{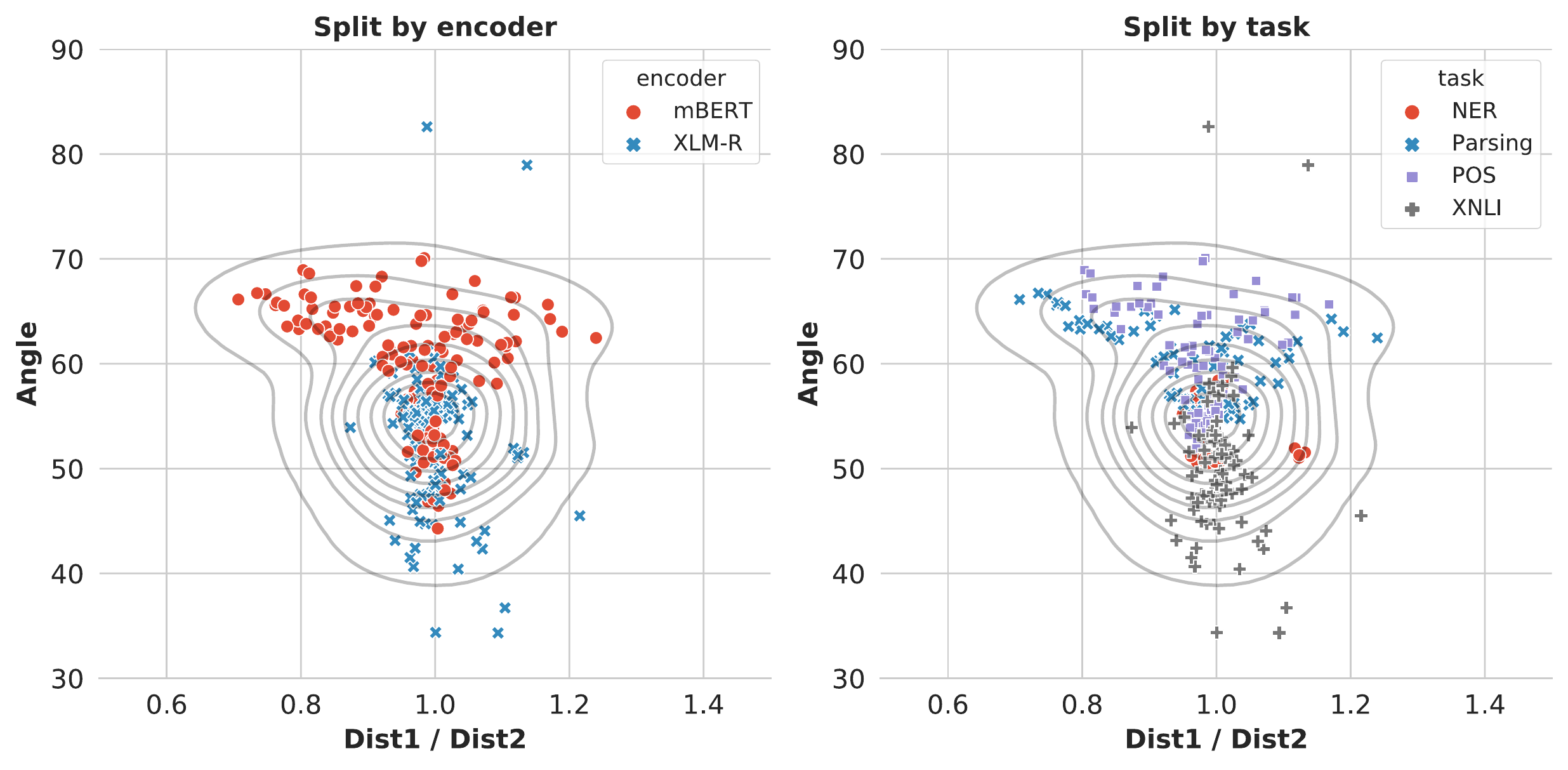}
\caption{
$\|\delta_{src}\|/\|\delta_{tgt}\|$ v.s. angle between $\delta_{src}$ and $\delta_{tgt}$.
Most $\delta_{src}$ and $\delta_{tgt}$ have similar norms, and the angle between them is around 55\degree.
}
\label{fig:dist-vs-angle}
\end{figure*}

where $\delta_{src} = \theta_{src}-\theta_{\{src,tgt\}}$, $\delta_{tgt} = \theta_{tgt}-\theta_{\{src,tgt\}}$, $\alpha_1$ and $\alpha_2$ are scalar mixing coefficients \cite{li2018visualizing}.\footnote{\citet{li2018visualizing} use two random directions and they normalize it to compensate scaling issue. In this setup, we find $\delta_{src}$ and $\delta_{tgt}$ have near identical norms, so we do not apply additional normalization. As these two directions are not random, we find that it spans around 55\degree. We plot the norm ratio and angle of these two vectors in \autoref{fig:dist-vs-angle}.} Finally, we can evaluate any interpolated models on the development set of source and target languages, testing the generalization error on the same language and across languages. 

The performance of the interpolated model illuminates the behavior of the model's parameters.
Take \autoref{eq:src-bi} as an example: if the linear interpolated model performs consistently high for our task on the source language, it suggests that both models lie within the same local minimum of source language generalization error surface. Additionally, if the linear interpolated model performs vastly differently on the target language, it would support our hypothesis. On the other hand, if the linear interpolated model performance drops on the source language, it suggests that both models lie in different local minimum of source language generalization error surface, suggesting the zero-shot optimization searching the wrong region.

\section{Experiments}

We consider four tasks: natural language inference \citep[XNLI;][]{conneau-etal-2018-xnli}, named entity recognition \citep[NER;][]{pan-etal-2017-cross}, POS tagging and dependency parsing \citep{ud2.7}.
We evaluate XNLI and POS tagging with accuracy (ACC), NER with span-level F1, and parsing with labeled attachment score (LAS).
We consider two encoders: base mBERT and large XLM-R.
For the task-specific layer, we use a linear classifier for XNLI, NER, and POS tagging, and \citet{dozat2016deep} for dependency parsing. 

To avoid English-centric experiments, we consider two source languages: English and Arabic. We choose 8 topologically diverse target languages: Arabic\footnote{Arabic is only used when English is the source language.}, German, Spanish, French, Hindi, Russian, Vietnamese, and Chinese.
We train the source language only and target language only monolingual model as well as a source-target bilingual model. 

We compute the linear interpolated models as described in \autoref{sec:interpolation} and test it on both the source and target language development set.
We loop over $\{-0.5,-0.4,\cdots,1.5\}$ for $\alpha$, $\alpha_1$ and $\alpha_2$.\footnote{We additionally select 0.025, 0.05, 0.075, 0.125, 0.15, 0.175, 0.825, 0.85, 0.875, 0.925, 0.95, and 0.975 for $\alpha$ due to preliminary experiment.}
We report the mean and variance of three runs by using different random seeds. We normalized both mean and variance of each interpolated model by the bilingual model performance, allowing us to aggregate across tasks and language pairs.

We follow the implementation and hyperparameter of \autoref{chap:crosslingual-signal}. We optimize with Adam \citep{kingma2014adam}. The learning rate is $\texttt{2e-5}$. The learning rate scheduler has 10\% steps linear warmup then linear decay till 0. We train for 5 epochs and the batch size is 32. For token level tasks, the task-specific layer takes the representation of the first subword, following previous chapters. Model selection is done on the corresponding dev set of the training set.

During fine-tuning, the maximum sequence length is 128. We use a sliding window of context to include subwords beyond the first 128 for NER and POS tagging. At test time, we use the same maximum sequence length with the exception of parsing, where the first 128 words instead of subwords of a sentence were used. We ignore words with POS tags of \texttt{SYM} and \texttt{PUNCT} during parsing evaluation. For NER, we adapt the same post-processing as \autoref{sec:task-ner}.
For POS tagging and dependency parsing, we use the following treebanks: Arabic-PADT, German-GSD, English-EWT, Spanish-GSD, French-GSD, Hindi-HDTB, Russian-GSD, Vietnamese-VTB, and Chinese-GSD.
Since the Chinese NER is labeled on character-level (including code-switched portion), we segment the Chinese character into word using the Stanford Word Segmenter and realign the label.

\section{Findings} \label{sec:analysis-result}

\begin{figure*}[]
\centering
\includegraphics[width=\columnwidth]{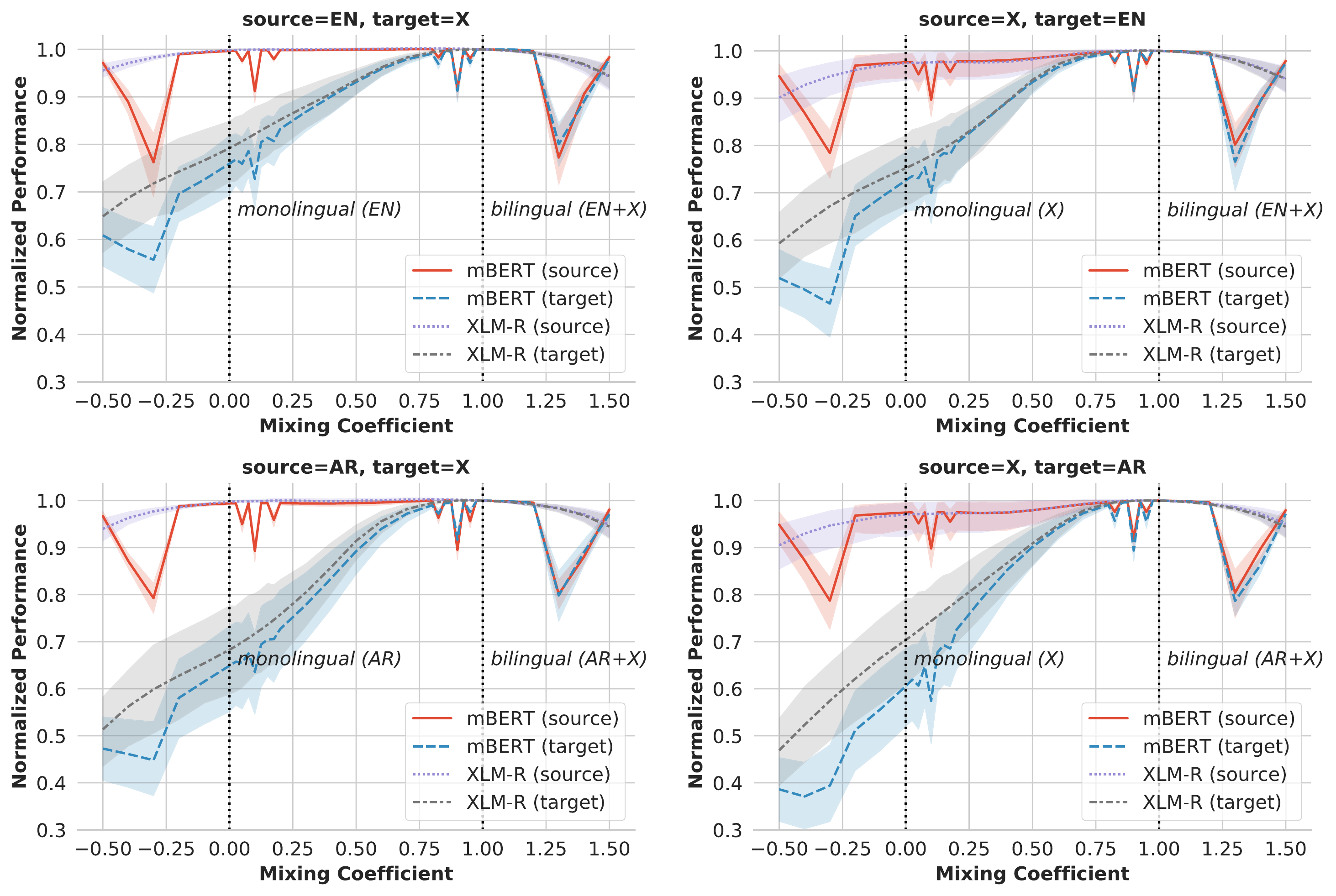}
\caption{Normalized performance of a linear interpolated model between a monolingual and bilingual model. A single plot line shows the performance normalized by the matching bilingual model and aggregated over eight language pairs and four tasks, with the shaded region representing 95\% confidence interval. 
The x-axis is the linear mixing coefficient $\alpha$ in \autoref{eq:src-bi} and \autoref{eq:tgt-bi}, with $\alpha=0$ and $\alpha=1$ representing source language monolingual model and source + target bilingual model, respectively.
To allow aggregating, for each encoder, language pair and task combination, we normalized the interpolated model performance by its corresponding bilingual performance.
Each subfigure title indicates the source and target languages.
Across all experiments, the source language dev performance stays consistently high (red and purple lines) during interpolation while the target language dev performance starts low and increases smoothly and linearly as it moves towards the bilingual model (gray and blue lines).
Break down of this figure by tasks can be found in 
\autoref{fig:interpolation-mean-ner} (NER), \autoref{fig:interpolation-mean-parsing} (Parsing), \autoref{fig:interpolation-mean-pos} (POS), and \autoref{fig:interpolation-mean-xnli} (XNLI), and we observe similar findings.}
\label{fig:interpolation-mean}
\end{figure*}

\begin{figure*}[]
\centering
\includegraphics[width=\columnwidth]{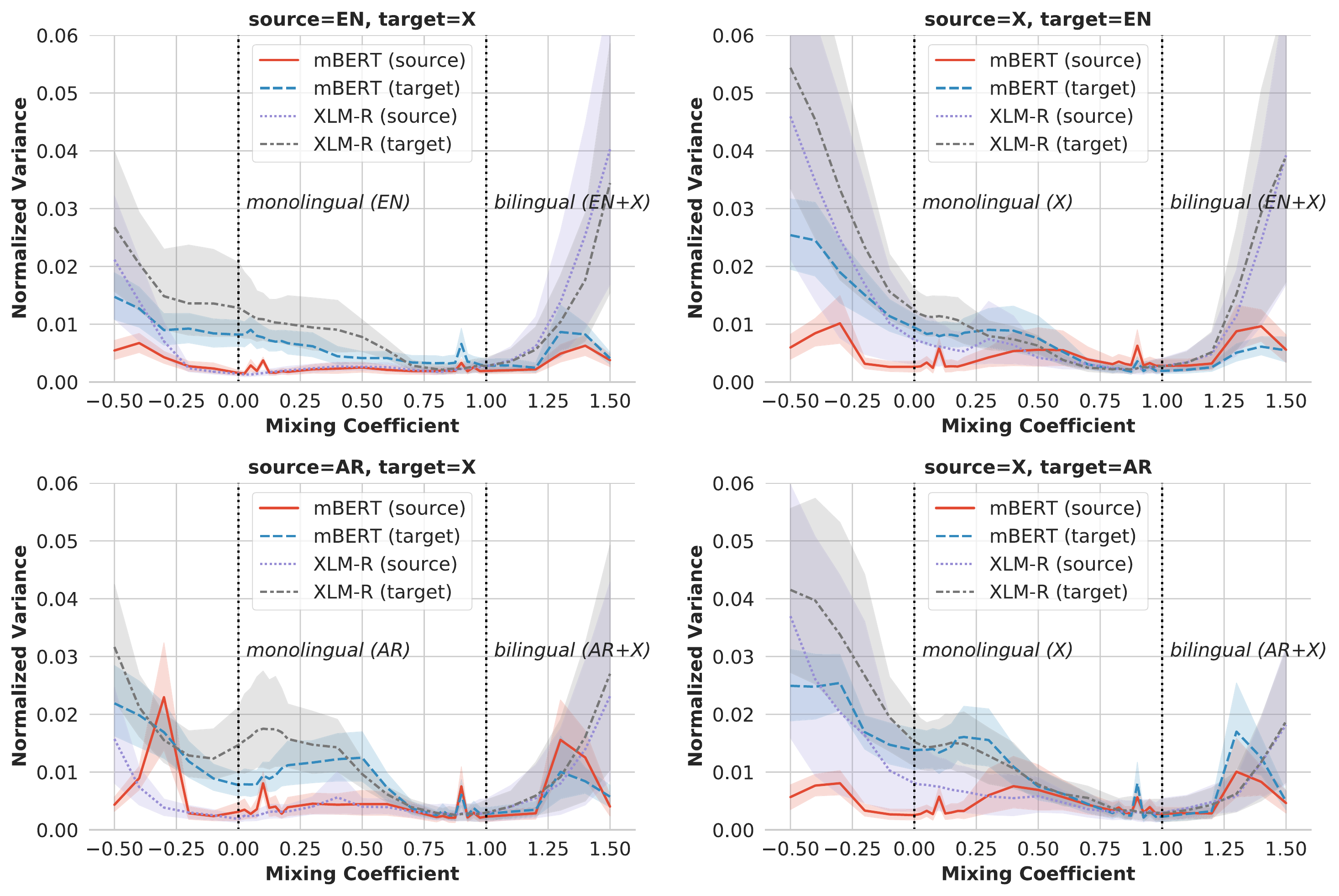}
\caption{
Normalized variance of linear interpolation between monolingual model and bilingual model.
The source language has much lower variance compared to target language on the monolingual side of the interpolated models.}
\label{fig:interpolation-std}
\end{figure*}

\begin{figure*}[]
\centering
\subfloat[][NER]{
\includegraphics[width=0.9\columnwidth]{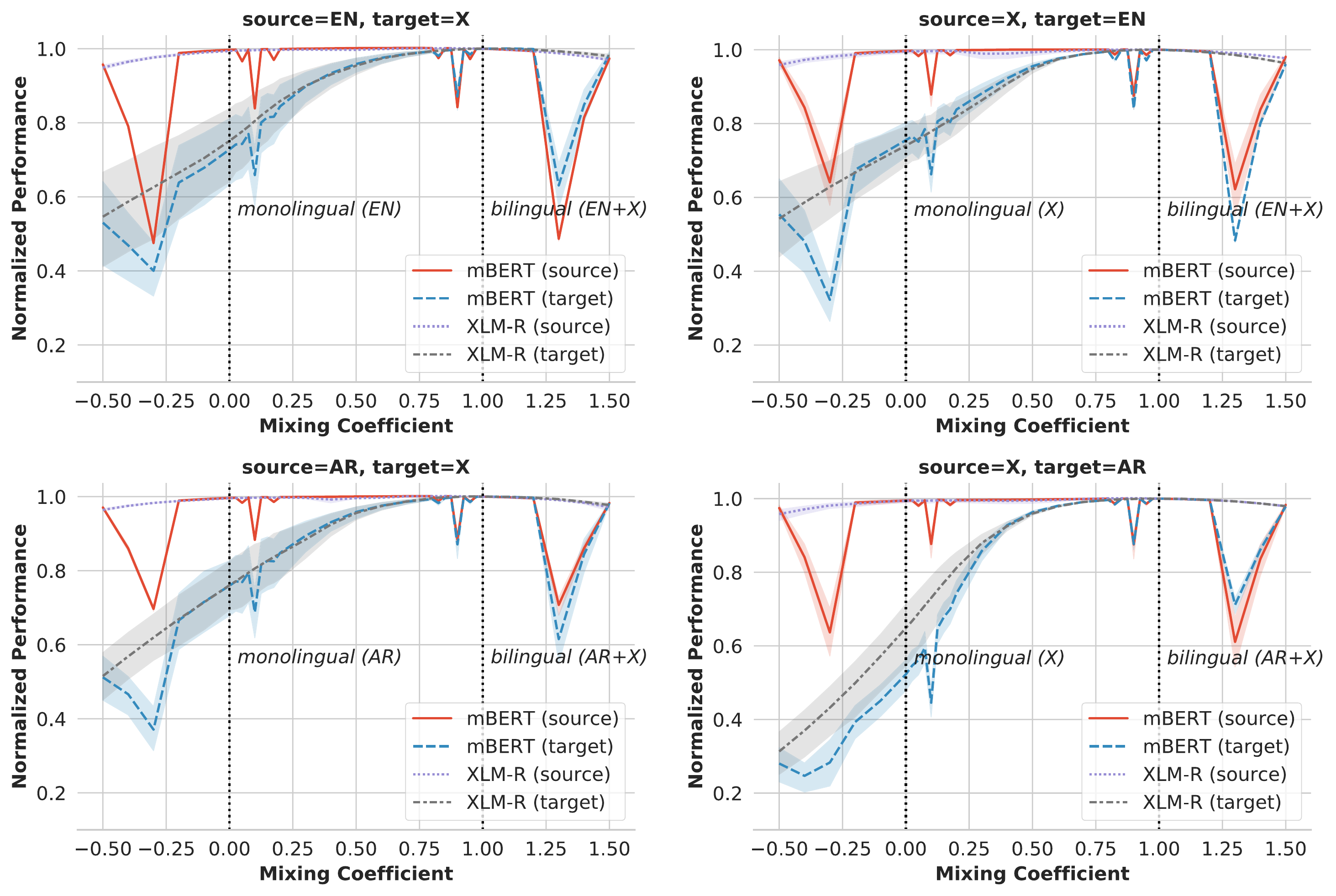}
\label{fig:interpolation-mean-ner}
}

\subfloat[][Parsing]{
\includegraphics[width=0.9\columnwidth]{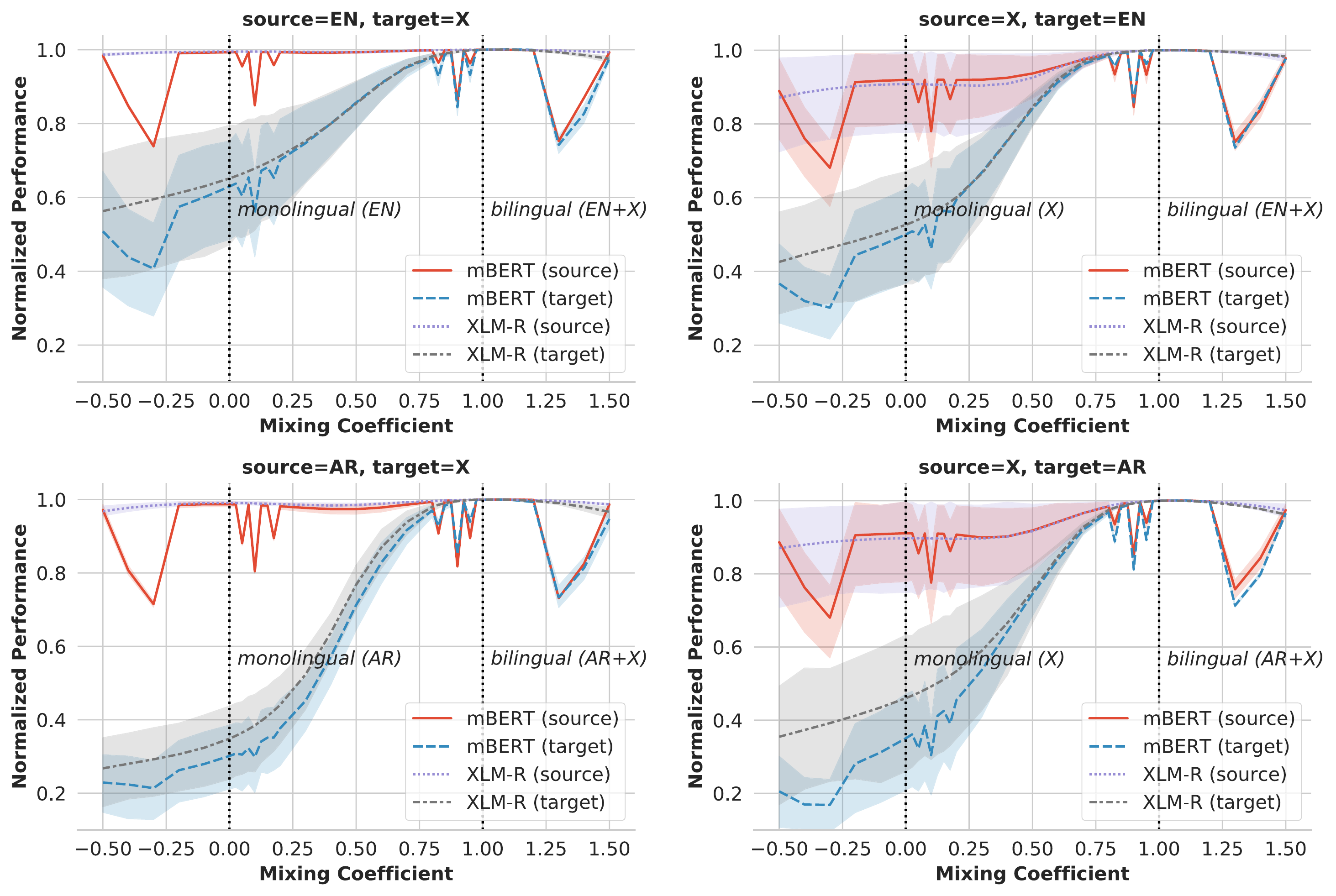}
\label{fig:interpolation-mean-parsing}
}
\caption{Normalized NER and Parsing performance of linear interpolated model between monolingual and bilingual model}
\end{figure*}

\begin{figure*}[]
\centering
\subfloat[][POS]{
\includegraphics[width=0.9\columnwidth]{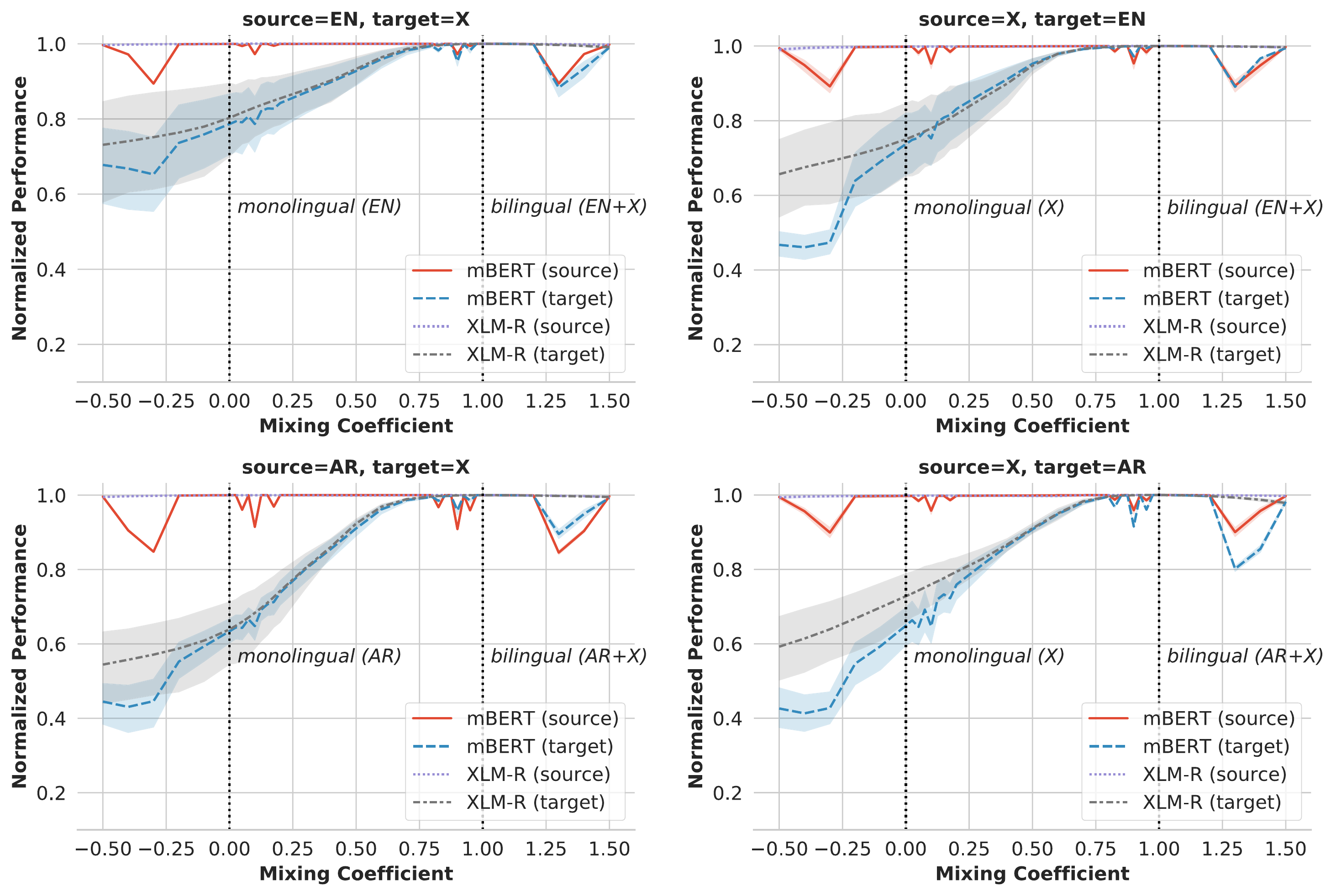}
\label{fig:interpolation-mean-pos}
}

\subfloat[][XNLI]{
\includegraphics[width=0.9\columnwidth]{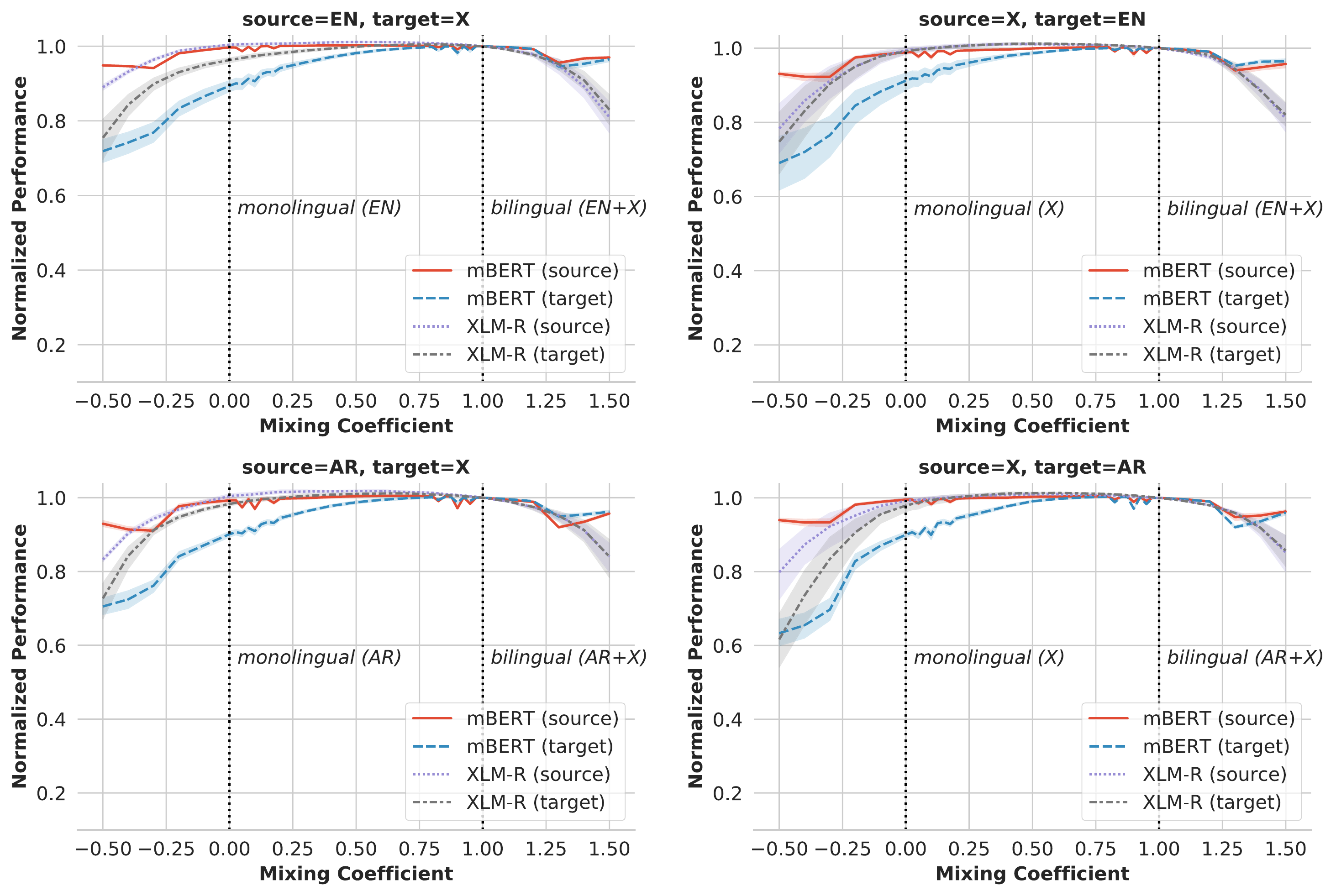}
\label{fig:interpolation-mean-xnli}
}
\caption{Normalized POS and XNLI performance of linear interpolated model between monolingual and bilingual model}
\end{figure*}

In \autoref{fig:interpolation-mean}, we observe that interpolations between the source monolingual and bilingual model have consistently similar source language performance. In contrast, surprisingly, the target language performance smoothly and linearly improves as the interpolated model moves from the zero-shot model to bilingual model.\footnote{We also show the variance of the interpolated models in \autoref{fig:interpolation-std}. The source language has much lower variance compared to target language on the monolingual side of the interpolated models, echoing findings in \autoref{chap:crosslingual-signal}.} Break down of \autoref{fig:interpolation-mean} by task shows different tasks follow similar trends. The only exception is mBERT, where the performance drops slightly around 0.1 and 0.9 locally. In contrast, XLM-R has a flatter slope and smoother interpolated models.

\begin{figure*}[]
\centering
\subfloat[][EN-HI Parsing with mBERT]{
\includegraphics[width=0.24\columnwidth]{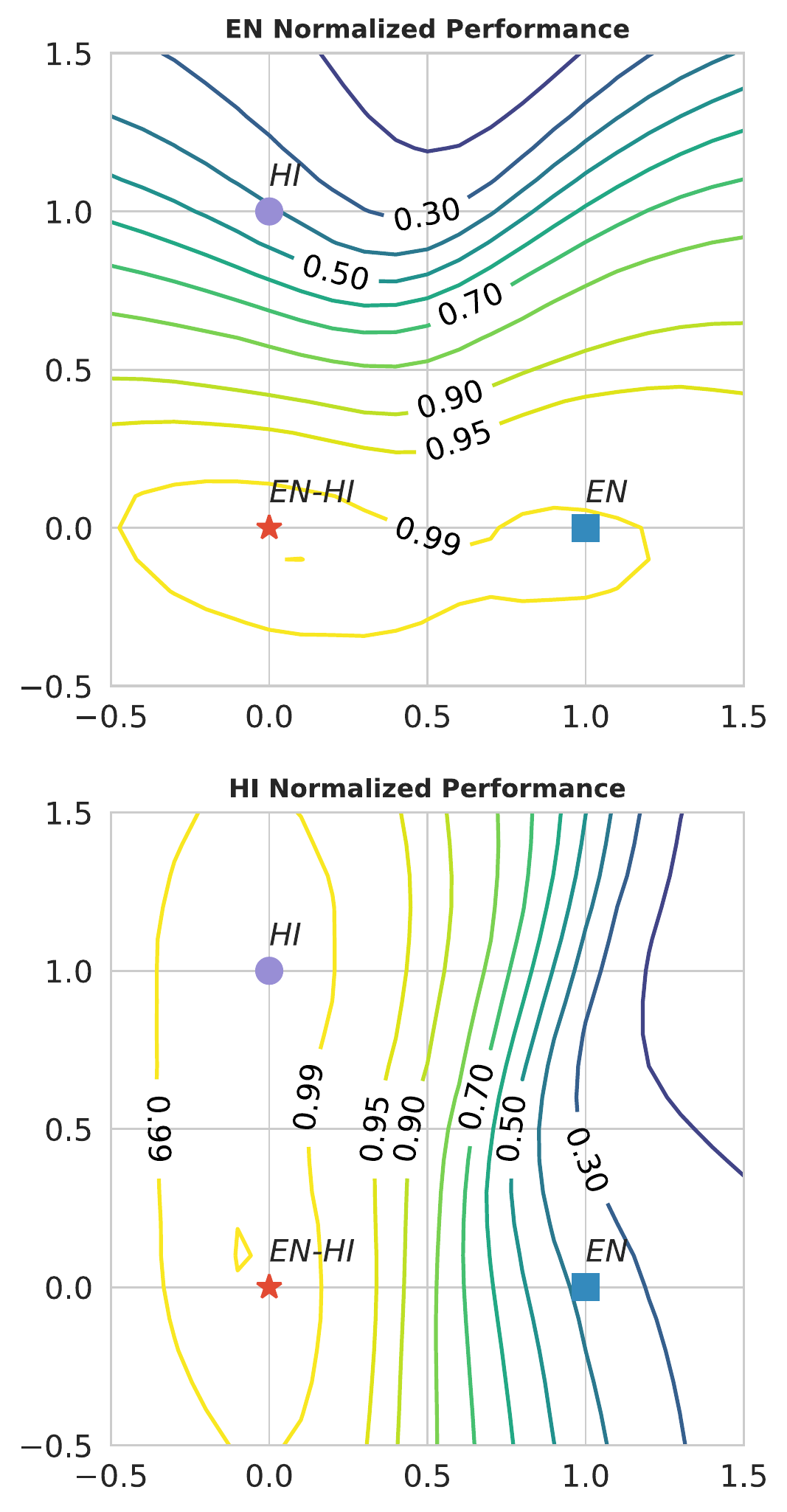}
}
\subfloat[][EN-HI Parsing with XLM-R]{
\includegraphics[width=0.24\columnwidth]{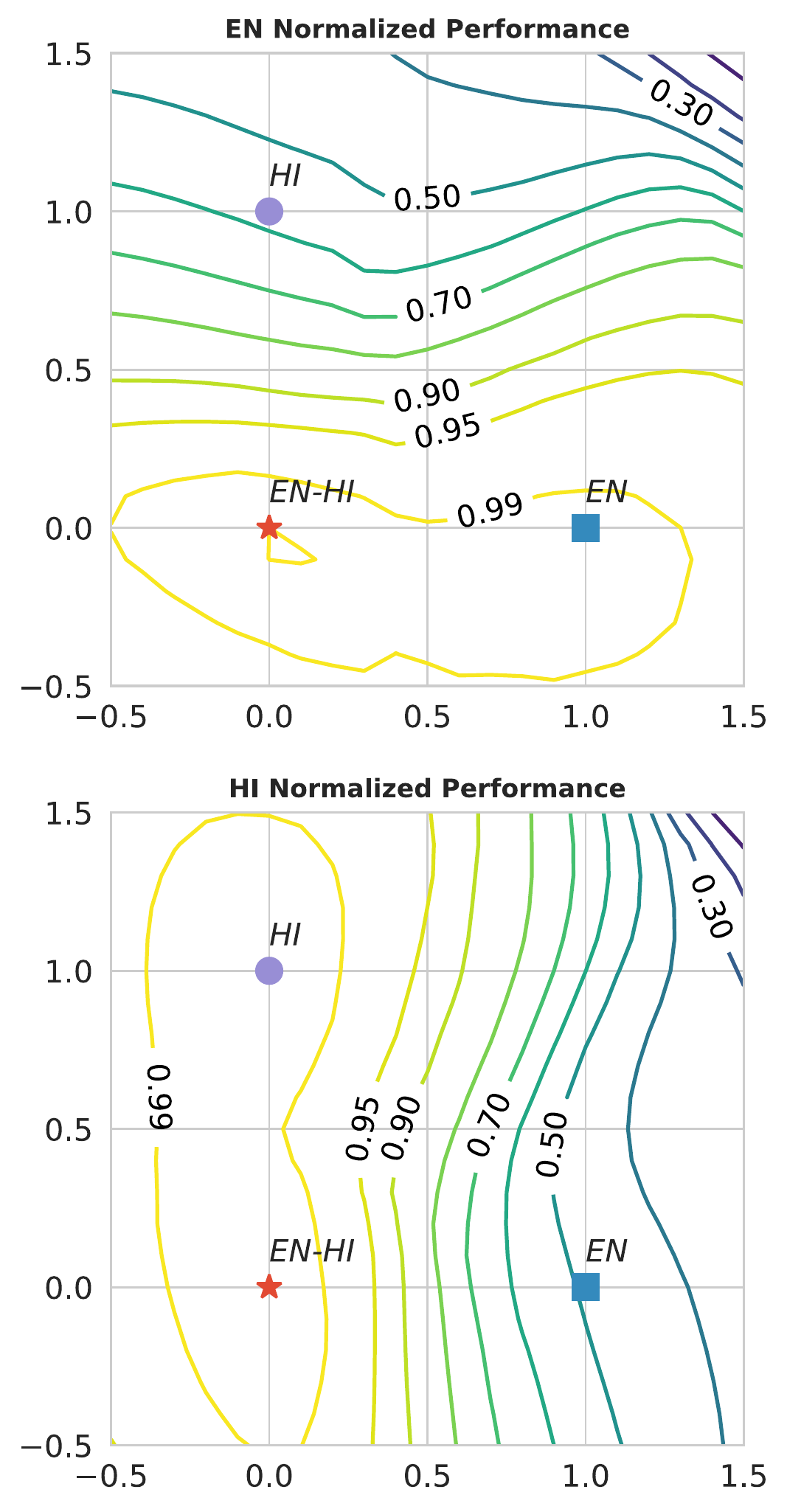}
}
\subfloat[][EN-RU NER with mBERT]{
\includegraphics[width=0.24\columnwidth]{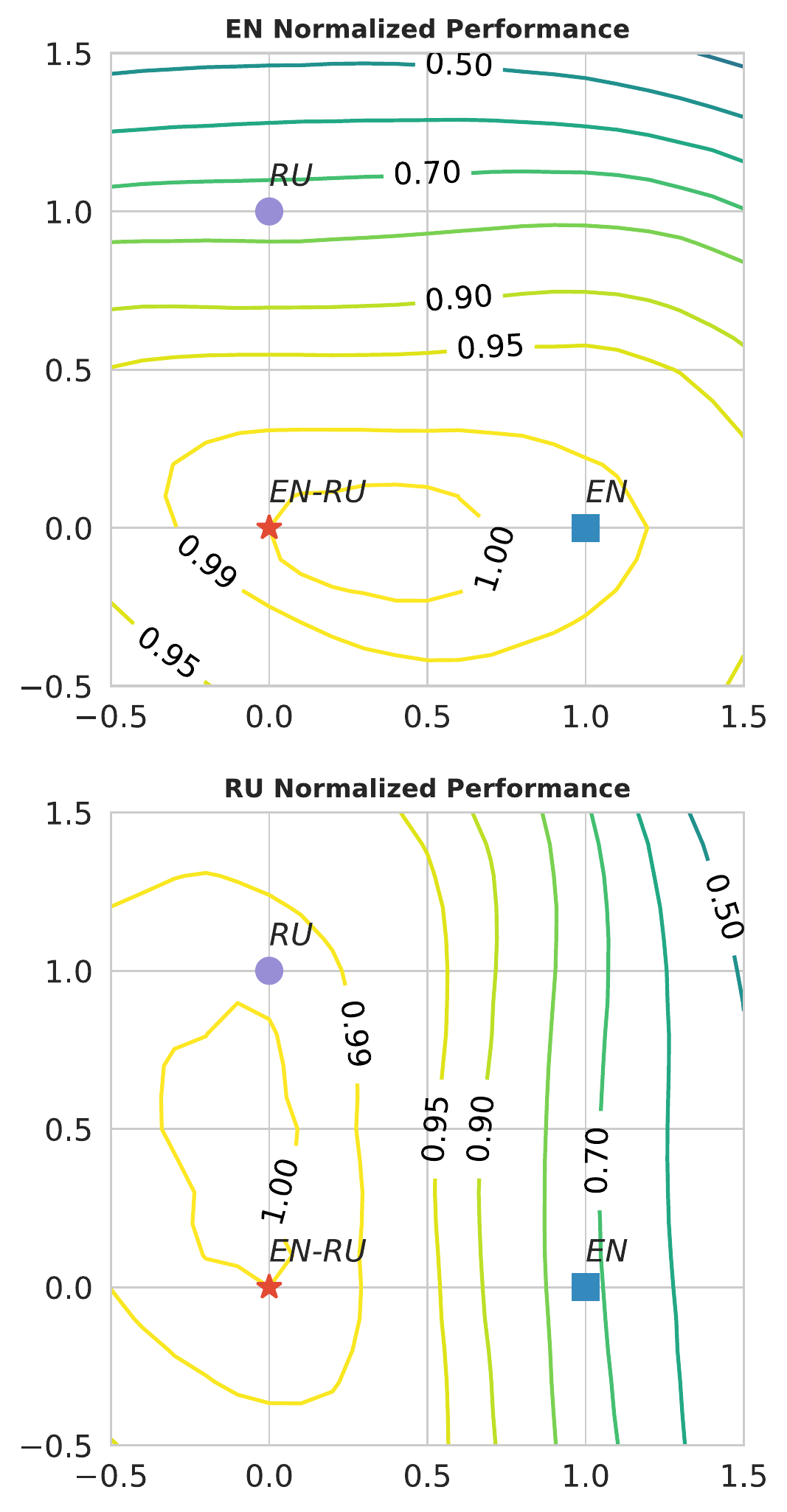}
}
\subfloat[][EN-RU NER with XLM-R]{
\includegraphics[width=0.24\columnwidth]{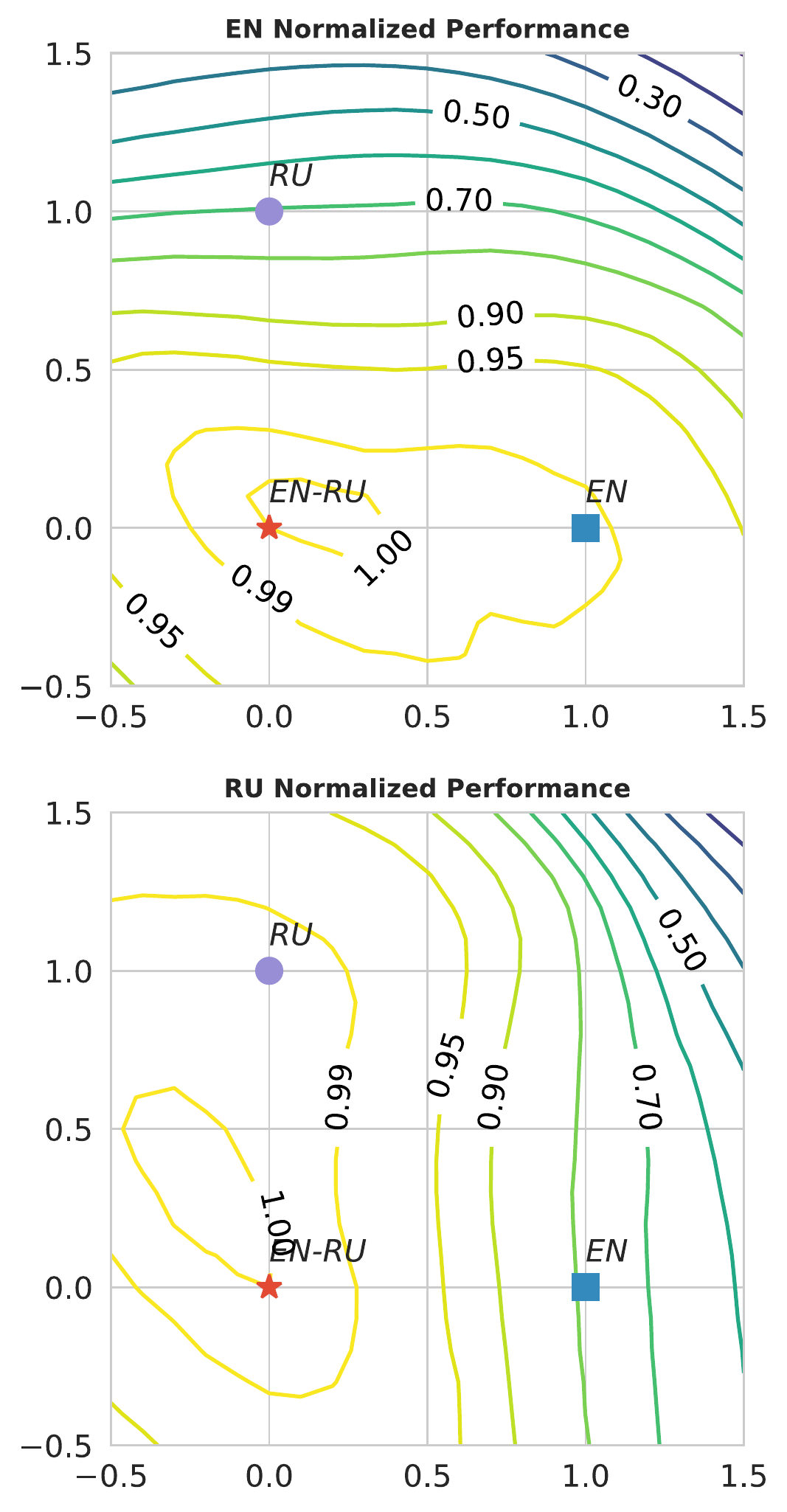}
}

\subfloat[][AR-DE POS with mBERT]{
\includegraphics[width=0.24\columnwidth]{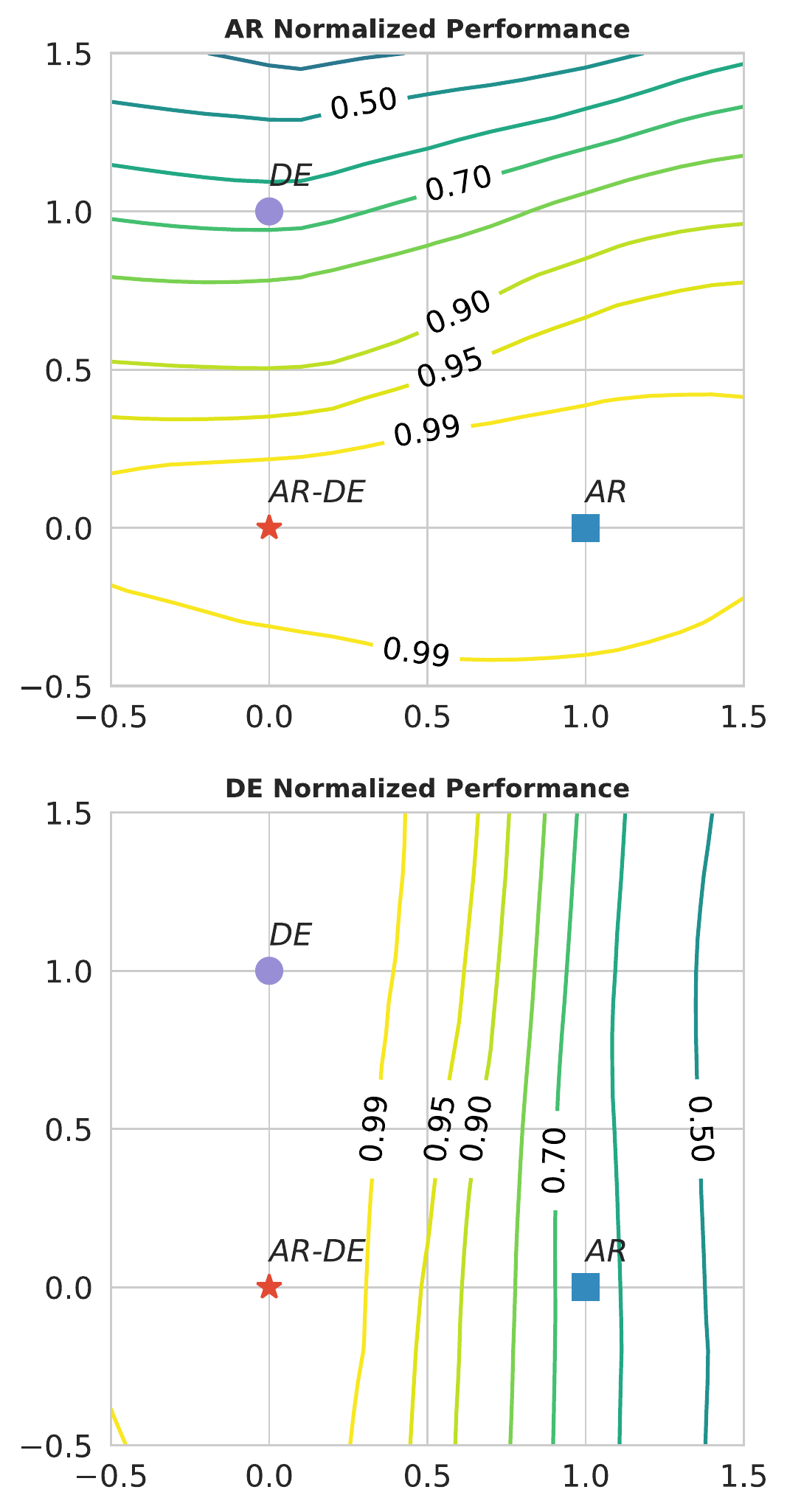}
}
\subfloat[][AR-DE POS with XLM-R]{
\includegraphics[width=0.24\columnwidth]{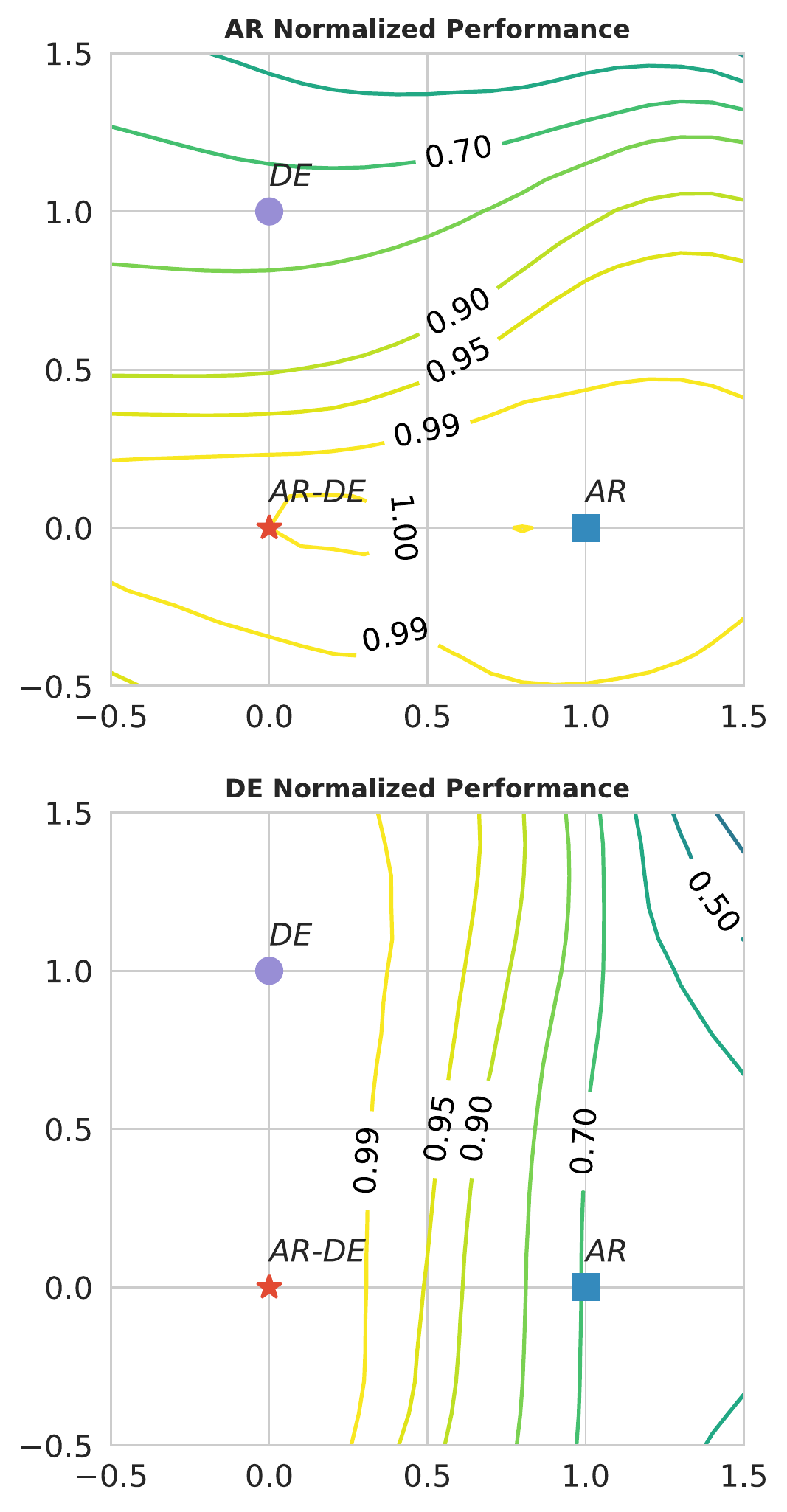}
}
\subfloat[][AR-ZH XNLI with mBERT]{
\includegraphics[width=0.24\columnwidth]{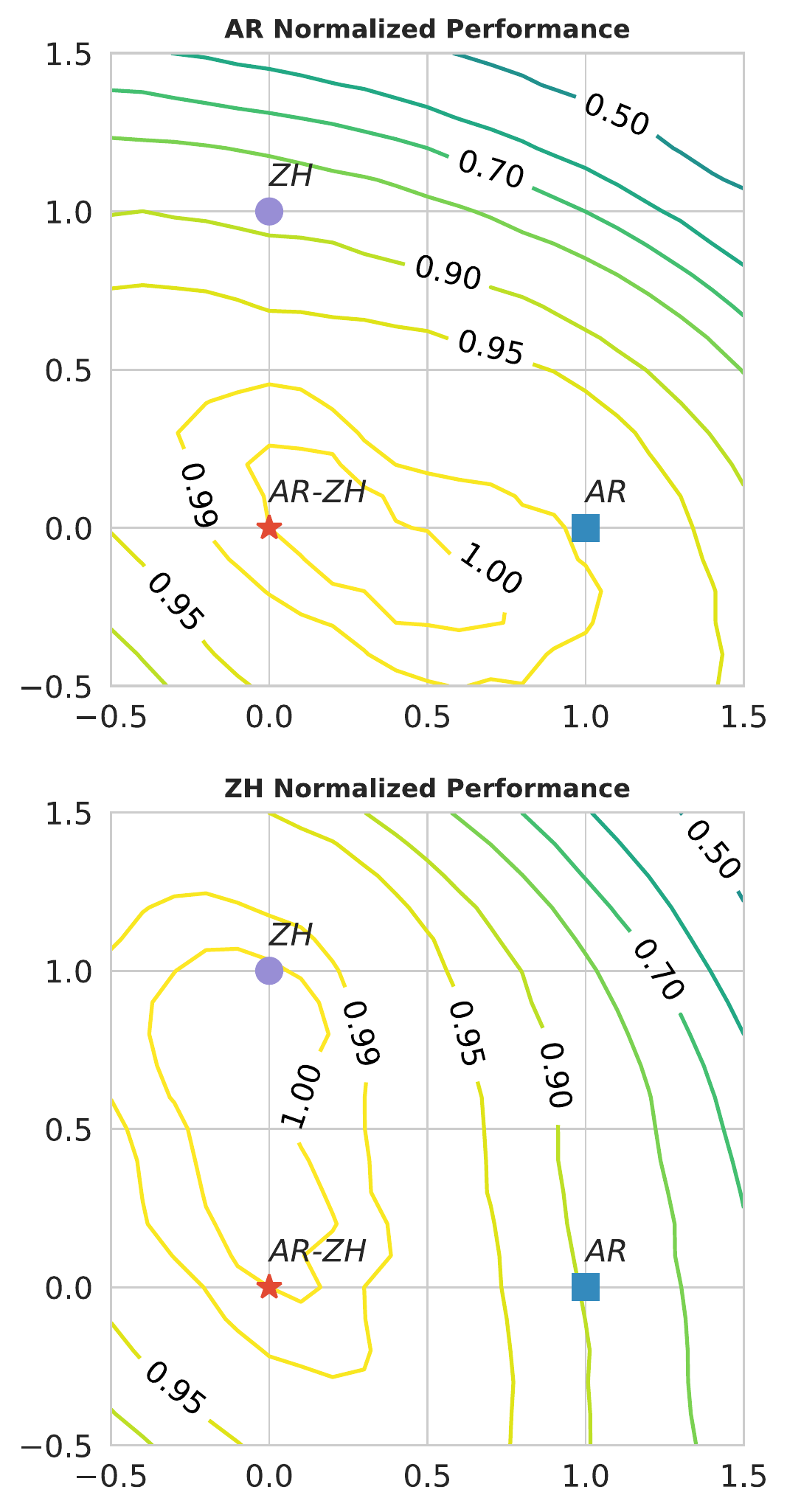}
}
\subfloat[][AR-ZH XNLI with XLM-R]{
\includegraphics[width=0.24\columnwidth]{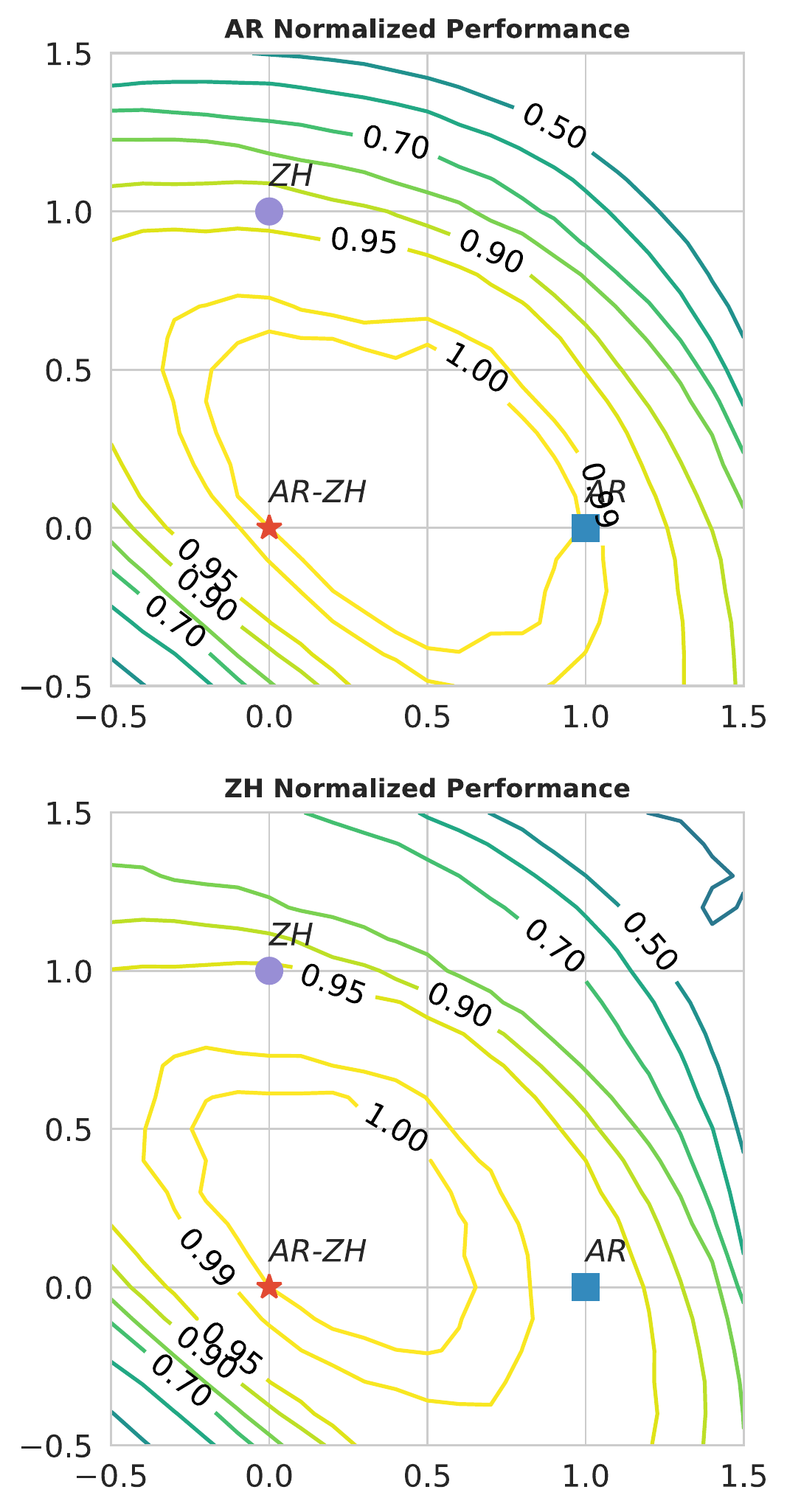}
}

\caption{Normalized performance of 2D linear interpolation between bilingual model and monolingual models.
The x-axis and the y-axis are the $\alpha_1$ and $\alpha_2$ in \autoref{eq:tri}, respectively.
By comparing mBERT and XLM-R, we observe that XLM-R has a flatter target language generalization error surface compared to mBERT.
Different language pairs and tasks combination shows similar trends.
}
\label{fig:interpolation-2d}
\end{figure*}

\autoref{fig:interpolation-2d} further demonstrates this finding with a 2D linear interpolation. The generalization error surface of the target language of XLM-R is much flatter compared to mBERT, perhaps the fundamental reason why XLM-R performs better than mBERT in zero-shot transfer, similar to findings in other computer vision models \cite{li2018visualizing}. As we discuss in \autoref{sec:underspecified}, these two findings support our hypothesis that zero-shot cross-lingual transfer is an under-specified optimization problem. As \cref{fig:interpolation-2d} shows, the solution found by zero-shot transfer lies in a non-flat region of target language generalization error surface, causing the high variance of zero-shot transfer on the target language. In contrast, the same solution lies in a flat region of source language generalization error surface, causing the low variance on the source language.

\section{Discussion}\label{sec:analysis-discussion}

In this chapter, we have presented evidence that zero-shot cross-lingual transfer is an under-specified optimization problem, and the cause of high variance on target language but not the source language tasks during zero-shot cross-lingual transfer. This finding holds across 4 tasks, 2 source languages and 8 target languages.
While this chapter focuses on zero-shot cross-lingual transfer, similar high variance has been observed in cross-lingual transfer with silver data (\autoref{chap:data-projection}) and few-shot cross-lingual transfer \cite{zhao-etal-2021-closer}, despite outperforming zero-shot cross-lingual transfer. It suggests that they are likely solving a similar under-specified optimization problem, due to the quality of the silver data or the variance of few-shot data selection impacting the gradient direction.

Therefore, addressing this issue may yield significant improvements to zero-shot cross-lingual transfer. 
Training bigger encoders addresses this issue indirectly by producing encoders with flatter cross-lingual generalization error surfaces. However, a more robust solution may be found by introducing constraints into the optimization problem that directly addresses the under-specification of the optimization. %

Silver target data is a potential way to further constrain the optimization problem. Silver target data can be created with machine translation and automatically labeling by either data projection or self training. In \autoref{chap:data-projection}, we will explore using silver target data to constrain the optimization problem and improve cross-lingual transfer performance.

Similarly, few-shot cross-lingual transfer is a potential way to further constrain the optimization problem. \citet{zhao-etal-2021-closer} finds that few-shot overall improves over zero-shot and it is important to first train on source language then fine-tune with the few-shot target language example. Through the lens of our analysis, this finding is intuitive since fine-tuning with a small amount of target data provides a guidance (gradient direction) to narrow down the solution space, leading to a potentially better solution for the target language. The initial fine-tuning with the source data is also important since it provides a good starting point. Additionally, \citet{zhao-etal-2021-closer} observes that the choice of shots matters. This is expected as it significantly impacts the quality of the gradient direction.

Unsupervised model selection like \citet{chen2020model} and optimization regularization like \citet{aghajanyan2021better} have been proposed in the literature to improve zero-shot cross-lingual transfer. Through the lens of our analysis, both solutions attempt to constrain the optimization problem. As none of the existing techniques fully constrain the optimization, future work should study the combination of existing techniques and develop new techniques on top of it instead of studying one technique at a time.

%% file: src/data-projection.tex
\section{Introduction}

In \autoref{chap:analysis}, we identify that zero-shot cross-lingual transfer is under-specified optimization. Additionally, performance on the target language in zero-shot cross-lingual transfer is often far below that of within-language supervision, especially in structured prediction tasks \cite{ruder-etal-2021-xtreme}. To address these challenges, additional constraints need to be added to the optimization problem. One way to achieve this is to add a learning signal for the target language. However, in zero-shot cross-lingual transfer, no target language supervision is available. Thus, we consider data projection and self-training. Before the advent of cross-lingual representations, such as in multilingual word embeddings and mBERT, cross-lingual transfer was approached largely as a data projection problem: one either translated and aligned the source training data to the target language, or at test time one translated target language inputs to the source language for prediction \cite{yarowsky-ngai-2001-inducing}. Instead of obtaining the label by alignment and projection, we could also obtain the label using the zero-shot model, similar to  traditional self-training \cite{yarowsky-1995-unsupervised}.

We show that by augmenting the source language training data with ``silver'' data in the target language---either via projection of the source data to the target language or via self-training with translated text---zero-shot performance can be improved, providing constraints to the optimization. Further improvements might come from using better pretrained encoders or improving on a projection strategy through better automatic translation models or better alignment models. In this chapter, we explore all the options above, finding that \emph{everything is all it takes} to best constrain the optimization and achieve our best empirical results, suggesting that a silver bullet strategy does not currently exist.

Specifically, we evaluate: cross-lingual data projection and self-training techniques with different machine translation and word alignment components, the impact of bilingual and multilingual contextualized encoders on each data projection and self-training component, and the use of different encoders in task-specific models. We also offer suggestions for practitioners operating under different computation budgets on four tasks: event extraction, named entity recognition, part-of-speech tagging, and dependency parsing, following recent work that uses English-to-Arabic tasks as a test bed \cite{lan-etal-2020-empirical}.
We then apply data projection and self-training to three structured prediction tasks---named entity recognition, part-of-speech tagging, and dependency parsing---in multiple target languages. Additionally, we use self-training as a control against data projection to determine in which situations data projection improves performance.

This chapter is adapted from \citet{yarmohammadi-etal-2021-everything}, which
is a distillation of the Phase 1 evaluation effort of the BETTER team at Johns
Hopkins University and University of Rochester, supported by IARPA BETTER (\#2019-19051600005). My colleagues on the BETTER team are responsible for the codebase for our BETTER system, and their efforts form the core of this paper. My co-first-author Mahsa Yarmohammadi and I designed and ran most of the experiments reported in this paper and drafted the paper. Marc Marone and Seth Ebner assisted in writing and editing. Other contributions made by my colleagues will be indicated throughout the chapter.

\begin{figure*}[t]
\centering
\includegraphics[width=0.9\textwidth]{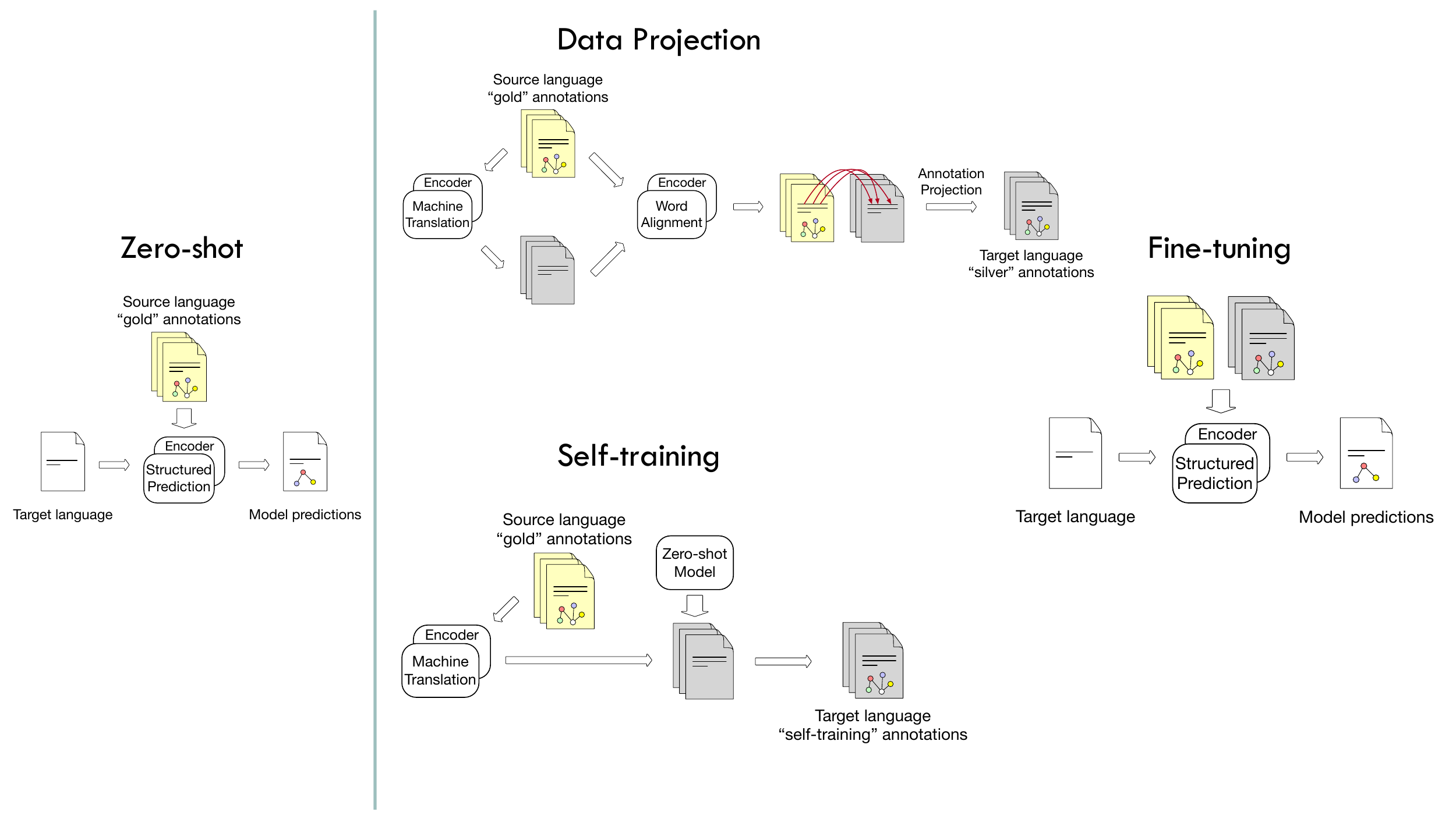}
\caption{Process for creating projected ``silver'' data from source ``gold'' data. Downstream models are trained on a combination of gold and silver data. Components in boxes have learned parameters. This figure is made by Seth Ebner and Mahsa Yarmohammadi.}
\label{fig:system}
\end{figure*}

\section{Universal Encoders}

\input{table/projection/encoder}

While massively multilingual encoders like mBERT and XLM-R enable strong zero-shot cross-lingual performance \cite{wu-dredze-2019-beto,conneau-etal-2020-unsupervised}, they suffer from the curse of multilinguality \cite{conneau-etal-2020-unsupervised}: cross-lingual effectiveness suffers as the number of supported languages increases for a fixed model size.
We would therefore expect that when restricted to only the source and target languages, a bilingual model should perform better than (or at least on par with) a multilingual model of the same size, assuming both languages have sufficient corpora \cite{wu-dredze-2020-languages}.
If a practitioner is interested in only a small subset of the supported languages, \textit{is the multilingual model still the best option}?

To answer this question, we use English and Arabic as a test bed. In \autoref{tab:encoder}, we summarize existing publicly available encoders that support both English and Arabic.\footnote{We do not include multilingual T5 \cite{xue-etal-2021-mt5} as it is still an open question on how to best utilize text-to-text models for structured prediction tasks \cite{ruder-etal-2021-xtreme}.}
Base models are 12-layer Transformers ($\texttt{d\_model} = 768$), and large models are 24-layer Transformers ($\texttt{d\_model} = 1024$) \cite{vaswani2017attention}.
As there is no publicly available large English--Arabic bilingual encoder, we train two encoders from scratch, named L64K and L128K, with vocabulary sizes of 64K and 128K, respectively.\footnote{L128K available at \url{https://huggingface.co/jhu-clsp/roberta-large-eng-ara-128k}} With these encoders, we can determine the impacts of model size and the number of supported languages.

\section{Data Projection and Self-Training}

We create silver versions of the data by automatically projecting annotations from source English gold data to their corresponding machine translations in the target language or labeling the translations with the zero-shot model.
Data projection transfers word-level annotations in a source language to a target language via word-to-word alignments \cite{yarowsky-etal-2001-inducing}. The technique has been used to create cross-lingual datasets for a variety of structured natural language processing tasks, including named entity recognition \cite{stengel-eskin-etal-2019-discriminative} and semantic role labeling \cite{akbik-etal-2015-generating,aminian-etal-2017-transferring,fei-etal-2020-cross}. Labeling data with learned models and using it to further train the model is referred to as self-training \cite{yarowsky-1995-unsupervised}. This technique has been used for cross-lingual transfer including text classification \cite{eisenschlos-etal-2019-multifit}. However, we differ with prior work as we label the translation of source gold data instead of assuming access to unlabeled corpus in target language.

For data projection, to create silver data, as shown in \autoref{fig:system}, we: (1) translate the source text to the target language using the MT system described in Section \ref{sec:mt}, (2) obtain word alignments between the original and translated parallel text using a word alignment tool, and (3) project the annotations along the word alignments. We then combine silver target data with gold source data to augment the training set for the structured prediction task. For self-training, the step (1) is shared with data projection, and we use the zero-shot model to label the translation.

For step (1), we rely on a variety of source-to-target MT systems.
To potentially leverage monolingual data, as well as contextualized cross-lingual information from pretrained encoders, we feed the outputs of the final layer of frozen pretrained encoders as the inputs to the MT encoders. We consider machine translation systems: (i) whose parameters are randomly initialized, (ii) that incorporate information from massively multilingual encoders, and (iii) that incorporate information from bilingual encoders that have been trained on only the source and target languages.

After translating source sentences to the target language, in step (2) we obtain a mapping of the source words to the target words using publicly available automatic word alignment tools. Similarly to our MT systems, we incorporate contextual encoders in the word aligner. We hypothesize that better word alignment yields better silver data, and better information extraction consequently.

For step (3), we apply direct projection to transfer labels from source sentences to target sentences according to the word alignments. Each target token receives the label of the source token aligned to it (token-based projection).  For multi-token spans, the target span is a contiguous span containing all aligned tokens from the same source span (span-based projection), potentially including tokens not aligned to the source span in the middle. Three of the IE tasks we consider---ACE, named entity recognition, and BETTER---use span-based projection, and we filter out projected target spans that are five times longer than the source spans. Two syntactic tasks---POS tagging and dependency parsing---use token-based projection. For dependency parsing, following \citet{tiedemann-etal-2014-treebank}, we adapt the disambiguation of many-to-one mappings by choosing as the head the node that is highest up in the dependency tree. In the case of a non-aligned dependency head, we choose the closest aligned ancestor as the head.

To address issues like translation shift, filtered projection \cite{akbik-etal-2015-generating,aminian-etal-2017-transferring} has been proposed to obtain higher precision but lower recall projected data. To maintain the same amount of silver data as gold data, in this chapter we do not use any task-specific filtered projection methods to remove any sentence.

\section{Tasks}
We employ our silver dataset creation approach on a variety of tasks.
For English--Arabic experiments, we consider ACE, BETTER, NER, POS tagging, and dependency parsing. For multilingual experiments, we consider NER, POS tagging, and dependency parsing. We use English as the source language and 8 typologically diverse target languages: Arabic, German, Spanish, French, Hindi, Russian, Vietnamese, and Chinese. Because of the high variance of cross-lingual transfer as shown in \autoref{chap:crosslingual-signal}, we report the average test performance of three runs with different predefined random seeds (except for ACE).\footnote{We report one run for ACE due to long fine-tuning time.} For model selection and development, we use the English dev set in the zero-shot scenario and the combined English dev and silver dev sets in the silver data scenario. Mahsa Yarmohammadi ran the experiment on English--Arabic ACE and BETTER with the help of Shabnam Behzad.

\subsection{ACE}
Automatic Content Extraction (ACE) 2005 \cite{ace2005} provides named entity, relation, and event annotations for English, Chinese, and Arabic. We conduct experiments on English as the source language and Arabic as the target language. We use the OneIE framework \cite{lin-etal-2020-joint}, a joint neural model for information extraction, which has shown state-of-the-art results on all subtasks. We use the same hyperparameters as in \citet{lin-etal-2020-joint} for all of our experiments.
We use the OneIE scoring tool to evaluate the prediction of entities, relations, event triggers, event arguments, and argument roles. For English, we use the same English document splits as \cite{lin-etal-2020-joint}. That work does not consider Arabic, so for Arabic we use the document splits from \cite{lan-etal-2020-empirical}.

We used the OneIE v0.4.8 codebase\footnote{http://blender.cs.illinois.edu/software/oneie/} with the following hyperparameters: Adam optimizer \citep{kingma2014adam} for 60 epochs with a learning rate of 5e-5 and weight decay of 1e-5 for the encoder, and a learning rate of 1e-3 and weight decay of 1e-3 for other parameters. Two-layer feed-forward network with a dropout rate of 0.4 for task-specific classifiers, 150 hidden units for entity and relation extraction, and 600 hidden units for event extraction. $\beta_v$ and $\beta_e$ set to 2 and $\theta$ set to 10 for global features.

\subsection{Named Entity Recognition}
We use WikiAnn \citep{pan-etal-2017-cross} for English--Arabic and multilingual experiments. The labeling scheme is BIO with 3 types of named entities: PER, LOC, and ORG. On top of the encoder, we use a linear classification layer with softmax to obtain word-level predictions. The labeling is word-level while the encoders operate at subword-level, thus, we mask the prediction of all subwords except for the first one. We evaluate NER performance by F1 score of the predicted entity.

We use the Adam optimizer with a learning rate of 2e-5 with linear warmup for the first 10\% of total steps and linear decay afterwards, and train for 5 epochs with a batch size of 32. We adapt the same post-processing step as \autoref{sec:task-ner} to obtain valid BIO sequences. We set the maximum sequence length to 128 during fine-tuning, and use a sliding window of context to include subwords beyond the first 128. At test time, we use the same maximum sequence length.

\subsection{Part-of-speech Tagging}
We use the Universal Dependencies (UD) Treebank \citep[v2.7;][]{ud2.7}. We use the following treebanks: Arabic-PADT, German-GSD, English-EWT, Spanish-GSD, French-GSD, Hindi-HDTB, Russian-GSD, Vietnamese-VTB, and Chinese-GSD.
Similar to NER, we use a word-level linear classifier on top of the encoder, and evaluate performance by the accuracy of predicted POS tags. We use the same fine-tuning hyperparameter and maximum sequence length as NER.

\subsection{Dependency Parsing}
We use the same treebanks as the POS tagging task. For the task-specific layer, we use the graph-based parser of \citet{DBLP:journals/corr/DozatM16}, but replace their LSTM encoder with our encoders of interest. We follow the same policy as that in NER for masking non-first subwords. We predict only the universal dependency labels, and we evaluate performance by labeled attachment score (LAS), ignoring punctuations (PUNCT) and symbols (SYM). We use the same fine-tuning hyperparameter as NER. We set the maximum sequence length to the first 128 subwords during fine-tuning, and the first 128 words at test time.

\subsection{BETTER}
 The Better Extraction from Text Towards Enhanced Retrieval (BETTER) Program\footnote{\url{https://www.iarpa.gov/index.php/research-programs/better}} develops methods for extracting increasingly fine-grained semantic information in a target language, given gold annotations only in English. 
We focus on the coarsest ``Abstract'' level, where the goal is to identify events and their agents and patients. %
The documents come from the news-specific portion of Common Crawl. We report the program-defined ``combined F1'' metric, which is the product of ``event match F1'' and ``argument match F1'', which are based on an alignment of predicted and reference event structures.%

To find all events in a sentence and their corresponding arguments,
we model the structure of the events as a tree, where event triggers are children of the ``virtual root'' of the sentence and arguments are children of event triggers
\citep{cai-etal-2018-full}.
Each node is associated with a span in the text and is labeled with an event or argument type label.

We use a model for event structure prediction that has three major components: a contextualized encoder, tagger, and typer~\cite{xia-etal-2021-lome}.\footnote{Code available at \url{https://github.com/hiaoxui/span-finder}}
This model is designed by my BETTER team colleague.
The tagger is a BiLSTM-CRF BIO tagger \citep{panchendrarajan-amaresan-2018-bidirectional} trained to predict child spans conditioned on parent spans and labels.
The typer is a feedforward network whose inputs are a parent span representation, parent label embedding, and child span representation.
The tree is produced level-wise at inference time, first predicting event triggers, typing them, and then predicting arguments conditioned on the typed triggers.

The codebase for event structure prediction uses AllenNLP \citep{gardner-etal-2018-allennlp}.
The contextual encoder produces representations for the tagger and typer modules. 
Span representations are formed by concatenating the output of a self-attention layer over the span's token embeddings with the embeddings of the first and last tokens of the span. 
The BiLSTM-CRF tagger has 2 layers, both with hidden size of 2048.
We use a dropout rate of 0.3 and maximum sequence length of 512.
Child span prediction is conditioned on parent spans and labels, so we represent parent labels with an embedding of size 128. 
We use Adam optimizer to fine-tune the encoder with a learning rate of 2e-5, and we use a learning rate of 1e-3 for other components. 
The tagger loss is negative log likelihood and the typer loss is cross entropy.
We equally weight both losses and train against their sum. The contextual encoder is not frozen.

\section{Experiments}

\subsection{Universal Encoders}

We train two English--Arabic bilingual encoders. Both of them are 24-layer Transformers ($\texttt{d\_model} = 1024$), the same size as XLM-R large. We use the same Common Crawl corpus as XLM-R for pretraining. Additionally, we also use English and Arabic Wikipedia, Arabic Gigaword \cite{parker2011arabic}, Arabic OSCAR \cite{ortiz-suarez-etal-2020-monolingual}, Arabic News Corpus \cite{el20161}, and Arabic OSIAN \cite{zeroual-etal-2019-osian}. In total, we train with 9.2B words of Arabic text and 26.8B words of English text, more than either XLM-R (2.9B words/23.6B words) or GBv4 (4.3B words/6.1B words).\footnote{We measure word count with \texttt{wc -w}.} We build two English--Arabic joint vocabularies using SentencePiece \cite{kudo-richardson-2018-sentencepiece}, resulting in two encoders: \textbf{L64K} and \textbf{L128K}. For the latter, we additionally enforce coverage of all Arabic characters after normalization.

We pretrain each encoder with a batch size of 2048 sequences and 512 sequence length for 250K steps from scratch,\footnote{While we use XLM-R as the initialization of the Transformer, due to vocabulary differences, the learning curve is similar to that of pretraining from scratch.} roughly 1/24 the amount of pretraining compute of XLM-R. Training takes 8 RTX 6000 GPUs roughly three weeks. We follow the pretraining recipe of RoBERTa \cite{liu2019roberta} and XLM-R. We omit the next sentence prediction task and use a learning rate of 2e-4, Adam optimizer, and linear warmup of 10K steps then decay linearly to 0, multilingual sampling alpha of 0.3, and the fairseq \cite{ott-etal-2019-fairseq} implementation.

\subsection{Machine Translation}
\label{sec:mt}
The machined translation component is developed by Haoran Xu and Kenton Murray, initially developed for the BETTER program and later improved by \citet{xu-etal-2021-bert}. The detailed description of this component can be found in \citet{yarmohammadi-etal-2021-everything}. In summary, it uses the final contextual embeddings from a frozen bilingual or multilingual encoder as the input of the MT encoder, instead of a randomly initialized embedding matrix (``None''). We also include a publicly released model (``public'') that has been demonstrated to perform well \citep{tiedemann-2020-tatoeba}.\footnote{The public MT model is available at \url{https://huggingface.co/Helsinki-NLP/opus-mt-en-ar}} \autoref{tab:mt} shows the denormalized and detokenized BLEU scores for English--Arabic MT systems with 
different encoders on the IWLST'17 test set using sacreBLEU \citep{post-2018-call}. 
The use of contextualized embeddings from pretrained encoders results in better performance than using a standard randomly initialized MT model regardless of which encoder is used. The best performing system uses our bilingual L64K encoder, but all pretrained encoder-based systems perform well and within 0.5 BLEU points of each other. 

\input{table/projection/mt}

\subsection{Word Alignment}
Until recently, alignments have typically been obtained using unsupervised statistical models such as  GIZA++ \cite{och-ney-2003-systematic} and fast-align \cite{dyer-etal-2013-simple}. Recent work has focused on using the similarities between contextualized embeddings to obtain alignments \cite{jalili-sabet-etal-2020-simalign,daza-frank-2020-x,dou-neubig-2021-word}, achieving state-of-the-art performance.

\input{table/projection/gale}

We use two automatic word alignment tools: fast-align, a widely used statistical alignment tool based on IBM models \cite{brown-etal-1993-mathematics}; and \awesome~\cite{dou-neubig-2021-word}, a contextualized embedding-based word aligner that extracts word alignments based on similarities of the tokens' contextualized embeddings. \awesome~achieves state-of-the-art performance on five language pairs. Optionally, \awesome~can be fine-tuned on
parallel text with objectives suitable for word alignment and on gold alignment data.
Mahsa Yarmohammadi ran the experiment on word alignment with the help of Shabnam Behzad, while I adapted the code to allow using any encoder.

We benchmark the word aligners on the gold standard alignments in the GALE Arabic--English Parallel Aligned Treebank \cite{li-etal-2012-parallel}. We use the same data splits as \citet{stengel-eskin-etal-2019-discriminative}, containing 1687, 299, and 315 sentence pairs in the train, dev, and test splits, respectively. %
To obtain alignments using fast-align, we append the test data to the MT training bitext and run the tool from scratch.
\awesome~extracts the alignments for the test set based on pretrained contextualized embeddings. These encoders can be fine-tuned using the parallel text in the train and dev sets. Additionally, the encoders can be further fine-tuned using supervision from gold word alignments.

\subsubsection{Intrinsic Evaluation} \autoref{tab:alignment-results} shows the performance of word alignment methods on the GALE English--Arabic alignment dataset. \awesome~outperforms fast-align, and fine-tuned \awesome~($\textit{ft}$) outperforms models that were not fine-tuned. Incorporating supervision from the gold alignments ($\textit{s}$) leads to the best performance. %

\section{Cross-lingual Transfer}

One might optimistically consider that the latest multilingual encoder (in this case XLM-R) in the zero-shot setting would achieve the best possible performance, which suggest data projection or self-training could not constrain the zero-shot cross-lingual optimization. However, in our extensive experiments in \autoref{tab:main} and \autoref{tab:multi}, we find that data projection and self-training could provide useful constraints and improve over zero-shot approach. In this section, we explore the impact of each factor within the silver data creation process.

\subsection{English--Arabic Experiments}

\input{table/projection/main-table-2x2}

In \autoref{tab:main}, we present the Arabic test performance of five tasks under all  combinations considered. The ``MT'' and ``Align'' columns indicate the models used for the translation and word alignment components of the silver data creation process. For ACE, we report results on the average of six metrics.
\footnote{Six metrics include entity, relation, trigger identification and classification, and argument identification and classification accuracies.
} For a large bilingual encoder, we use L128K instead of L64K due to its slightly better performance on English ACE. %

\subsubsection{Impact of Data Projection} By comparing any group against group Z, we observe adding silver data yields better or equal performance to zero-shot in at least some setup in the IE tasks (ACE, NER, and BETTER). For syntax-related tasks, we observe similar trends, with the exception of XLM-R. We hypothesize that XLM-R provides better syntactic cues than those obtainable from the alignment, which we discuss later in relation to self-training.

\subsubsection{Impact of Word Aligner} By comparing groups A, B, and C of the same encoder, we observe that \awesome~performs overall better than statistical MT-based fast-align (FA). Additional fine-tuning ($\textit{ft}$) on MT training bitext further improves its performance. As a result, we use fine-tuned aligners for further experiments. Moreover, incorporating supervised signals from gold alignments in the word alignment component ($\textit{ft.s}$) often helps performance of the task. In terms of computation budget, these three groups use a publicly available MT system \citep[``public'';][]{tiedemann-2020-tatoeba} and require only fine-tuning the encoder for alignment, which requires small additional computation.%

\subsubsection{Impact of Encoder Size} Large bilingual or multilingual encoders tend to perform better than base encoders in the zero-shot scenario, with the exception of the bilingual encoders on ACE and BETTER.
While we observe base size encoders benefit from reducing the number of supported languages (from 100 to 2), for large size encoders trained much longer, the zero-shot performance of the bilingual model is worse than that of the multilingual model.
After adding silver data from group C based on the public MT model and the fine-tuned aligner, the performance gap between base and large models tends to shrink, with the exception of both bilingual and multilingual encoders on NER. %
In terms of computation budget, training a bilingual encoder requires significant additional computation.

\subsubsection{Impact of Encoder on Word Aligner} By comparing groups C and D (in multilingual encoders) or groups E and F (in bilingual encoders), we observe bilingual encoders tend to perform slightly worse than multilingual encoders for word alignment. If bilingual encoders exist, using them in aligners requires little additional computation.

\subsubsection{Impact of Encoder on MT} By comparing groups C and E, we observe the performance difference between the bilingual encoder based MT and the public MT depends on the task and encoder, and neither MT system clearly outperforms the other in all settings, despite the bilingual encoder having a better BLEU score.
The results suggest that both options should be explored if one's budget allows.
In terms of computation budget, using pretrained encoders in a custom MT system requires medium additional computation.%

\subsubsection{Impact of Label Source} 
To assess the quality of the projected annotations in the silver data, we consider a different way to automatically label translated sentences: self-training \citep[ST;][]{yarowsky-1995-unsupervised}. For self-training, we translate the source data to the target language, label the translated data using a zero-shot model trained on source data, and combine the labeled translations with the source data to train a new model.\footnote{This setup differs from traditional zero-shot self-training in cross-lingual transfer, as the traditional setup assumes unlabeled corpora in the target language(s) \cite{eisenschlos-etal-2019-multifit} instead of translations of the source language data.}
Compared to the silver data, the self-training data has the same underlying text but a different label source. 

We first observe that self-training for parsing leads to significantly worse performance due to the low quality of the predicted trees. 
By comparing groups S and C, which use the same underlying text, we observe that data projection tends to perform better than self-training, with the exceptions of POS tagging with a large encoder and NER with a large multilingual encoder. 
These results suggest that the external knowledge\footnote{``External knowledge'' refers to knowledge introduced into the downstream model as a consequence of the particular decisions made by the aligner (and subsequent projection).}
in the silver data complements the knowledge obtainable when the model is trained with source language data alone, but when the zero-shot model is already quite good (like for POS tagging) data projection can harm performance compared to self-training.

\subsection{Multilingual Experiments}

\input{table/projection/multilingual}

In \autoref{tab:multi}, we present the test performance of three tasks for eight target languages. We use the public MT system \cite{tiedemann-2020-tatoeba} and non-fine-tuned \awesome~with mBERT as the word aligner for data projection---a setup with the smallest computation budget---due to computation constraints. We consider both data projection (+Proj) and self-training (+Self). We use silver data in addition to English gold data for training. We use multilingual training with +Self and +Proj, and bilingual training with +Proj (Bi).

We observe that data projection (+Proj (Bi)) sometimes benefits languages with the lowest zero-shot performance (Arabic, Hindi, and Chinese), with the notable exception of XLM-R on syntax-based tasks (excluding Chinese). For languages closely related to English, data projection tends to hurt performance. %
We observe that for data projection, training multiple bilingual models (+Proj (Bi)) outperforms joint multilingual training (+Proj). This could be the result of noise from alignments of various quality mutually interfering. In fact, self-training with the same translated text (+Self) outperforms data projection and zero-shot scenarios, again with the exception of parsing. As data projection and self-training use the same translated text and differ only by label source, the results indicate that the external knowledge from frozen mBERT-based alignment is worse than what the model learns from source language data alone. Thus, further performance improvement could be achieved with an improved aligner.

\section{Related Work}

Although projected data may be of lower quality than the original source data due to errors in translation or alignment, it is useful for tasks such as semantic role labeling \cite{akbik-etal-2015-generating,aminian-etal-2019-cross}, information extraction \cite{riloff-etal-2002-inducing}, POS tagging \cite{yarowsky-ngai-2001-inducing}, and dependency parsing \cite{ozaki-etal-2021-project}. The intuition is that although the projected data may be noisy, training on it gives a model useful information about the statistics of the target language.

\citet{akbik-etal-2015-generating} and \citet{aminian-etal-2017-transferring} use bootstrapping algorithms to iteratively construct projected datasets for semantic role labeling. \citet{akbik-etal-2015-generating} additionally use manually defined filters to maintain high data quality, which results in a projected dataset that has low recall with respect to the source corpus.
\citet{fei-etal-2020-cross} and \citet{daza-frank-2020-x} find that a non-bootstrapped approach works well for cross-lingual SRL. Advances in translation and alignment quality allow us to avoid bootstrapping while still constructing projected data that is useful for downstream tasks.

\citet{fei-etal-2020-cross} and \citet{daza-frank-2020-x} also find improvements when training on a mixture of gold source language data and projected silver target language data.
The intuition of using both gold and projected silver data is to allow the model to see high quality gold data as well as data with target language statistics.
Ideas from domain adaptation can be used to make more effective use of gold and silver data to mitigate the effects of language shift \cite{xu-etal-2021-gradual}.

\section{Discussion}

In this chapter, we explore the use of silver data via data projection or self-training to constrain the zero-shot cross-lingual transfer optimization, facilitated by neural machine translation and word alignment. Recent advances in pretrained encoders have improved machine translation systems and word aligners in terms of intrinsic evaluation. We conduct an extensive extrinsic evaluation and study how the encoders themselves---and components containing them---impact performance on a range of downstream tasks and languages.

With a test bed of English--Arabic IE tasks, we find that adding projected silver training data overall yields improvements over zero-shot learning. Comparisons of how each factor in the data projection process impacts performance show that while one might hope for the existence of a silver bullet strategy, the best setup is usually task dependent.
In multilingual experiments, we find that silver data tends to help languages with the weakest zero-shot performance, and that it is best used separately for each desired language pair instead of in joint multilingual training.

We also examine self-training with translated text to assess when data projection helps cross-lingual transfer, and find it to be another viable option for obtaining labels for some tasks.
Further directions include how to improve alignment quality and how to combine data projection and self-training techniques. 

As we observe in this chapter, the best setup to constrain the optimization for each task is task-specific. Thus, to further constrain the optimization and to produce the best performance with the existing encoders, we should consider a bag of techniques mentioned in this chapter and \autoref{sec:analysis-discussion}, depending on the computation budget.

%% file: table/projection/encoder.tex
\begin{table}
\begin{center}
\begin{tabular}{l|ll}
\toprule
& \textbf{Base} & \textbf{Large}\\
\midrule
\textbf{Multilingual} & mBERT & XLM-R \\
 & (\citeauthor{devlin-etal-2019-bert}) & (\citeauthor{conneau-etal-2020-unsupervised}) \\
\textbf{Bilingual} & GBv4 & L64K \& L128K \\
 & (\citeauthor{lan-etal-2020-empirical}) &  \textbf{(Ours)} \\
\bottomrule
\end{tabular}
\caption{Encoders supporting English and Arabic.}
\label{tab:encoder}
\end{center}
\end{table}

%% file: table/projection/mt.tex
\begin{table}[]
\begin{center}
\begin{tabular}{cc}
\toprule
\textbf{Encoder} & \textbf{BLEU}  \\
\midrule
Public & 12.7 \\
\midrule
None & 14.9 \\
\midrule
mBERT & 15.7 \\
GBv4 & 15.7 \\
\midrule
XLM-R & 16.0 \\
L64K & \bf 16.2 \\
L128K & 15.8\\
\bottomrule
\end{tabular}
\caption{BLEU scores of MT systems with different pre-trained encoders on English--Arabic IWSLT'17.}
\label{tab:mt}
\end{center}
\end{table}

%% file: table/projection/gale.tex
\begin{table}
\begin{center}
\begin{tabular}{lccccc}

\toprule
\textbf{Model} & \textbf{Layer}\dag & \textbf{AER} & \textbf{P} & \textbf{R}  & \textbf{F}\\
\midrule
fast-align*& n/a & 47.4&53.9&51.4&52.6\\
\midrule
\multicolumn{6}{l}{\textit{Awesome-align w/o FT}} \\
\midrule
mBERT           & 8  & 35.6 & 78.5 & 54.5 & 64.4 \\
GBv4     & 8  & \textbf{32.7} & \textbf{85.6} & 55.4 & \textbf{67.3} \\
\midrule
XLM-R & 16 & 40.1 & 78.6 & 48.4 & 59.9 \\
L64K  & 17 & 34.0 & 81.5 & \textbf{55.5} & 66.0 \\
L128K & 17 & 35.1 & 80.0 & 54.5 & 64.9 \\
\midrule
\multicolumn{6}{l}{\textit{Awesome-align w/ FT}} \\
\midrule
mBERT$_\textit{ft}$   & 8  & 30.0 & 81.9 & \textbf{61.2} & 70.0 \\
GBv4$_\textit{ft}$       & 8  & 29.3 & 86.9 & 59.7 & 70.7 \\
\midrule
XLM-R$_\textit{ft}$   & 18 & \textbf{27.8} & \textbf{90.3} & 60.2 & \textbf{72.2} \\
L64K$_\textit{ft}$    & 17 & 29.1 & 84.9 & 60.9 & 70.9 \\
L128K$_\textit{ft}$   & 16 & 32.2 & 80.3 & 58.7 & 67.8 \\
\midrule
\multicolumn{6}{l}{\textit{Awesome-align w/ FT \& supervision}} \\
\midrule
XLM-R$_\textit{ft.s}$   & 16 & \textbf{23.3} & 92.5 & \textbf{65.6} & \textbf{76.7} \\
L128K$_\textit{ft.s}$   & 17 & 23.5 & \textbf{93.7}& 64.6 & 76.5\\
\bottomrule

\end{tabular}
\caption{Alignment performance on GALE EN--AR. *Trained on MT bitext. \dag We report the best layer of each encoder based on dev alignment error rate (AER).}
\label{tab:alignment-results}
\end{center}
\end{table}

%% file: table/projection/main-table-2x2.tex
\begin{table*}[]
\small
\begin{center}
\resizebox{\linewidth}{!}{
\begin{tabular}{lll| ccccc|| ccccc}

\toprule
 & \textbf{MT} & \textbf{Align} & \textbf{ACE} & \textbf{NER} & \textbf{POS} & \textbf{Parsing} & \textbf{BET.} & \textbf{ACE} & \textbf{NER} & \textbf{POS} & \textbf{Parsing} & \textbf{BET.}\\
\midrule
& & & \multicolumn{5}{l||}{\textit{mBERT (base, multilingual)}} & \multicolumn{5}{l}{\textit{XLM-R (large, multilingual)}}   \\
\midrule
(Z) & - & - & 27.0 & 41.6 & 59.7 & 29.2 & 39.9  & \textbf{45.1} & 46.4 & 73.3 & \best \textbf{48.0} & 50.8\\
\midrule
(A) & public & FA & \better +2.5 & \worse -3.8 & \muchbetter +8.5 & \muchbetter +7.3 & \better +2.6 & \muchworse -7.5 & -0.1 & \muchworse -7.7 & \muchworse -9.5 & \worse -1.6\\
\midrule
(B) & public & mBERT & \muchbetter +6.5 & +0.2 & \muchbetter +8.5 & \muchbetter +7.6 & \better +2.3 & \worse -4.4 & \muchbetter +6.9 & \muchworse -6.1 & \muchworse -8.4 & \worse -2.6 \\
(B) & public & XLM-R & +0.9 & \worse -2.9 & \muchbetter +9.5 & \muchbetter +9.0 & \worse -1.2 & \muchworse -10.0 & +0.0 & \muchworse -5.9 & \muchworse -8.8 & \muchworse -6.3\\
\midrule
(C) & public & mBERT$_\textit{ft}$ & \muchbetter +7.8 & \muchbetter \textbf{+5.6} & \muchbetter +7.7 & \muchbetter +10.0 & \better +4.1 & -0.6 & \muchbetter +7.4 & \muchworse -8.0 & \muchworse -6.8 & +0.3\\
(C) & public & XLM-R$_\textit{ft}$ & \muchbetter +7.7 & \better +4.9 & \muchbetter +6.2 & \muchbetter +9.3 & \better +4.5 & \worse -2.6 & \muchbetter +7.0 & \muchworse -9.0 & \muchworse -7.6 & \better +1.0 \\
(C) & public & XLM-R$_\textit{ft.s}$ & \muchbetter +7.3 & \better +1.5 & \muchbetter +10.1 & \muchbetter \textbf{+12.4} & \better +4.8 & \worse -3.0 & \muchbetter +9.1 & \worse -3.8 & \worse -3.7 & \best \textbf{+2.3}\\
\midrule
(D) & public & GBv4$_\textit{ft}$ & \muchbetter +8.5 & \better +4.3 & \muchbetter +5.9 & \muchbetter +8.9 & \muchbetter +5.0 & \worse -1.5 & \muchbetter +7.7 & \muchworse -9.4 & \muchworse -9.1 & -0.1\\
(D) & public & L128K$_\textit{ft}$ & \muchbetter +6.4 & \better +3.1 & \muchbetter +6.5 & \muchbetter +8.2 & \better +1.6 & \worse -1.6 & \muchbetter +6.1 & \muchworse -9.0 & \muchworse -9.4 & \worse -3.6\\
(D) & public & L128K$_\textit{ft.s}$ & \muchbetter +7.0 & \better +3.7 & \muchbetter \textbf{+10.3} & \muchbetter +11.8 & \muchbetter \textbf{+5.4} & -0.3 & \muchbetter +5.2 & \worse -4.4 & \worse -4.6 & \better +2.1 \\
\midrule
(E) & GBv4 & mBERT$_\textit{ft}$ & \muchbetter +8.4 & \better +3.2 & \muchbetter +7.7 & \muchbetter +9.9 & \better +4.7 & \worse -1.5 & \better +3.2 & \muchworse -7.1 & \muchworse -6.7 & +0.7\\
(E) & GBv4 & XLM-R$_\textit{ft}$ & \muchbetter +9.6 & \better +1.8 & \muchbetter +7.0 & \muchbetter +9.5 & \muchbetter +5.2 & -0.4 & \better +1.4 & \muchworse -8.3 & \muchworse -7.7 & \better +1.4\\
(E) & L128K & mBERT$_\textit{ft}$ & \muchbetter \textbf{+12.1} & \better +3.3 & \muchbetter +7.9 & \muchbetter +9.9 & \better +4.7 & \worse -1.4 & \muchbetter +7.2 & \muchworse -8.1 & \muchworse -6.7 & \better +1.3\\
(E) & L128K & XLM-R$_\textit{ft}$ & \muchbetter +10.2 & \worse -1.9 & \muchbetter +6.1 & \muchbetter +9.4 & \better +4.8 & -0.5 & \better +4.6 & \muchworse -9.8 & \muchworse -7.5 & \better +2.0\\
\midrule
(S) & public & ST & - & \muchbetter +5.5 & +0.1 & \muchworse -20.3 & +0.3 & - & \best \textbf{+10.0} & \best \textbf{+1.8} & \muchworse -29.6 & \better +1.2 \\
\midrule
\midrule
& & &\multicolumn{5}{l||}{\textit{GBv4 (base, bilingual)}} & \multicolumn{5}{l}{\textit{L128K (large, bilingual)}}  \\
\midrule
(Z) & - & - & 46.0 & 45.4 & 64.7 & 33.2 & 41.7 & 42.7 & 46.3 & 67.9 & 36.7 & 40.9\\
\midrule
(C) & public & mBERT$_\textit{ft}$ & +0.6 & \better +3.7 & \better +2.6 & \muchbetter +6.9 & \muchbetter +7.5 & \better +2.7 & \muchbetter +8.2 & -0.9 & \better +4.9 & \muchbetter +11.7\\
(C) & public & XLM-R$_\textit{ft}$ & \worse -1.4 & \better \textbf{+4.5} & \better +1.8 & \muchbetter +6.0  & \muchbetter +8.4 & \better +1.2 & \muchbetter \textbf{+9.0} & \worse -2.5 & \better +3.9 & \muchbetter +10.5 \\
(C) & public & XLM-R$_\textit{ft.s}$ & -0.1 & \better +3.4 & \muchbetter \textbf{+5.1} & \muchbetter \textbf{+9.2} & \muchbetter +8.0 & \better +2.7 & \muchbetter +7.0 & \better +1.2 & \muchbetter \textbf{+7.2} & \muchbetter \textbf{+12.1}\\
\midrule
(E) & GBv4 & mBERT$_\textit{ft}$ & -0.1 & +0.1 & \better +3.3 & \muchbetter +7.2 & \muchbetter +8.1 & \better +4.2 & -0.5 & -0.1 & \muchbetter +5.1  & \muchbetter +11.2\\
(E) & GBv4 & XLM-R$_\textit{ft}$ & +0.1 & +0.4 & \better +1.5 & \muchbetter +6.0  & \muchbetter \textbf{+9.7} & \better +2.4 & +0.0 & \worse -1.3 & \better +4.2  & \muchbetter +10.8\\
(E) & L128K & mBERT$_\textit{ft}$ & -0.6 & \better +1.0 & \better +2.6 & \muchbetter +6.1  &  \muchbetter +7.4& \best \textbf{+5.5} & +0.8 & -0.7 & \better +4.7  & \muchbetter +10.6 \\
(E) & L128K & XLM-R$_\textit{ft}$ & +0.9 & \worse -2.1 & \better +1.1 & \muchbetter +5.5  & \muchbetter +7.8 & \better +4.4 & \worse -3.6 & \worse -2.2 & \better +4.1  & \muchbetter +11.3\\
\midrule
(F) & GBv4 & GBv4$_\textit{ft}$ & +0.0 & \worse -1.9 & \better +1.6 & \better +4.5 & \muchbetter +9.1  & \better +2.0 & -0.3 & \worse -1.7 & \better +3.2  & \muchbetter +10.9\\
(F) & GBv4 & L128K$_\textit{ft}$ & -0.9 & \worse -1.4 & \better +1.5 & \better +4.1 & \muchbetter +5.7  & \better +2.3 & \worse -1.7 & \worse -2.4 & \better +2.6 & \muchbetter +8.3\\
(F) & L128K & GBv4$_\textit{ft}$ & \worse -4.3 & \worse -1.0 & +0.4 & \better +4.1 & \muchbetter +7.4 & \better +4.1 & \worse -3.6 & \worse -2.1 & \better +2.3 & \muchbetter +11.4\\
(F) & L128K & L128K$_\textit{ft}$ & \worse -3.5 & \worse -1.1 & +0.3 & \better +3.8  & \better + 4.5 & \better +2.9 & +0.1 & \worse -2.9 & \better +2.0 & \muchbetter +6.7\\
(F) & L128K & L128K$_\textit{ft.s}$ & \better \textbf{+1.9} & +0.2 & \better +3.3 & \muchbetter +7.4 & \muchbetter +7.2 & \better +2.8 & \worse -1.8 & +0.8 & \muchbetter \muchbetter +6.0 & \muchbetter +11.8 \\
\midrule
(S) & public & ST & - & \worse -2.5 & \worse -1.3 & \muchworse -18.6 & \better +1.9 & - & \muchbetter +7.1 & \better \textbf{+1.5} & \muchworse -21.7 & \muchbetter +8.1\\
\bottomrule
\end{tabular}
}
\caption{Performance of Arabic on 5 tasks under various setups. %
Cells are colored by performance difference over zero-shot baseline: \colorbox{muchbetter}{+5 or more}, \colorbox{better}{+1 to +5}, \colorbox{worse}{-1 to -5}, \colorbox{muchworse}{-5 or more}. \colorbox{best}{\textbf{Highlights}} indicate the best setting for each task (best viewed in color). The best setting for each task and encoder combination is \textbf{bolded}. We order four encoders along two axes, similar to \autoref{tab:encoder}. 
}
\label{tab:main}
\end{center}
\end{table*}

%% file: table/projection/multilingual.tex
\begin{table*}[]
\begin{center}
\resizebox{\linewidth}{!}{
\begin{tabular}{ll ccccccccc c}

\toprule

\textbf{Encoder} & \textbf{Data} & \textbf{ar} & \textbf{de} & \textbf{en} & \textbf{es} & \textbf{fr} & \textbf{hi} & \textbf{ru} & \textbf{vi} & \textbf{zh} & \textbf{Average} \\

\midrule
\multicolumn{12}{l}{\textit{NER (F1)}} \\ 
\midrule
mBERT & Zero-shot & 41.6 & \textbf{78.8} & 83.9 & 73.1 & 79.5 & 66.2 & 63.4 & 70.8 & 51.8 & 67.7 \\
 & + Self & \muchbetter \textbf{+7.7} & -0.5 & \textbf{+0.4} & \better +4.8 & \best \textbf{+2.4} & \worse -2.5 & \better \textbf{+2.7} & \better \textbf{+1.2} & \better +1.4 & \better \textbf{+2.0} \\
 & + Proj & \muchworse -5.8 & -0.6 & +0.3 & \better +3.6 & +0.2 & \textbf{+0.4} & \worse -1.7 & \worse -2.0 & \better \textbf{+2.3} & -0.4 \\
 & + Proj (Bi) & +0.3 & -0.7 & +0.1 & \best \textbf{+5.2} & -0.6 & \worse -2.1 & \worse -1.1 & +0.3 & +0.0 & +0.2 \\
\midrule
XLM-R & Zero-shot & 46.4 & 79.5 & 83.9 & 76.1 & 80.0 & 70.9 & \best \textbf{70.5} & 77.0 & 40.2 & 69.4 \\
 & + Self & \best \textbf{+11.2} & \best \textbf{+0.9} & \best \textbf{+0.6} & \better \textbf{+1.0} & \textbf{+0.5} & \better +2.1 & \worse -1.5 & \best \textbf{+1.7} & \better +2.3 & \best \textbf{+2.1} \\
 & + Proj & \better +1.7 & -0.7 & -0.1 & \worse -3.9 & \worse -1.2 & \better +1.2 & \worse -4.8 & \muchworse -9.1 & \muchbetter +14.2 & -0.3 \\
 & + Proj (Bi) & \muchbetter +6.9 & +0.4 & -0.2 & \worse -4.3 & \worse -1.5 & \best \textbf{+3.2} & \worse -3.3 & \muchworse -5.2 & \best \textbf{+15.1} & \better +1.2 \\
\midrule
\multicolumn{12}{l}{\textit{POS (ACC)}} \\ 
\midrule
mBERT & Zero-shot & 59.7 & 89.6 & \textbf{96.9} & 87.5 & 88.7 & 69.5 & 81.9 & 62.6 & 66.6 & 78.1 \\
 & + Self & +0.3 & \textbf{+0.5} & +0.0 & \textbf{+0.4} & \textbf{+0.4} & -0.3 & \textbf{+0.5} & \textbf{+0.4} & \best \textbf{+1.7} & \textbf{+0.4} \\
 & + Proj & \muchbetter +6.9 & \worse -3.2 & +0.0 & \worse -3.8 & \worse -3.9 & \better +1.3 & \muchworse -6.6 & \muchworse -7.4 & \worse -4.1 & \worse -2.3 \\
 & + Proj (Bi) & \muchbetter \textbf{+8.5} & \worse -2.6 & -0.1 & \worse -3.2 & \worse -3.0 & \better \textbf{+1.6} & \muchworse -5.7 & \muchworse -6.9 & \worse -3.9 & \worse -1.7 \\
\midrule
XLM-R & Zero-shot & 73.3 & \best \textbf{91.5} & \best \textbf{98.0} & \best \textbf{89.3} & \best \textbf{90.0} & 78.6 & 86.8 & \best \textbf{65.2} & 53.6 & 80.7 \\
 & + Self & \best \textbf{+1.6} & -0.3 & +0.0 & +0.0 & +0.0 & \better \best \textbf{+2.0} & \best \textbf{+0.1} & -0.4 & \muchbetter \textbf{+11.7} & \best \textbf{+1.6} \\
 & + Proj & \muchworse -7.1 & \muchworse -5.4 & -0.5 & \muchworse -6.3 & \muchworse -5.9 & \muchworse -6.0 & \muchworse -10.5 & \muchworse -8.9 & \muchbetter +9.7 & \worse -4.6 \\
 & + Proj (Bi) & \muchworse -6.1 & \worse -4.6 & -0.1 & \worse -4.9 & \worse -4.6 & \muchworse -5.5 & \muchworse -10.4 & \muchworse -8.7 & \muchbetter +9.4 & \worse -4.0 \\
\midrule
\multicolumn{12}{l}{\textit{Parsing (LAS)}} \\ 
\midrule
mBERT & Zero-shot & 29.2 & \textbf{67.7} & 79.7 & \textbf{68.9} & \textbf{73.2} & 31.2 & \textbf{60.6} & \textbf{33.6} & \textbf{29.4} & \textbf{52.6} \\
 & + Self & \muchworse -20.6 & \muchworse -34.2 & +0.1 & \muchworse -41.6 & \muchworse -41.1 & \muchworse -15.3 & \muchworse -35.2 & \muchworse -17.8 & \muchworse -14.5 & \muchworse -24.5 \\
 & + Proj & \muchbetter \textbf{+9.1} & \worse -2.1 & \better \textbf{+1.1} & \worse -4.9 & \muchworse -5.8 & \muchbetter \textbf{+6.0} & \muchworse -5.6 & \muchworse -7.2 & \worse -2.1 & \worse -1.3 \\
 & + Proj (Bi) & \muchbetter +7.6 & \worse -1.6 & +0.5 & \worse -3.8 & \worse -4.5 & \muchbetter +5.7 & \worse -4.8 & \muchworse -7.2 & \worse -2.5 & \worse -1.2 \\
\midrule
XLM-R & Zero-shot & \best \textbf{48.0} & \best \textbf{69.6} & 82.6 & \best \textbf{73.6} & \best \textbf{76.1} & \best \textbf{43.1} & \best \textbf{70.3} & \best \textbf{38.4} & 15.0 & \best \textbf{57.4} \\
 & + Self & \muchworse -30.4 & \muchworse -29.4 & \best \textbf{+0.1} & \muchworse -39.9 & \muchworse -40.0 & \muchworse -18.3 & \muchworse -33.9 & \muchworse -16.1 & \muchworse -9.7 & \muchworse -24.2 \\
 & + Proj & \muchworse -8.5 & \worse -4.3 & +0.0 & \muchworse -10.3 & \muchworse -10.1 & \muchworse -5.7 & \muchworse -14.8 & \muchworse -11.1 & \muchbetter +14.5 & \muchworse -5.6 \\
 & + Proj (Bi) & \muchworse -8.4 & \worse -1.6 & \best \textbf{+0.1} & \muchworse -7.7 & \muchworse -7.4 & \worse -3.1 & \muchworse -12.7 & \muchworse -9.8 & \best \textbf{+15.1} & \worse -3.9 \\
\bottomrule
\end{tabular}
}
\caption{Performance of NER, POS, and parsing for eight target languages. We use the same color code as \autoref{tab:main}.}
\label{tab:multi}
\end{center}
\end{table*}

%% file: src/conclusions.tex
\section{Contributions}
In this thesis, we have attempted to answer a set of questions raised by the surprising cross-lingual effectiveness of Multilingual BERT. By understanding these models through analysis, we have identified ways to improve its cross-lingual representation.
In \autoref{chap:background}, we review the progress of representation learning in NLP and its impact on cross-lingual transfer, and observe that cross-lingual transfer performance improves as better representation learning techniques are developed.
With the release of Multilingual BERT (mBERT), in \autoref{chap:surprising-mbert}, we show that surprisingly mBERT learns cross-lingual representation even without explicit cross-lingual signal, and probes the released model to gain more insight.
In \autoref{chap:emerging-structure}, we conduct an ablation study on these multilingual models, demonstrating that parameter sharing of transformer contribute the most to the learning of cross-lingual representation, and show monolingual BERT of different language are similar to each other.
In \autoref{chap:low-resource}, we find that mBERT does not learn high quality representation for its lower resource languages, despite trying its best, as monolingual BERT or bilingual BERT paired with similar high resource language performs worse that mBERT for lower resource languages.
In \autoref{chap:crosslingual-signal}, we propose two methods for injecting different cross-lingual signal---bilingual dictionary and bitext---into these models, and show that while despite improvement on cross-lingual representation, it is eclipsed by the improvement of scaling up the model.
In \autoref{chap:analysis}, we show that zero-shot cross-lingual transfer is under-specified optimization, causing its high variance on target languages and much lower variance on source language. To address this issue, constraints need to be introduced into the optimization.
Thus, in \autoref{chap:data-projection}, we consider using silver target data---created automatically with machine translation based on supervision in source language---to constrain the optimization. We show that it indeed improves zero-shot cross-lingual transfer, despite the best setup of data creation pipeline with encoders is task specific.

\section{Future Works}

\subsection{Continue Scaling of Multilingual Encoders}

As we observe in \autoref{chap:crosslingual-signal}, while injecting cross-lingual signal explicitly into the model helps improve cross-lingual representation for smaller model, simply scaling up the model capacity and data size produces much better representation, as evidenced by \citet{xue-etal-2021-mt5} and \citet{goyal-etal-2021-larger}. In an orthogonal direction, as we discuss in \autoref{chap:low-resource}, improving the sample-efficiency of BERT pretraining objective likely leads to better representation, as evidenced by \citet{clark2020electra} and \citet{chi2021xlm}.

However, in this direction, there is one important open question: \textit{Is there a limit of scaling up model size?} In other words, Is there a peak model size, beyond which we will get the same or even worse cross-lingual representation? \citet{kaplan2020scaling} study the scaling law of monolingual language model, and observe that the cross-entropy loss of language model scales as a power-law with model size and data size. However, it is unclear that whether low cross-entropy loss in this scale translate to better cross-lingual representation. \citet{chi-etal-2021-infoxlm} argues that the cross-lingual representation of pretrained multilingual models is the product of the bottleneck effect, which suggests a scenario where the model learn each language well with low cross-entropy loss but fail to align the representation in the cross-lingual fashion. As we observe in \autoref{chap:emerging-structure}, such natural alignment is the key to the success of models like mBERT.

We hypothesize that this scenario might exist in theory but not in practise for model sharing all parameters across languages. As the model is trained with gradient descent mixing sentences of different languages, we will not discover such a solution with optimization. Additionally, scaling up model size is much slower compared to scaling up data size, as new data is continuously produced on the Web. Realistically, it is unlikely that we will hit such a limit any time soon even if it exists.

Looking at the current literature, we still observe better cross-lingual representation by scaling up to 10B parameters \cite{xue-etal-2021-mt5,goyal-etal-2021-larger}. There are models like GPT-3 \cite{brown2020language} with 100B+ parameters or GShard \cite{lepikhin2021gshard} and Switch Transformer \cite{fedus2021switch} with 1T+ parameters. Notably, these models are decoder-only or encoder-decoder based models. While the exact cross-lingual capability of these models are unclear at the moment, as these models are not open sourced. There is early evidence suggesting that these models might process similar cross-lingual capability \cite{winata2021language}. Thus, a further model size scaling of 100x is possible with the current techniques. However, such scaling requires answering new questions.

\textit{Can we develop a more scalable architecture and algorithm without sacrificing the natural alignment?} Scaling to 1T \cite{lepikhin2021gshard,fedus2021switch} rely heavily on sparse network or conditional computation---activating certain component of the network depending on the input---with mixture of experts \cite{jacobs1991adaptive,jordan1994hierarchical,shazeer2017outrageously}. However, it might harm the natural alignment across languages, as less parameter is shared across all languages compared to dense network. Thus, the model might learns better representation for each language, but worse cross-lingual representation in comparison. To address this challenge, possible directions include continue improvement of mixture of experts, Transformers, and more sample-efficient pretraining algorithm.

\textit{Can we efficiently adapt a large pretrained model to a task?} Such large models introduce challenges for fine-tuning. GPT-3 \cite{brown2020language}, a model with 175B parameters, addresses this challenge by relying on context-based few-shot learning, showing that with a prompt and some examples, the model can perform new tasks without fine-tuning. However, there is still much room for improvement. This is an active research area with new techniques like prompt fine-tuning \cite{li-liang-2021-prefix,qin-eisner-2021-learning,lester-etal-2021-power}. \citet{liu2021pre} survey the ongoing research.

\textit{Can we distill knowledge from a large pretrained model for deployment?} Large models introduce significant challenges for deployment. We could distill the knowledge from large pretrained models to smaller models to reduce inference time. It includes technique like knowledge-distillation \cite{hinton2015distilling}---transferring logits of large models on target languages (soft label)---and self-training \cite{yarowsky-1995-unsupervised} or semi-supervised learning---transferring the prediction of large models on target languages (hard label). As we show in \autoref{chap:data-projection}, self-training with the same model improves the cross-lingual transfer except for parsing. Additionally, the large model could be potentially pruned or quantized.

\subsection{Multilingual Multi-modals Models}

While this thesis mainly focuses on multilingual models for text, the lessons we learned may transfer to multilingual multi-modals models, e.g. speech, text + speech, text + image, text + video, and text + code. Similar to modern NLP systems, multi-modals systems usually need to support more than one language. Similar to text, we do not have the same amount of supervision for all languages. Thus, it is beneficial to consider the multilingual approach that enable cross-lingual transfer.
The main research question in this direction is that how well do lessons we learn from text encoder transfer to multi-modals models? Does cross-lingual representation emerge in multilingual multi-modals models?

%% file: src/CV.tex
\begin{vita}

Shijie Wu received a B.S. degree in Informational and Computational Science from Sun Yat-sen University in 2016 and a M.S.E. degree in Computer Science from Johns Hopkins University in 2018. Shijie enrolled in the Ph.D. program in Computer Science at Johns Hopkins University in 2018. Shijie's research focuses on cross-lingual transfer with pretrained multilingual encoders, ranging from discovering its cross-lingual potential, understanding why it works, documenting its caveat, to figuring out ways to improve it. Outside of the main area, Shijie also worked on papers related to morphology and computational linguistics.

\end{vita}